\documentclass[journal]{IEEEtran}
\usepackage{times}
\usepackage{ifthen}

\usepackage[numbers,sort&compress]{natbib}
\usepackage{multicol}
\usepackage[bookmarks=true]{hyperref}
\usepackage{xspace}
\usepackage{multirow}
 
\usepackage{amsmath}  
\usepackage{amssymb}
\usepackage{subfigure}
\usepackage[font=footnotesize]{caption}
\usepackage{color}
\usepackage{graphicx} % to trim image
\usepackage{booktabs} % for nice tables
\usepackage{tabularx} % to set the width of the table
\usepackage{lipsum} 
\usepackage{framed}

\newcommand{\hide}[1]{}
\newcommand{\Figure}{Fig.~}
\newcommand{\Equation}{Eq.~}

\newcommand{\cf}{\emph{cf.~}}
\newcommand{\bdmath}{\begin{dmath}}
\newcommand{\edmath}{\end{dmath}}
\newcommand{\beq}{\begin{equation}}
\newcommand{\eeq}{\end{equation}}
\newcommand{\bdm}{\begin{displaymath}}
\newcommand{\edm}{\end{displaymath}}
\newcommand{\bea}{\begin{eqnarray}}
\newcommand{\eea}{\end{eqnarray}}
\newcommand{\beal}{\beq \begin{array}{ll}}
\newcommand{\eeal}{\end{array} \eeq}
\newcommand{\beas}{\begin{eqnarray*}}
\newcommand{\eeas}{\end{eqnarray*}}
\newcommand{\ba}{\begin{array}}
\newcommand{\ea}{\end{array}}
\newcommand{\bit}{\begin{itemize}}
\newcommand{\eit}{\end{itemize}}
\newcommand{\ben}{\begin{enumerate}}
\newcommand{\een}{\end{enumerate}}

\newcommand{\SO}{\mathrm{SO}}
\newcommand{\so}{\mathfrak{so}}

\newcommand{\Real}{\mathbb{R}}
\newcommand{\Int}{\mathbb{Z}}

\newcommand{\SEthree}{\ensuremath{\mathrm{SE}(3)}\xspace}

\newcommand{\SOthree}{\ensuremath{\SO(3)}\xspace}
\newcommand{\sothree}{\ensuremath{\so(3)}\xspace}
\newcommand{\setdef}[2]{ \{#1 \; {:} \; #2 \} }

\newcommand{\calJ}{{\cal J}}

\newcommand{\calL}{{\cal L}}
\newcommand{\calM}{{\cal M}}
\newcommand{\calN}{{\cal N}}

\newcommand{\calR}{{\cal R}}

\newcommand{\T}{\mathtt{T}}
\newcommand{\R}{\mathtt{R}}

\newcommand{\Identity}{\mathbf{I}}
\newcommand{\inv}{^{-1}}

\newcommand{\trace}[1]{\mathrm{tr}\left(#1\right)}

\newcommand{\residual}{\mathbf{r}}
\newcommand{\transpose}{\mathsf{T}}
\newcommand{\Zero}{\mathbf{0}}
\newcommand{\eye}{{\mathbf I}}

\newcommand{\Ithree}{\eye_{3 \times 3}}

\newcommand{\rotvel}{\boldsymbol\omega}
\newcommand{\acc}{\mathbf{a}}
\newcommand{\rotvec}{\boldsymbol\phi}
\newcommand{\rotvecpert}{\delta\rotvec}
\newcommand{\rotvecang}{\varphi}	% rotation vector angle
\newcommand{\rotvecdir}{\mathbf{a}} % rotation vector direction
\newcommand{\tran}{\mathbf{p}}
\newcommand{\tranpert}{\delta\tran}
\newcommand{\vel}{\mathbf{v}}
\newcommand{\velpert}{\delta\vel}
\newcommand{\bias}{\mathbf{b}}
\newcommand{\biaspert}{\delta\bias}
\newcommand{\gravity}{\mathbf{g}}
\newcommand{\noise}{\boldsymbol\eta}
\newcommand{\VIO}{VIO\xspace}

\newcommand{\tango}{{\rm Tango}\xspace}

\newcommand{\Cov}{\mathbf{\Sigma}}

\newcommand{\expmap}{\mathrm{Exp}}
\newcommand{\logmap}{\mathrm{Log}}
\newcommand{\normsq}[2]{\left\|#1\right\|^2_{#2}}
\renewcommand{\d}{\text{d}}
\newcommand{\Rmean}{\R}
\newcommand{\Rrand}{\tilde\R}
\newcommand{\eq}{Eq.}
\newcommand{\eqs}{Eqs.}

\newcommand{\MAP}{MAP\xspace}

\newcommand{\GN}{GN\xspace}
\newcommand{\States}{\mathcal{X}}
\newcommand{\meascam}{\mathcal{C}}
\newcommand{\measimu}{\mathcal{I}}
\newcommand{\covprior}{\mathbf{\Sigma}}
\newcommand{\covimu}{\mathbf{\Sigma}}
\newcommand{\covcam}{\mathbf{\Sigma}_{\meascam}}
\newcommand{\World}{\text{W}}
\newcommand{\Imu}{\text{B}}

\newcommand{\world}{\text{\tiny{W}}}
\newcommand{\imu}{\text{\tiny{B}}}
\newcommand{\camera}{\text{\tiny{C}}}

\newcommand{\accshort}{\tilde\acc(t)\!-\! \bias^a(t) \!-\! \noise^{ad}(t)}

\newcommand{\biasFixed}{\bar{\bias}}

\newcommand{\biasHat}{\hat{\bias}}
\newcommand{\meters}{\rm{m}}
\newcommand{\preintRmeas}{\Delta\tilde\R}
\newcommand{\preintVmeas}{\Delta\tilde\vel}
\newcommand{\preintPmeas}{\Delta\tilde\tran}

\newcommand{\pixel}{\mathbf{z}}

\newcommand{\landmark}{\mathbf{\rho}}
\newcommand{\landpert}{\delta\landmark}
\newcommand{\posepert}{\delta\mathbf{T}}
\newcommand{\bvec}{\mathbf{b}}
\newcommand{\Amat}{\mathbf{A}}
\newcommand{\Bmat}{\mathbf{B}}
\newcommand{\Fmat}{\mathbf{F}}
\newcommand{\Emat}{\mathbf{E}}
\newcommand{\Qmat}{\mathbf{Q}}

\newcommand{\subimu}{{ij}} 
\newcommand{\indmeas}{(i,j)}

\newcommand{\DLog}{\mathtt{J}_{r}^{-1}} 
\newcommand{\DExp}{\mathtt{J}_{r}} 
\newcommand{\DExpk}{\mathtt{J}^k_{r}} 
\newcommand{\jacobian}[2]{\frac{\partial #1}{\partial #2}}

\newcommand{\rup}{\rotvecpert} 
\newcommand{\tup}{\tranpert} 
\newcommand{\vup}{\velpert} 
\newcommand{\bupa}{\tilde\delta{\bias^a_i}} 
\newcommand{\bupg}{\tilde\delta{\bias^g_i}}

\newcommand{\pRot}{\Delta\bar\R_{ij} } % integration value
\newcommand{\pRotk}{\Delta\bar\R_{ik} } % integration value
\newcommand{\pTran}{\Delta\bar\tran_{ij}}
\newcommand{\pVel}{\Delta\bar\vel_{ij}}
\newcommand{\pVelk}{\Delta\bar\vel_{ik}}
\newcommand{\mTran}{\Delta\tilde\tran_{ij}}
\newcommand{\mVel}{\Delta\tilde\vel_{ij}}

\newcommand{\figdirvin}{.}

\newcommand{\videolink}{https://youtu.be/CsJkci5lfco}

\begin{document}

\onecolumn

\hspace{3cm}
\begin{center}
This paper has been accepted for publication in \emph{IEEE Transactions on Robotics}.\\

\hspace{1cm}

DOI: \href{http://dx.doi.org/10.1109/TRO.2016.2597321}{10.1109/TRO.2016.2597321}\\

IEEE Explore: \url{http://ieeexplore.ieee.org/document/7557075/} \\

\hspace{1cm}

Please cite the paper as: \\

\hspace{1cm}

Christian Forster, Luca Carlone, Frank Dellaert, Davide Scaramuzza, \\
``On-Manifold Preintegration for Real-Time Visual-Inertial Odometry'', \\
in \emph{IEEE Transactions on Robotics}, 2016.\\

\end{center}
\twocolumn

\author{Christian Forster, Luca Carlone, Frank Dellaert, Davide Scaramuzza\thanks{
C.\,Forster is with the Robotics and Perception Group, University of Zurich, Switzerland.
E-mail: {\sf forster@ifi.uzh.ch}

L.\,Carlone is with the Laboratory for Information \& Decision Systems, Massachusetts Institute of Technology, USA. 
E-mail: {\sf lcarlone@mit.edu}

F.\,Dellaert is with the College of Computing, Georgia Institute of Technology, USA. 
E-mail: {\sf dellaert@cc.gatech.edu}

D.\,Scaramuzza is with the Robotics and Perception Group, University of Zurich, Switzerland.
E-mail: {\sf sdavide@ifi.uzh.ch}

This research was partially funded by the Swiss National Foundation (project number 200021-143607, ``Swarm of Flying Cameras''), the National Center of Competence in Research Robotics (NCCR), the UZH Forschungskredit, the NSF Award 11115678, and the USDA NIFA Award GEOW-2014-09160.
}}

\title{On-Manifold Preintegration for Real-Time Visual-Inertial Odometry}

\maketitle

%%%%%%%%%%%%%%%%%%%%%%%%%%%%%%%%%%%%%%%%%%%%%%%%%%%%%%%%%%%%%%%%%%%%%%%
\begin{abstract}
%!TEX root = main.tex
Current approaches for visual-inertial odometry (VIO) are able to attain highly accurate 
state estimation via nonlinear optimization.
However, real-time optimization quickly becomes infeasible as the trajectory grows over time; 
this problem is further emphasized by the fact that inertial measurements come at high rate, 
hence leading to fast growth of the number of variables in the optimization.
% 1. Contribution
In this paper, we address this issue by preintegrating inertial measurements between selected keyframes into single relative motion constraints.
Our first contribution is a \emph{preintegration theory} that properly addresses the manifold structure of the rotation group.
We formally discuss the generative measurement model as well as the nature of the rotation noise and derive the expression 
for the \emph{maximum a posteriori} state estimator.
Our theoretical development enables the computation of all necessary Jacobians for the optimization and a-posteriori bias correction in analytic form.
% 2. Contribution 
The second contribution is to show that the preintegrated IMU model can be seamlessly integrated into a visual-inertial pipeline under the unifying framework of factor graphs.
This enables the application of incremental-smoothing algorithms and the use of a \emph{structureless} model for visual measurements, 
which avoids optimizing over the 3D points, further accelerating the computation. 
% 3. Contribution
We perform an extensive evaluation of our monocular \VIO pipeline on real and simulated datasets.
The results confirm that our modelling effort leads to accurate state estimation in 
real-time, outperforming state-of-the-art approaches.
\end{abstract}

\section*{Supplementary Material}
\begin{itemize}
	\item Video of the experiments: \url{\videolink}
	\item Source-code for preintegrated IMU and structureless vision factors \url{https://bitbucket.org/gtborg/gtsam}.
\end{itemize}

%!TEX root = main.tex
\section{Introduction}

%% IMPORTANCE OF VISUAL INERTIAL NAVIGATION
The use of cameras and inertial sensors for three-dimensional structure and motion estimation has 
received considerable attention from the robotics community. 
Both sensor types are cheap, ubiquitous, and complementary.
A single moving camera is an exteroceptive sensor that allows us to measure appearance and geometry of a three-dimensional scene, up to an unknown metric scale; an inertial measurement unit (IMU) is a proprioceptive sensor that renders metric scale of monocular vision and gravity observable \cite{Martinelli12tro} and provides robust and accurate inter-frame motion estimates.
Applications of \VIO range from autonomous navigation in GPS-denied environments, to 3D reconstruction, and augmented reality.

%% Limitations of related work
The existing literature on VIO imposes a trade-off between accuracy and computational efficiency 
(a detailed review is given in Section~\ref{sec:related_work}).
On the one hand, filtering approaches enable fast inference, but their accuracy is deteriorated by 
the accumulation of linearization errors.
On the other hand, full smoothing approaches, based on nonlinear optimization, are accurate, but computationally demanding. 
Fixed-lag smoothing offers  a compromise between accuracy for efficiency; however, 
it is not clear how to set the length of the estimation window so to 
guarantee a given level of performance. 

In this work we show that it is possible to overcome this trade-off. We design a \VIO system that 
enables fast incremental smoothing and computes the optimal \emph{maximum  a posteriori} (\MAP) estimate in real time.
An overview of our approach is given in Section~\ref{sec:ml_vi}.

%% CONTRIBUTION 1
The first step towards this goal is the development of a novel preintegration theory. 
The use of \emph{preintegrated IMU measurements} was first proposed in \cite{Lupton12tro} and consists of combining many inertial measurements between two keyframes into a single relative motion constraint.
We build upon this work and present a preintegration theory that properly addresses the manifold structure of the rotation group $\SOthree$. 
Our preintegration theory is presented in Sections~\ref{sec:imu_model}-\ref{sec:preintegration}.
Compared with~\cite{Lupton12tro}, our theory offers a more formal treatment of the rotation noise, and avoids singularities in the representation of rotations.
Furthermore, we are able to derive all necessary Jacobians in analytic form: specifically, we report the analytic Jacobians of the residuals, the noise propagation, and the a-posteriori bias correction in the appendix of this paper.

% CONTRIBUTION 2
Our second contribution is to frame the IMU preintegration theory into a factor graph model.
This enables the application of incremental smoothing algorithms, as iSAM2~\cite{Kaess12ijrr}, which avoid the accumulation of linearization errors and offer an elegant way to trade-off  accuracy with efficiency.
Inspired by~\cite{Carlone14icra,Mourikis07icra}, we also adopt a \emph{structureless} model for visual measurements, which allows eliminating a large number of variables (\emph{i.e.}, all 3D points) during incremental smoothing, further accelerating the computation (Section~\ref{sec:structureless}).
In contrast to \cite{Mourikis07icra}, we use the structureless model in an incremental smoothing framework.
This has two main advantages: we do not need to  delay the processing of visual measurements, and we can relinearize the visual measurements multiple times.

In order to demonstrate the effectiveness of our model, we integrated the proposed IMU preintegration in a state-of-the-art \VIO pipeline and tested it on real and simulated datasets (Sections~\ref{sec:experiments}).
Our theoretical development leads to tangible practical advantages: an implementation of the approach proposed in this paper performs full-smoothing at a rate of 100~Hz and achieves superior accuracy with respect to competitive state-of-the-art filtering and optimization approaches.

% Tutorial
Besides the technical contribution, the paper also provides a tutorial contribution for practitioners.
In Section \ref{sec:preliminaries} and across the paper, we provide a short but concise summary of uncertainty representation on manifolds and exemplary derivations for uncertainty propagation and Jacobian computation. 
The complete derivation of all equations and Jacobians -- necessary to implement 
our model -- are given in the appendix.

This paper is an extension of our previous work \cite{Forster15rss} with additional experiments, an in-depth discussion of related work, and comprehensive technical derivations.
The results of the new experiments highlight the accuracy of bias estimation, demonstrate the consistency of our approach, and provide comparisons against full batch estimation.
We release our implementation of the preintegrated IMU and structureless vision factors in the GTSAM 4.0 optimization toolbox \cite{Dellaert12tr}.
%!TEX root = main.tex
\section{Related Work}
\label{sec:related_work}

% Nomunclature
Related work on visual-inertial odometry can be sectioned along three main dimensions.
The first dimension is the number of camera-poses involved in the estimation. While \emph{full smoothers} (or \emph{batch 
nonlinear least-squares} algorithms) estimate the complete history of poses,
\emph{fixed-lag smoothers} (or \emph{sliding window estimators}) consider a window of the latest poses, 
and \emph{filtering} approaches only estimate the latest state.
Both fixed-lag smoothers and filters marginalize older states and absorb the corresponding information in a Gaussian prior.

The second dimension regards the representation of the uncertainty for the measurements and the Gaussian priors: 
the \emph{Extended Kalman Filter} (EKF) represents the uncertainty using a covariance matrix; instead, \emph{information filters} and smoothers resort to the information matrix (the inverse of the covariance) or the square-root of the information matrix \cite{Kaess12ijrr,Wu15rss}.

Finally, the third dimension distinguishes existing approaches by looking at the
number of times in which the measurement model is linearized. While a standard EKF (in contrast to the \emph{iterated} EKF) processes a measurement only once, a smoothing approach allows linearizing multiple times.

While the terminology is vast, the underlying algorithms are tightly related. For instance, it can be shown that the iterated Extended Kalman filter equations are equivalent to the Gauss-Newton algorithm, commonly used for smoothing~\cite{Bell93}.

%%%%%%%%%%%%%%%%%%%%%%%%%%%%%%%%%%%%%%%%%%%%%%%%%%%%%%%%%%%%%%%%%%%%%%%%%%%%%%%%%%%%%%%%%%%%%%%%%%%%%%%%%%%%
\subsection{Filtering}
Filtering algorithms enable efficient estimation by restricting the inference process to the latest state of the system.
The complexity of the EKF grows quadratically in the number of estimated landmarks, therefore, 
a small number of landmarks (in the order of 20) are typically tracked to allow real-time operation~\cite{Davison07pami,Blosch15iros,Jones11ijrr}.
An alternative is to adopt a ``structureless'' approach where landmark positions are marginalized out of the state vector. 
An elegant example of this strategy is the \emph{Multi-State Constraint Kalman filter} (MSC-KF)~\cite{Mourikis07icra}.
The structureless approach requires to keep previous poses in the state vector, 
by means of \emph{stochastic cloning} \cite{Roumeliotis02icra}.

A drawback of using a structureless approach for filtering, 
is that the processing of landmark measurements needs to be delayed until all measurements of a landmark are obtained~\cite{Mourikis07icra}.
This hinders accuracy as the filter cannot use all current visual information.
Marginalization is also a source of errors as it locks in linearization errors and erroneous outlier measurements.
Therefore, it is particularly important to filter out spurious measurements as a single 
outlier can irreversibly corrupt the filter \cite{Tsotsos15icra}.
Further, linearization errors introduce drift in the estimate and render the filter \emph{inconsistent}.
An effect of inconsistency is that the estimator becomes over-confident, resulting in non-optimal information fusion.
Generally, the \VIO problem has four unobservable directions: the global position and the orientation around the gravity direction (yaw) \cite{Martinelli13fntrob,Kottas2012iser}. 
In \cite{Kottas2012iser} it is shown that linearization at the wrong estimate results in only three unobservable directions (the global position); hence, erroneous linearization adds spurious information in yaw direction to the Gaussian prior, which renders the filter inconsistent.
This problem was addressed with the \emph{first-estimates jacobian} approach \cite{Huang08iser}, which ensures that a state is not updated with different linearization points --- a source of inconsistency.
In the \emph{observability-constrained} EKF (OC-EKF) an estimate of the unobservable directions is maintained which allows to update the filter only in directions that are observable \cite{Kottas2012iser,Hesch14ijrr}.
A thorough analysis of \VIO observability properties is given in \cite{Martinelli12tro,Martinelli13fntrob,Hernandez15icra}.

%%%%%%%%%%%%%%%%%%%%%%%%%%%%%%%%%%%%%%%%%%%%%%%%%%%%%%%%%%%%%%%%%%%%%%%%%%%%%%%%%%%%%%%%%%%%%%%%%%%%%%%%%%%%
\subsection{Fixed-lag Smoothing}
Fixed-lag smoothers estimate the states that fall within a given time 
window, while marginalizing out older states~\cite{Mourikis08wvlmp,Sibley10jfr,DongSi11icra,Leutenegger13rss,Leutenegger15ijrr}.
In a maximum likelihood estimation setup, fixed-lag smoothers lead to an optimization problem over a set of recent states.
For nonlinear problems, smoothing approaches are generally more accurate than filtering, since they relinearize past measurements~\cite{Maybeck79}.
Moreover, these approaches are more resilient to outliers, which can be discarded a posteriori 
(i.e., after the optimization), or can be alleviated by using robust cost functions.
On the downside, the marginalization of the states outside the estimation window 
 leads to dense Gaussian priors 
which hinder efficient inference.  
For this reason, it has been proposed to drop certain measurements in the interest of sparsity  \cite{Leutenegger15ijrr}.
Furthermore, due to marginalization, 
fixed-lag smoothers share part of the issues of filtering (consistency, build-up of linearization errors)~\cite{Huang11iros,DongSi11icra,Hesch14ijrr}.

%%%%%%%%%%%%%%%%%%%%%%%%%%%%%%%%%%%%%%%%%%%%%%%%%%%%%%%%%%%%%%%%%%%%%%%%%%%%%%%%%%%%%%%%%%%%%%%%%%%%%%%%%%%%
\subsection{Full Smoothing}
Full smoothing methods estimate the entire history of the states (camera trajectory and 3D landmarks), by solving a large nonlinear optimization problem
\cite{Jung01cvpr,Sterlow04ijrr,Bryson09icra,Indelman13ras,PatronPerez15}. 
Full smoothing guarantees the highest accuracy;
however, real-time operation quickly becomes infeasible as the trajectory and the map grow over time.
Therefore, it has been proposed to discard frames except selected \emph{keyframes} \cite{Strasdat10icra,Klein09ismar,Nerurkar14icra,Leutenegger15ijrr} or to run the optimization in a parallel thread, using a tracking and mapping dual architecture \cite{Klein07ismar,Mourikis08wvlmp}.
A breakthrough has been the development of \emph{incremental smoothing techniques} (iSAM~\cite{Kaess08tro}, iSAM2~\cite{Kaess12ijrr}), 
which leverage the expressiveness of~\emph{factor graphs} to maintain sparsity and to identify and update only the typically small subset of variables affected by a new measurement.

% IMU Preintegration
Nevertheless, the high rate of inertial measurements (usually 100~Hz to 1~kHz) still constitutes a challenge for smoothing approaches. A naive implementation would require adding a new state at every IMU measurement, which quickly becomes impractically slow~\cite{Indelman12fusion}.
Therefore, inertial measurements are typically integrated between frames to form relative motion constraints \cite{Indelman13iros,Indelman13ras,Shen14thesis,keivan14iser,Leutenegger15ijrr}.
For standard IMU integration between two frames, the initial condition is given by the state estimate at the first frame.
However, at every iteration of the optimization, the state estimate changes, which requires to repeat the IMU integration between all frames \cite{Leutenegger15ijrr}.
Lupton and Sukkarieh \cite{Lupton12tro} show that this repeated integration can be 
avoided by a reparametrization of the relative motion constraints. Such reparametrization is called \emph{IMU preintegration}.

In the present work, we build upon the seminal work \cite{Lupton12tro} and bring the theory of IMU preintegration to maturity by properly addressing the manifold structure of the rotation group $\SOthree$.
The work~\cite{Lupton12tro} adopted Euler angles as global parametrization for rotations.
Using Euler angles and applying the usual averaging and smoothing techniques of Euclidean spaces for state propagation and covariance estimation is not properly invariant under the action of rigid transformations~\cite{Hornegger99,Moakher02siam}.
Moreover, Euler angles are known to have singularities.
Our work, on the other hand, provides a formal treatment of the rotation measurements 
(and the corresponding noise), and provides a complete derivation of the maximum a posteriori 
estimator. 
We also derive analytic expressions for the Jacobians (needed for the optimization), which, to the best of our knowledge, 
have not been previously reported in the literature.
In the experimental section, we show that a proper representation of the rotation manifold results in higher 
accuracy and robustness, leading to tangible advantages over the original proposal~\cite{Lupton12tro}.

%!TEX root = main.tex
\section{Preliminaries}
\label{sec:preliminaries}

In this paper we formulate \VIO in terms of MAP estimation. 
In our model, MAP estimation leads to a nonlinear optimization problem that involves quantities living on smooth manifolds (e.g., rotations, poses).
Therefore, before delving into details, we conveniently review some useful geometric concepts.
This section can be skipped by the expert reader. 

We structure this section as follows:
Section~\ref{sec:manifolds} provides useful notions related to two main Riemannian manifolds: the Special Orthogonal Group \SOthree and the Special Euclidean Group \SEthree.
Our presentation is based on~\cite{Chirikjian12book, Wang08ijrr}.
Section~\ref{sec:uncertaintySO3} describes a suitable model to describe uncertain rotations in \SOthree.  
Section~\ref{sec:GNmanifold} reviews optimization on manifolds, following standard references~\cite{Absil07fcm}.

\subsection{Notions of Riemannian geometry}
\label{sec:manifolds}

\paragraph{Special Orthogonal Group} 

\SOthree describes the group of 3D rotation matrices and it is formally defined as
$
  \SOthree \doteq \setdef{\R \in \Real^{3\times3}}{\R^\transpose \R = \eye, \det(\R)=1}
$.
The group operation is the usual matrix multiplication, and the inverse is the matrix transpose.
The group \SOthree also forms a smooth manifold.
The tangent space to the manifold (at the identity) is denoted  as \sothree, which is also called the \emph{Lie algebra} and coincides with the space of $3 \times 3$ skew symmetric matrices. 
We can identify every skew symmetric matrix with a vector in $\Real^3$ using the \emph{hat} operator:
\beq
  \boldsymbol\omega^\wedge = 
  \begin{bmatrix}
    \omega_1         \\
    \omega_2  \\
    \omega_3 
   \end{bmatrix}^\wedge =
   \begin{bmatrix}
    0         & -\omega_3 & \omega_2 \\
    \omega_3  & 0         & -\omega_1 \\
    -\omega_2 & \omega_1  & 0
   \end{bmatrix} \in \sothree.
\eeq 
Similarly, we can map a skew symmetric matrix to a vector in $\Real^3$ using the \emph{vee} operator $(\cdot)^\vee$:
for a skew symmetric matrix $\mathtt{S} = \boldsymbol\omega^\wedge$, the vee operator is such that $\mathtt{S}^\vee = \boldsymbol\omega$.
A property of skew symmetric matrices that will be useful later on is:
\beq\label{eq:skewProperty}
  \mathbf{a}^\wedge \; \mathbf{b} = -\mathbf{b}^\wedge \; \mathbf{a}, \quad \forall \; \mathbf{a},\mathbf{b}\in \Real^3.
\eeq

The \emph{exponential map} (at the identity) $\exp: \sothree \rightarrow \SOthree$ associates an element of the Lie Algebra to a rotation and coincides with standard matrix exponential (Rodrigues' formula): 
\beq
\textstyle
  \exp(\rotvec^\wedge)
  = \eye + \frac{\sin(\|\rotvec\|)}{\|\rotvec\|} \rotvec^\wedge
  + \frac{1-\cos(\|\rotvec\|)}{\|\rotvec\|^2} \left( \rotvec^\wedge \right)^2.
\eeq 
A first-order approximation of the exponential map that we will use later on is:
\beq\label{eq:expFirstOrder}
\textstyle
  \exp(\rotvec^\wedge) \approx \Identity + \rotvec^\wedge \;.
\eeq

\begin{figure}[t]
  \centering
  \includegraphics[width=0.75\linewidth]{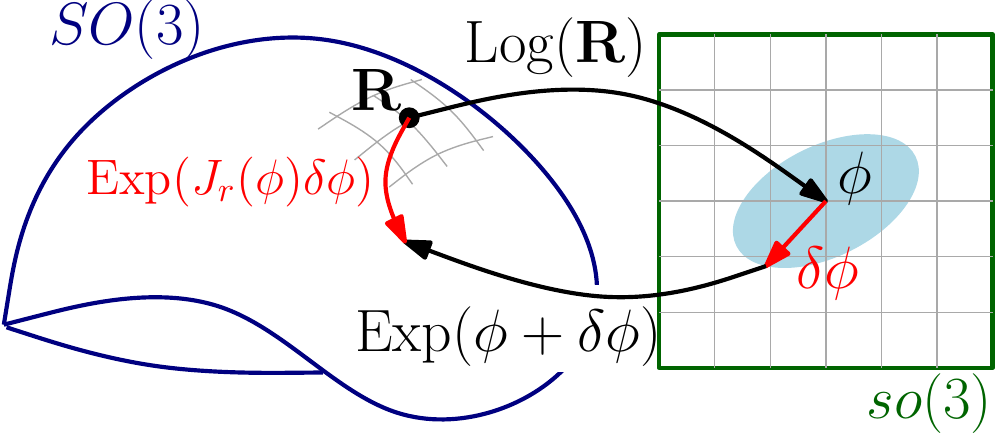}
  \caption{The right Jacobian $\DExp$ relates an additive perturbation $\rotvecpert$ in the tangent space to a multiplicative perturbation on the manifold $\SOthree$,  as per \eq~\eqref{eq:expExpansion}.}
  \label{fig:tangent_space}
  \vspace{-15px}
\end{figure}

The \emph{logarithm map} (at the identity) associates a matrix $\R\neq\Identity$ in $\SOthree$ to a skew symmetric matrix:
\beq
  \log(\R) = \frac{\rotvecang \cdot ( \R - \R^\transpose ) }{2\sin(\rotvecang)} 
  \text{ with }
  \rotvecang = \cos\inv \left(\frac{\trace{\R}-1}{2} \right).
\eeq
Note that $\log(\R)^\vee = \rotvecdir \rotvecang$, where $\rotvecdir$ and $\rotvecang$ are the rotation axis and the rotation angle of $\R$, respectively. 
If $\R=\Identity$, then $\rotvecang=0$ and $\rotvecdir$ is undetermined and can therefore be chosen arbitrarily.
 
The exponential map is a bijection if restricted to the open ball $\|\rotvec\|<\pi$, and the corresponding inverse is the logarithm map.
However, if we do not restrict the domain, the exponential map becomes surjective as every vector $\rotvec = (\rotvecang+2k\pi)\rotvecdir$, $k\in\Int$ would be an admissible logarithm of $\R$. 

For notational convenience, we adopt ``vectorized'' versions of the exponential and logarithm map:
\hide{
\beq
\begin{array}{cccccccc}
  \label{eq:vectorExpLog}
  \expmap :& \Real^3  &\ni& \rotvec & \rightarrow & \exp(\rotvec^\wedge) &\in& \SOthree, \\
  \logmap :& \SOthree &\ni& \R      & \rightarrow & \log(\R)^\vee        &\in& \Real^3,
\end{array}
\eeq
}
\beq
\begin{array}{cccccccc}
  \label{eq:vectorExpLog}
  \expmap :& \Real^3  &\rightarrow& \SOthree &;& \rotvec  &\mapsto& \exp(\rotvec^\wedge)\\
  \logmap :& \SOthree &\rightarrow& \Real^3  &;& \R       &\mapsto& \log(\R)^\vee,
\end{array}
\eeq
which operate directly on vectors, rather than on skew symmetric matrices in \sothree. 
   
Later, we will use the following first-order approximation:
\beq\label{eq:expExpansion}
  \expmap(\rotvec+\rotvecpert) \approx \expmap(\rotvec) \; \expmap(\DExp(\rotvec)\rotvecpert).
\eeq
The term $\DExp(\rotvec)$ is the \emph{right Jacobian} of $\SOthree$ \cite[p.40]{Chirikjian12book} and relates additive increments in the tangent space to multiplicative increments applied on the right-hand-side (\Figure~\ref{fig:tangent_space}):

\beq
  \label{eq:rightJacobian}
  \textstyle
  \DExp(\rotvec) = \Identity - \frac{1-\cos(\|\rotvec\|)}{\|\rotvec\|^2}\rotvec^\wedge
  +\frac{\| \rotvec\| - \sin(\|\rotvec\|)}{\|\rotvec^3\|}(\rotvec^\wedge)^2.
  \;
\end{equation}
A similar first-order approximation holds for the logarithm:
\beq
  \label{eq:logExpansion}
  \logmap\big(\; \expmap(\rotvec) \; \expmap(\rotvecpert) \;\big)\approx \rotvec+\DLog(\rotvec)\rotvecpert.
\eeq
Where the inverse of the right Jacobian is
\beq
  \DLog(\rotvec) = 
  \Identity + \frac{1}{2}\rotvec^\wedge 
  + \left( \frac{1}{\|\rotvec\|^2} 
          +\frac{1+\cos(\|\rotvec\|)}{2\|\rotvec\|\sin(\|\rotvec\|)} \right)(\rotvec^\wedge)^2.
  \; \nonumber
\eeq
The right Jacobian $\DExp(\rotvec)$ and its inverse $\DLog(\rotvec)$ reduce to the identity matrix for $\|\rotvec\| \!=\!0$.

Another useful property of the exponential map is:
\begin{align}
  \R\ \expmap(\rotvec)\ \R^\transpose &= \exp(\R\rotvec^\wedge\R^\transpose) = \expmap(\R\rotvec)
  \label{eq:adjointProperty1}\\
  \Leftrightarrow \quad \expmap(\rotvec)\ \R &= \R\ \expmap(\R^\transpose\rotvec).
  \label{eq:adjointProperty2}
\end{align}

%%%%%%%%%%%%%%%%%%%%%%%%%%%%%%%%%%%%%%%%%%%%%%%%%%%%%%%%%%%%%%%%%%%%%%%%%%%%%%%
\paragraph{Special Euclidean Group} 
\SEthree describes the group of rigid motion in 3D, which is the semi-direct product of $\SOthree$ and $\Real^3$, and it is defined as $\SEthree \doteq 
\setdef{(\R,\tran)}{\R \in \SOthree, \tran \in \Real^{3}}$.
Given $\T_1, \T_2 \in\SEthree$, the group operation is 
$\T_1 \cdot \T_2 = (\R_1 \R_2 \;,\; \tran_1 + \R_1 \tran_2)$, 
and the inverse is 
$\T_1\inv = (\R_1^\transpose \;,\; -\R_1^\transpose \tran_1)$. 
The \emph{exponential map} and the \emph{logarithm map} for \SEthree are defined in~\cite{Wang08ijrr}.
However, these are not needed in this paper for reasons that will be clear in Section~\ref{sec:GNmanifold}.  

%%%%%%%%%%%%%%%%%%%%%%%%%%%%%%%%%%%%%%%%%%%%%%%%%%%%%%%%%%%%%%%%%%%%%%%%%%%%%%%
\subsection{Uncertainty Description in \texorpdfstring{$\SOthree$}{SO(3)}}
\label{sec:uncertaintySO3}

A natural definition of uncertainty in $\SOthree$ is to define a distribution in the tangent space, and then map it to $\SOthree$ via the exponential map~\eqref{eq:vectorExpLog}~\cite{Barfoot14tro,Wang06tro,Wang08ijrr}:
\beq
\label{eq:rightPerturbation}
\Rrand =\Rmean \; \expmap(\epsilon),\qquad \epsilon \sim \calN(0,\Sigma),
\eeq
where $\Rmean$ is a given noise-free rotation (the \emph{mean}) and $\epsilon$ is a small normally distributed perturbation with zero mean and covariance $\Sigma$.
 
To obtain an explicit expression for the distribution of $\Rrand$, we start from the integral of the Gaussian distribution in $\Real^3$:
\beq
  \label{eq:integralR3}
  \int_{\Real^3} p(\epsilon) d\epsilon = \int_{\Real^3} \alpha e^{-\frac{1}{2} \normsq{\epsilon}{\Sigma} } \d\epsilon = 1,
\eeq
where $\alpha = 1 / \sqrt{(2\pi)^3 \det(\Sigma)}$ and $\normsq{\epsilon}{\Sigma} \doteq \epsilon^\transpose\Sigma^{-1}\epsilon$ is the squared Mahalanobis distance with covariance $\Sigma$.
Then, applying the change of coordinates $\epsilon = \logmap(\Rmean\inv \Rrand)$ (this is the inverse of~\eqref{eq:rightPerturbation} when $\|\epsilon\| < \pi$), the integral~\eqref{eq:integralR3} becomes:
\beq
  \label{eq:integralSO3}
  \int_{\SOthree} \beta(\Rrand) \; e^{-\frac{1}{2} \normsq{\logmap(\Rmean\inv \Rrand)}{\Sigma} } \; \d \Rrand = 1,
\eeq
where $\beta(\Rrand)$ is a normalization factor.  
The normalization factor assumes the form $\beta(\Rrand) = \alpha / |\det(\calJ(\Rrand)|$, where $\calJ(\Rrand) \doteq \DExp(\logmap(\Rmean\inv \Rrand))$ and $\DExp(\cdot)$ is the right Jacobian~\eqref{eq:rightJacobian}; $\calJ(\Rrand)$ is a by-product of the change of variables, see~\cite{Barfoot14tro} for a derivation.

From the argument of~\eqref{eq:integralSO3} we can directly read our ``Gaussian'' distribution in \SOthree:
\beq
\label{eq:gaussianSO3}
p(\Rrand) = \beta(\Rrand) \; e^{-\frac{1}{2} \normsq{\logmap(\Rmean\inv \Rrand)}{\Sigma} }.
\eeq

For small covariances we can approximate $\beta \simeq \alpha$, as $\DExp(\logmap(\Rmean\inv \Rrand))$ is well approximated by the identity matrix when $\Rrand$ is close to $\Rmean$. Note that~$\eqref{eq:integralSO3}$ already assumes relatively a small covariance $\Sigma$, since it ``clips'' the probability tails outside the open ball of radius $\pi$ (this is due to the re-parametrization $\epsilon = \logmap(\Rmean\inv \Rrand)$, which restricts $\epsilon$ 
to $\|\epsilon\| < \pi$).
Approximating $\beta$ as a constant, the negative log-likelihood of a rotation $\R$, given a measurement $\Rrand$ distributed as in~\eqref{eq:gaussianSO3}, is:
\beq \label{eq:minusloglikeSO3-2}
  \calL(\R) = \frac{1}{2} \normsq{\logmap(\R\inv \Rrand)}{\Sigma} + \text{const} = \frac{1}{2} \normsq{\logmap(\Rrand\inv \R)}{\Sigma} + \text{const},
\eeq
which geometrically can be interpreted as the squared angle (geodesic distance in \SOthree) between $\Rrand$ and $\R$ weighted by the inverse uncertainty $\Sigma^{-1}$.
\hide{
The ``Gaussian'' distribution~\eqref{eq:gaussianSO3} only shares some of the desirable properties of a Gaussian defined in a vector space.
A Gaussian in $\Real^n$ solves the \emph{heat equation}, and it is the maximum-entropy distribution among the family of distributions with the same mean and covariance.
The distribution~\eqref{eq:gaussianSO3} is still a maximum-entropy distribution but only converges to the solution of the heat equation for small $|\Sigma|$~\cite[p.356]{Chirikjian12book}.
}
%Alternative distributions on \SOthree are the \emph{folded Normal distribution}~\cite{Lee05ijrr}, the \emph{von Mises-Fisher} (or \emph{Langevin}) distribution~\cite{Mardia99book}, and the \emph{wrapped Gaussian}~\cite{Fletcher03ipmi}.  

%%%%%%%%%%%%%%%%%%%%%%%%%%%%%%%%%%%%%%%%%%%%%%%%%%%%%%%%%%%%%%%%%%%%%%%%%%%%%%%%%%%%%%%%%%%%%%%%%%
\subsection{Gauss-Newton Method on Manifold} 
\label{sec:GNmanifold}

A standard Gauss-Newton method in Euclidean space works by repeatedly optimizing a quadratic approximation of the (generally non-convex) objective function. 
Solving the quadratic approximation reduces to solving a set of linear equations (\emph{normal equations}), and the solution of this local approximation is used to update the current estimate.
Here we recall how to extend this approach to (unconstrained) optimization problems whose variables belong to some manifold~$\calM$. 

Let us consider the following optimization problem:
\beq
  \label{eq:GNmanifold}
  \min_{x \in \calM} f(x),
\eeq
where the variable $x$ belongs to a manifold $\calM$; for the sake of simplicity we consider a single variable in~\eqref{eq:GNmanifold}, while the description easily generalizes to multiple variables. 

Contrarily to the Euclidean case, one cannot directly approximate~\eqref{eq:GNmanifold} as a quadratic function of $x$. This is due to two main reasons.
First, working directly on $x$ leads to an over-parametrization of the problem (e.g., we parametrize a rotation matrix with 9 elements, while a 3D rotation is completely defined by a vector in $\Real^3$) 
and this can make the \emph{normal equations} under-determined. 
Second, the solution of the resulting approximation does not belong to $\calM$ in general.

A standard approach for optimization on manifold~\cite{Absil07fcm,Smith94ams}, consists of defining a \emph{retraction} $\calR_x$, which is a bijective map between an element $\delta x$ of the tangent space (at $x$) and a neighborhood of $x\in\calM$.
Using the retraction, we can re-parametrize our problem as follows: 
\beq
  \label{eq:lifting}
  \min_{x \in \calM} f(x) \quad \Rightarrow \quad
  \min_{\delta x \in \Real^n} f( \calR_x(\delta x) ).
\eeq
The re-parametrization is usually called \emph{lifting}~\cite{Absil07fcm}.
Roughly speaking, we work in the tangent space defined at the current estimate, which locally behaves as an Euclidean space.
The use of the retraction allows framing the optimization problem over an Euclidean space 
of suitable dimension (e.g., $\delta x \in \Real^{3}$ when we work in \SOthree).
We can now apply standard optimization techniques to the problem on the right-hand side of~\eqref{eq:lifting}.
In the Gauss-Newton framework, we square the cost around the current estimate.
Then we solve the quadratic approximation to get a vector $\delta x^\star$ in the tangent space.
Finally, the current guess on the manifold is updated as
\beq
  \label{eq:update}
  \hat{x} \leftarrow \calR_{\hat{x}} (\delta x^\star).
\eeq

This ``lift-solve-retract'' scheme can be generalized to any trust-region method~\cite{Absil07fcm}.
Moreover, it provides a grounded and unifying generalization of the \emph{error state model}, commonly used in aerospace literature for filtering~\cite{Farrell08book} and recently adopted in robotics for optimization~\cite{Leutenegger13rss,Nerurkar14icra}.

We conclude this section by discussing the choice of the retraction $\calR_x$.
A possible retraction is the exponential map.
It is known that, computationally, this may not be the most convenient choice, see~\cite{Manton02tsp}.

In this work, we use the following retraction for \SOthree, % $\R$: 
\beq
  \label{eq:retractionR}
  \calR_\R(\rotvec) = \R \; \expmap(\rotvecpert), \qquad \rotvecpert \in \Real^3,
\eeq
and for \SEthree, we use the retraction at $\T \doteq (\R,\tran)$:
\beq
  \label{eq:retractionT}
  \calR_\T(\rotvecpert, \tranpert) = (\R \; \expmap(\rotvecpert), \ \tran + \R \; \tranpert), 
  \qquad \left[ \rotvecpert \ \ \tranpert \right] \in \Real^6,
\eeq
which explains why in Section~\ref{sec:manifolds} we only defined the exponential map for \SOthree: 
with this choice of retraction we never need to compute the exponential map for \SEthree.

%!TEX root = main.tex
\section{Maximum a Posteriori Visual-Inertial \texorpdfstring{\\}{} State Estimation}
\label{sec:ml_vi}

\begin{figure}[t]
  \centering
  \includegraphics[height=0.40\linewidth]{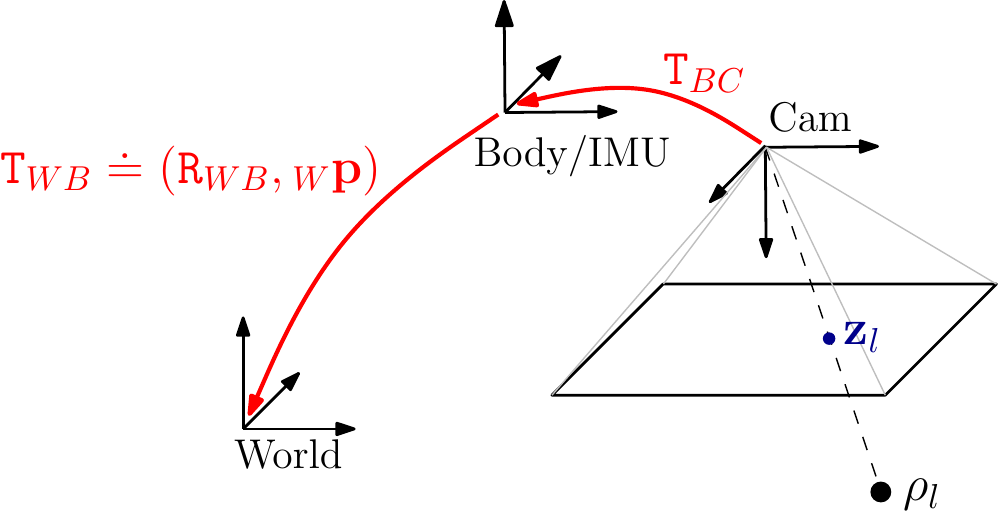}
  \caption{
  $\T_{\world\imu} \doteq (\R_{\world\imu},{}_{\world}\tran)$ is the pose of the 
  body frame $\Imu$ w.r.t. the world frame $\World$. We assume that the body frame coincides with the 
  IMU frame. $\T_{\imu\camera}$ is the pose of the camera in the body frame,  known from prior calibration.}
  \label{fig:frames}
  \vspace{-15px}
\end{figure}

We consider a \VIO problem in which we want to track the state of a \emph{sensing system} (e.g., a mobile robot, a UAV, or a hand-held device), 
equipped with an IMU and a monocular camera.
We assume that the IMU frame ``$\Imu$'' coincides with the body frame we want to track, and that the transformation between the camera and the IMU is fixed and known from prior calibration (Fig.~\ref{fig:frames}). 
Furthermore, we assume that a \emph{front-end} provides image measurements of 3D landmarks at unknown position. 
The front-end also selects a subset of images, called \emph{keyframes}~\cite{Strasdat10icra}, for which we want to compute a pose estimate. Section~\ref{sec:implementation} discusses implementation aspects, including the choice of the front-end in our experiments.
 
\subsection{The State}

The state of the system at time $i$ is described by the IMU orientation, position, velocity and biases:
\beq
  \mathbf{x}_i \doteq [\R_i,\tran_i,\vel_i,\bias_i].
\eeq
The pose $(\R_i, \tran_i)$ belongs to $\SEthree$, while velocities live in a vector space, i.e., $\vel_i \in \Real^3$.
IMU biases can be written as $\bias_i = [\bias^g_i \;\; \bias^a_i] \in \Real^6$, where $\bias^g_i, \bias^a_i \in \Real^3$ are the gyroscope and accelerometer bias, respectively. 

Let $\mathcal{K}_k$ denote the set of all keyframes up to time $k$.
In our approach we estimate the state of all keyframes:
\beq
	\States_k \doteq \{ \mathbf{x}_i  \}_{i \in \mathcal{K}_k}.
\eeq
In our implementation, we adopt a structureless approach (\emph{cf.}, Section \ref{sec:structureless}), hence the 3D landmarks are not part of the variables to be estimated.
However, the proposed approach generalizes in a straightforward manner to also estimating the landmarks and the camera intrinsic and extrinsic calibration parameters.

\subsection{The Measurements}
The input to our estimation problem are the measurements from the camera and the IMU.
We denote with $\meascam_i$ the image measurements at keyframe $i$. 
At time $i$, the camera can observe multiple landmarks $l$, hence $\meascam_i$ contains multiple image measurements $\mathbf{z}_{il}$. 
With slight abuse of notation we write $l \in \meascam_i$ when a landmark $l$ is seen at time $i$.

We denote with $\measimu_\subimu$  the set of IMU measurements acquired between two consecutive keyframes $i$ and  $j$. 
Depending on the IMU measurement rate and the frequency of selected keyframes, each set $\measimu_\subimu$ can contain from a small number to hundreds of IMU measurements.
The set of measurements collected up to time $k$ is 
\beq
	\mathcal{Z}_k \doteq \{ \meascam_i, \measimu_\subimu \}_{\indmeas \in \mathcal{K}_k}.
\eeq

\subsection{Factor Graphs and MAP Estimation} 

%A \emph{factor graph} encodes 
The posterior probability of the variables $\States_k$, given the available visual and inertial measurements $\mathcal{Z}_k$ and priors $p(\States_0)$ is:
\begin{align}
  \label{eq:mapDistribution}
  p(\States_k |& \mathcal{Z}_k)
  \propto \ p(\States_0) p(\mathcal{Z}_k | \States_k)
  \overset{(a)}{=} 
  p(\States_0) \!\!\! \prod_{\indmeas \in \mathcal{K}_k} \!\!\! p(\meascam_i, \measimu_\subimu | \States_k)  \nonumber  \\
  %\!\!\!\!&\overset{(a)}{=}&\!\!\!\! p(\States_0) \!\!\! \prod_{\indmeas \in \mathcal{K}_k} \!\!\! p(\measimu_\subimu | \States_k) \!\!
  %\; \prod_{i \in \mathcal{K}_k} \prod_{l \in \meascam_i} p(\mathbf{z}_{il} | \States_k)   
  \overset{(b)}{=} & \
  p(\States_0) \!\!\!\! \prod_{\indmeas \in \mathcal{K}_k} \!\!\! \! 
  p(\measimu_\subimu | \mathbf{x}_i, \mathbf{x}_j) 
  \; \prod_{i \in \mathcal{K}_k} \prod_{l \in \meascam_i} p(\mathbf{z}_{il} | \mathbf{x}_i). 
\end{align}
The factorizations (a) and (b) follow from a standard independence assumption among the measurements. Furthermore, the Markovian property is applied in (b) (e.g., an image measurement at time $i$ only depends on the state at time $i$).

\begin{figure}[t]
  \centering
  \hbox{
  \includegraphics[height=0.25\linewidth]{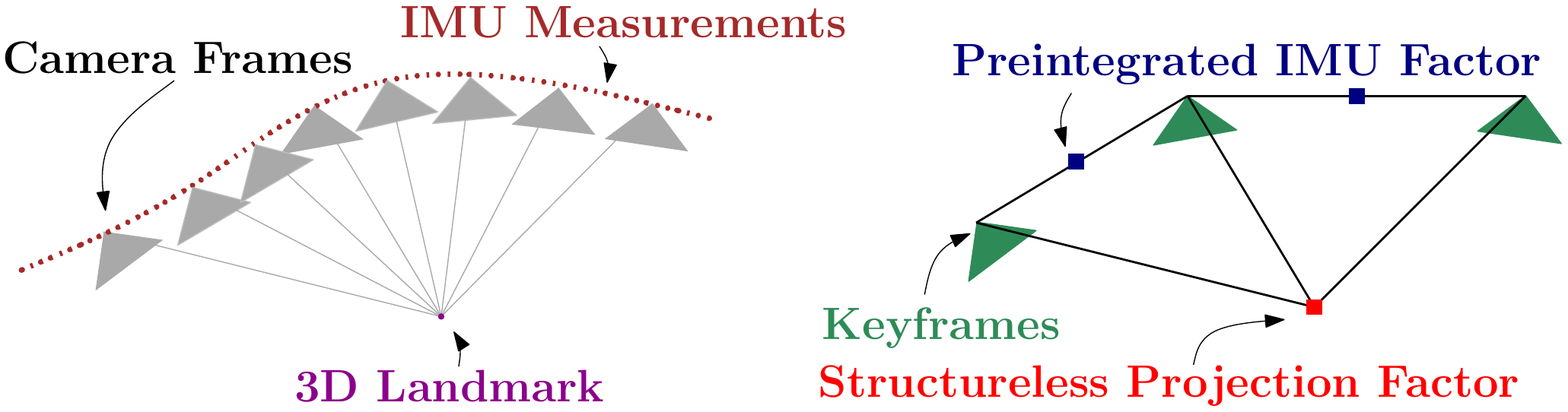}
  }
  \caption{Left: visual and inertial measurements in \VIO. Right: factor graph 
  in which several IMU measurements are summarized in a single preintegrated 
  IMU factor and a structureless vision factor constraints keyframes observing the same landmark.}
  \label{fig:factor_graph}
  \vspace{-10px}
\end{figure}

As the measurements $\mathcal{Z}_k$ are known, we are free to eliminate them as variables and consider them as parameters of the joint probability factors over the actual unknowns. 
This naturally leads to the well known factor graph representation, a class of bipartite graphical models that can be used to represent such factored densities \cite{Kschischang01it,Dellaert05tr}.
%The terms in the factorization~\eqref{eq:mapDistribution} are called \emph{factors}.
A schematic representation of the connectivity of the factor graph underlying the \VIO problem is given in \Figure~\ref{fig:factor_graph} (the connectivity of the structureless vision factors will be clarified in Section~\ref{sec:structureless}).
The factor graph is composed of nodes for unknowns and nodes for the probability factors defined on them, and the graph structure expresses which unknowns are involved in each factor.

The \MAP estimate $\States_k^\star$ corresponds to the maximum of~\eqref{eq:mapDistribution}, or equivalently, the minimum of the negative log-posterior.
Under the assumption of zero-mean Gaussian noise, the negative log-posterior can be written as a sum of squared residual errors:
\begin{align}
\label{eq:cost} 
  & \States_k^\star \doteq  \arg\min_{\States_k} \; -\log_e  \;p(\States_k | \mathcal{Z}_k) \\
  &= \arg\min_{\States_k} \; 
     \|\residual_{0}\|^2_{\covprior_0} \!\!\!
     + \sum_{\indmeas \in \mathcal{K}_k} \| \residual_{\measimu_\subimu} \|^2_{\covimu_\subimu}
     + \sum_{i \in \mathcal{K}_k}\sum_{l \in \meascam_i} \| \residual_{\meascam_{il}} \|^2_{\covcam} \nonumber
\end{align}
where $\residual_{0}$, $\residual_{\measimu_\subimu}$, $\residual_{\meascam_{il}}$ are the residual errors associated to the measurements, and 
$\covprior_0$, $\covimu_\subimu$, and $\covcam$ are the corresponding covariance matrices.  
Roughly speaking, the residual error is a function of $\States_k$ that quantifies the mismatch between a measured quantity and the predicted value of this quantity given the state  $\States_k$ and the priors. 
The goal of the following sections is to provide expressions for the residual errors and the covariances.

%!TEX root = main.tex
\section{IMU Model and Motion Integration}
\label{sec:imu_model}

An IMU commonly includes a 3-axis accelerometer and a 3-axis gyroscope and allows measuring the rotation rate and the acceleration of the sensor with respect to an inertial frame. 
The measurements, namely ${}_\imu\tilde\acc(t)$, and ${}_\imu\tilde\rotvel_{\world\imu}(t)$, are affected by additive white noise $\noise$ and a slowly varying sensor bias $\bias$:
\begin{align}
  {}_\imu\tilde\rotvel_{\world\imu}(t)
    &= {}_\imu\rotvel_{\world\imu}(t) + \bias^g(t) + \noise^{g}(t) \label{eq:omegaModel}\\
  {}_\imu\tilde\acc(t)
    &= \R_{\world\imu}^\transpose(t)\left({}_\world\acc(t) - {}_\world\gravity \right) + \bias^a(t) + \noise^{a}(t), \label{eq:accModel}
\end{align}
In our notation, the prefix $\Imu$ denotes that the corresponding quantity is expressed in the frame $\Imu$ (\emph{c.f.,} Fig. \ref{fig:frames}).
The pose of the IMU is described by the transformation $\{\R_{\world\imu}, {}_\world\tran\}$, which maps a point from sensor frame $\Imu$ to $\World$.
The vector ${}_\imu\rotvel_{\world\imu}(t) \!\in\! \Real^3$ is the  instantaneous angular velocity of $\Imu$ relative to $\World$ expressed in coordinate frame $\Imu$, while ${}_\world\acc(t) \!\in\! \Real^3$ is the acceleration of the sensor; ${}_\world\gravity$ is the gravity vector in world coordinates. We neglect effects due to earth's rotation, which amounts 
to assuming that $\World$ is an inertial frame.

The goal now is to infer the motion of the system from IMU measurements.
For this purpose we introduce the following kinematic model~\cite{Murray94book,Farrell08book}:
\beq\label{eq:kinematic_model}
  \dot\R_{\world\imu} = \R_{\world\imu} \ {}_\imu\rotvel_{\world\imu}^\wedge, \qquad
  {}_\world\dot\vel = {}_\world\acc, \qquad
  {}_\world\dot\tran = {}_\world\vel,
\eeq
which describes the evolution of the pose and velocity of $\Imu$.

The state at time $t+\Delta t$ is obtained by integrating \eq~\eqref{eq:kinematic_model}:
\begin{align*}
  \R_{\world\imu}(t+\Delta t)
    &= \R_{\world\imu}(t) \; \expmap\left(\int_t^{t+\Delta t} \!\!\!\! {}_\imu\rotvel_{\world\imu}(\tau) d\tau \right)\\
  {}_\world\vel(t+\Delta t)
    &= {}_\world\vel(t) + \int_t^{t+\Delta t} \!\!\!\! {}_\world\acc(\tau) d\tau \nonumber\\
  {}_\world\tran(t+\Delta t)
    &= {}_\world\tran(t) + \int_t^{t+\Delta t} \!\!\!\! {}_\world\vel(\tau) d\tau + \iint_t^{t+\Delta t} {}_\world\acc(\tau) d\tau^2.
\end{align*}
Assuming that ${}_\world\acc$ and ${}_\imu\rotvel_{\world\imu}$ remain constant in the time interval $[t,t+\Delta t]$, we can write:
\begin{align}
\label{eq:euler}
  \R_{\world\imu}(t+\Delta t) 
    &= \R_{\world\imu}(t) \; \expmap\left({}_\imu\rotvel_{\world\imu}(t) \Delta t\right) \nonumber\\
  {}_\world\vel(t+\Delta t)
    &= {}_\world\vel(t) + {}_\world\acc(t) \Delta t \nonumber\\
  {}_\world\tran(t+\Delta t)
    &= {}_\world\tran(t) + {}_\world\vel(t) \Delta t + \frac{1}{2}{}_\world\acc(t) \Delta t^2.
\end{align}
Using \eqs~\eqref{eq:omegaModel}--\eqref{eq:accModel}, we can write ${}_\world\acc$ and ${}_\imu\rotvel_{\world\imu}$ as a function of the IMU measurements, hence~\eqref{eq:euler} becomes
\begin{align}\label{eq:discreteStatePropagation}
  \R(t+\Delta t) 
    &=\ \R(t) \ \expmap\left(\left(\tilde\rotvel(t) - \bias^g(t)-\noise^{gd}(t) \right) \Delta t\right) \nonumber\\
  \vel(t+\Delta t)
    &=\ \vel(t) + \gravity\Delta t
    + \R(t) \left( \accshort \right) \Delta t \nonumber\\
  \tran(t+\Delta t)
    &=\ \tran(t) + \vel(t) \Delta t + \frac{1}{2}\gravity\Delta t^2 \nonumber
    \\&
    + \frac{1}{2}\R(t) \left( \accshort \right) \Delta t^2,
\end{align}
where we dropped the coordinate frame subscripts for readability (the notation should be unambiguous from now on). 
This numeric integration of the velocity and position assumes a constant orientation $\R(t)$ for the time of integration between two measurements, which is not an exact solution of the differential equation \eqref{eq:kinematic_model} for measurements with non-zero rotation rate. In practice, 
the use of a high-rate IMU mitigates the effects of this approximation.
We adopt the integration scheme \eqref{eq:discreteStatePropagation} as it is simple and amenable for modeling and uncertainty propagation.
While we show that this integration scheme performs very well in practice, we remark that for slower IMU measurement rates one may consider using
 higher-order numerical integration methods \cite{Crouch93,MuntheKaas99,Park05tro,Andrle13jgcd}.

The covariance of the discrete-time noise $\noise^{gd}$ is a function of the sampling rate and relates to the continuous-time spectral noise $\noise^{g}$ via
$\text{Cov}(\noise^{gd}(t)) = \frac{1}{\Delta t}\text{Cov}(\noise^{g}(t))$.
The same relation holds for $\noise^{ad}$ (\emph{cf.}, \cite[Appendix]{Crassidis06taes}).
%!TEX root = main.tex

\section{IMU Preintegration on Manifold}
\label{sec:preintegration}

\begin{figure}[t]
  \centering
  \includegraphics[width=0.9\linewidth]{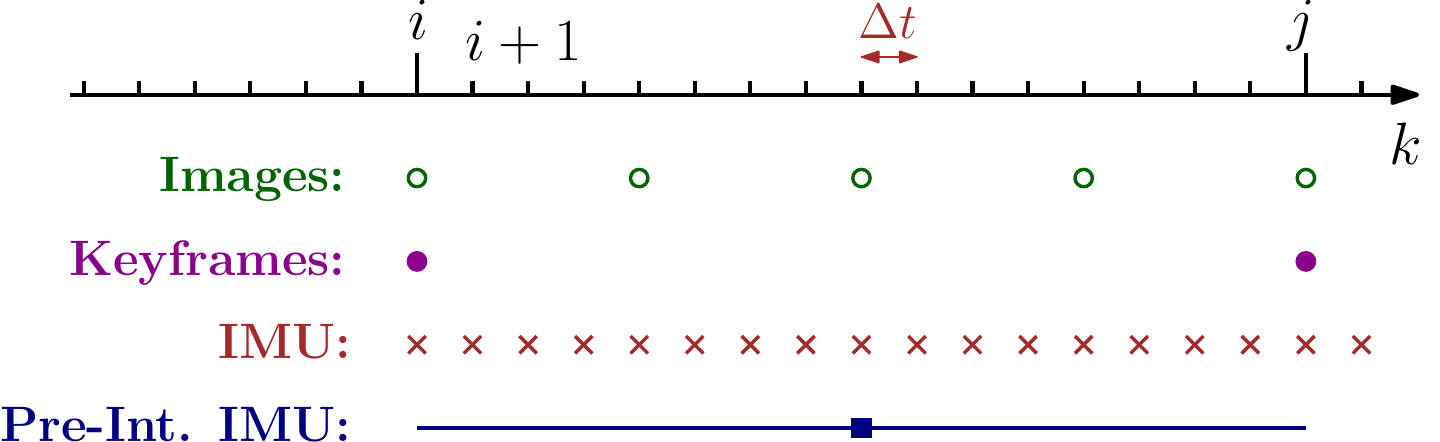}
  \caption{Different rates for IMU and camera. \vspace{-10pt}}
  \label{fig:preintegration_notation}
\end{figure}

While \eq~\eqref{eq:discreteStatePropagation} could be readily seen as a probabilistic constraint in a factor graph, it would require to include states in the factor graph at high rate. 
Intuitively, \eq~\eqref{eq:discreteStatePropagation} relates states at time $t$ and $t+\Delta t$, where $\Delta t$ 
is the sampling period of the IMU, hence we would have to add new states in the estimation at every new IMU measurement \cite{Indelman12fusion}.

Here we show that all measurements between two keyframes at times $k=i$ and $k=j$ (see \Figure~\ref{fig:preintegration_notation}) can be summarized in a single compound measurement, named \emph{preintegrated IMU measurement}, which constrains the motion between consecutive keyframes.
This concept was first proposed in \cite{Lupton12tro} using Euler angles and we extend it, by developing a suitable theory for preintegration on the manifold $\SOthree$. 

We assume that the IMU is synchronized with the camera and provides measurements at discrete times~$k$ (\emph{cf.}, \Figure~\ref{fig:preintegration_notation}).\footnote{We calibrate the IMU-camera delay using the \emph{Kalibr} toolbox \cite{Furgale13iros}. An alternative is to add the delay as a state in the estimation process \cite{Li14ijrra}.}
Iterating the IMU integration~\eqref{eq:discreteStatePropagation} for all $\Delta t$ intervals between two consecutive keyframes at times $k=i$ and $k=j$ (\emph{c.f.}, Fig. \ref{fig:preintegration_notation}), we find:
\begin{align}\label{eq:preintegrationGlobalFrame}
	\R_j
    =&\ \R_i
    \prod_{k=i}^{j-1}\expmap\left(\left(\tilde\rotvel_k - \bias^g_k-\noise^{gd}_k \right) \Delta t\right),  \nonumber\\
  \vel_j
    =&\ \!  \vel_i \!  + \gravity \Delta t_{ij}
    + \sum_{k=i}^{j-1} \R_k \Big( \tilde\acc_k-\bias^a_k-\noise^{ad}_k \Big)\Delta t \\
  \tran_j
    =&\ \! \tran_i \!
    + \sum_{k=i}^{j-1}
      \Big[ \vel_k \Delta t
    + \frac{1}{2}\gravity \Delta t^2
    + \frac{1}{2}\R_k \Big( \tilde\acc_k \!-\! \bias^a_k \!-\! \noise^{ad}_k \Big)\Delta t^2 \Big] \nonumber
\end{align}
where we introduced the shorthands $\Delta t_{ij}\doteq\sum_{k=i}^{j-1}\Delta t$ and $(\cdot)_i\doteq(\cdot)(t_i)$ for readability.
While \eq~\eqref{eq:preintegrationGlobalFrame} already provides an estimate of the motion between time $t_i$ and $t_j$, it has the drawback that the integration in~\eqref{eq:preintegrationGlobalFrame} has to be repeated whenever the linearization point at time $t_i$ changes~\cite{Leutenegger15ijrr} (intuitively, a change in the rotation $\R_i$ implies a change in all future rotations $\R_k$, $k=i,\ldots,j-1$, and makes necessary to re-evaluate summations and products in~\eqref{eq:preintegrationGlobalFrame}).

We want to avoid to recompute the above integration whenever the linearization point at time $t_i$ changes. 
Therefore, we follow \cite{Lupton12tro} and \emph{define} the following relative motion increments that are independent of the pose and velocity at $t_i$:
\begin{align}
\label{eq:preintVar}
  \Delta\R_{ij}
    &\doteq \R_i^\transpose \R_j
    = \prod_{k=i}^{j-1}\expmap\left(\left(\tilde\rotvel_k - \bias^g_k-\noise^{gd}_k \right) \Delta t\right) \nonumber\\
  \Delta\vel_{ij} 
    &\doteq \R_i^\transpose \left(\vel_j \!-\! \vel_i \!-\! \gravity\Delta t_{ij}\right) 
    \!=\! \sum_{k=i}^{j-1} \Delta\R_{ik} \!\left( \tilde\acc_k \!-\! \bias^a_k \!-\! \noise^{ad}_k\right)\!\Delta t \nonumber\\
  \Delta\tran_{ij} 
    &\doteq \textstyle \R_i^\transpose \left( \tran_j-\tran_i - \vel_i\Delta t_{ij} - \frac{1}{2} \gravity \Delta t_{ij}^2\right) \nonumber\\
    &= \sum_{k=i}^{j-1} 
    \left[ \Delta\vel_{ik}\Delta t
     + \frac{1}{2}\Delta\R_{ik}\left(\tilde\acc_k \!-\! \bias^a_k \!-\! \noise^{ad}_k\right)\Delta t^2 \right]
\end{align}
where $\Delta\R_{ik} \doteq \R_i^\transpose \R_k$ and $\Delta\vel_{ik} \doteq \R_i^\transpose \left(\vel_k \!-\! \vel_i \!-\! \gravity\Delta t_{ik}\right)$.
We highlight that, in contrast to the ``delta'' rotation $\Delta\R_{ij}$, neither 
$\Delta\vel_{ij}$ nor  $\Delta\tran_{ij}$ correspond to the true \emph{physical} change in velocity and position but are defined in a way that make the right-hand side of~\eqref{eq:preintVar} independent from the state at time $i$ as well as gravitational effects.
Indeed, we will be able to compute the right-hand side of~\eqref{eq:preintVar} directly from the inertial measurements between the two keyframes.

Unfortunately, summations and products in~\eqref{eq:preintVar} are still function of the bias estimate.
We tackle this problem in two steps.
In Section~\ref{sec:preintmeas}, we assume $\bias_i$ is known; then, in Section~\ref{sec:biasUpdate} we show how to avoid repeating the integration when the bias estimate changes.

In the rest of the paper, we assume that the bias remains constant between two keyframes:
\begin{equation}
  \bias^g_i = \bias^g_{i+1} = \ldots = \bias^g_{j-1}, \quad 
  \bias^a_i = \bias^a_{i+1} = \ldots = \bias^a_{j-1}.
\end{equation}

%%%%%%%%%%%%%%%%%%%%%%%%%%%%%%%%%%%%%%%%%%%%%%%%%%%%%%%%%%%%%%%%%%%%%%%
\subsection{Preintegrated IMU Measurements}
\label{sec:preintmeas}

Equation~\eqref{eq:preintVar} relates the states of keyframes $i$ and $j$ (left-hand side) to the measurements (right-hand side).
In this sense, it can be already understood as a measurement model.
Unfortunately, it has a fairly intricate dependence on the measurement noise and this complicates a direct application of MAP estimation; intuitively, the MAP estimator requires to clearly define the densities (and their log-likelihood) of the measurements.
In this section we manipulate~\eqref{eq:preintVar} so to make easier the derivation of the measurement log-likelihood.   
In practice, we isolate the noise terms of the individual inertial measurements in~\eqref{eq:preintVar}.
As discussed above, across this section assume that the bias at time $t_i$ is known. 

Let us start with the rotation increment $\Delta\R_{ij}$ in~\eqref{eq:preintVar}.
We use the first-order approximation~\eqref{eq:expExpansion} (rotation noise is ``small'') and rearrange the terms, by ``moving'' the noise to the end, using the relation~\eqref{eq:adjointProperty2}:
\begin{align}
\Delta\R_{ij}
    &\stackrel{\text{eq.}\eqref{eq:expExpansion}}{\simeq} \prod_{k=i}^{j-1} 
    \left[ 
    \expmap\left(\left(\tilde\rotvel_k - \bias^g_i \right) \Delta t\right) 
    \expmap\left( - \DExpk  \; \noise^{gd}_k \;  \Delta t \right)
    \right]
    \nonumber\\
    &\!\stackrel{\text{eq.}\eqref{eq:adjointProperty2}}{=}
    \preintRmeas_{ij}  \prod_{k=i}^{j-1} \expmap\left( - \preintRmeas_{k+1j}^\transpose \; \DExpk \; \noise^{gd}_k \; \Delta t\right) 
    \nonumber \\
    & \ \doteq \preintRmeas_{ij} \expmap\left( - \rotvecpert_{ij} \right)\label{eq:preintVarR} 
\end{align}
with $\DExpk \doteq \DExpk((\tilde\rotvel_k - \bias^g_i)\Delta t)$.
In the last line of~\eqref{eq:preintVarR}, we defined the \emph{preintegrated rotation measurement} $\preintRmeas_{ij} \doteq \prod_{k=i}^{j-1} \expmap\left(\left( \tilde\rotvel_k - \bias^g_i \right) \Delta t\right)$, and its noise $\rotvecpert_{ij}$, which will be further analysed in the next section.

Substituting~\eqref{eq:preintVarR} back into the expression of $\Delta\vel_{ij}$ in~\eqref{eq:preintVar}, using the first-order approximation~\eqref{eq:expFirstOrder} for $\expmap\left( - \rotvecpert_{ij} \right)$, and dropping higher-order noise terms, we obtain:
\begin{align}\label{eq:preintVarV}
 \Delta\vel_{ij} 
    &\stackrel{\text{eq.}\eqref{eq:expFirstOrder}}{\simeq} 
    \sum_{k=i}^{j-1} \preintRmeas_{ik}(\eye - \rotvecpert_{ik}^\wedge) 
    \left( \tilde\acc_k \!-\! \bias^a_i \right)\Delta t 
    \!-\!  \preintRmeas_{ik} \noise^{ad}_k \Delta t   \nonumber \\
 	& \overset{\text{eq.} \eqref{eq:skewProperty}}{=}
    \preintVmeas_{ij}  \!+\! \sum_{k=i}^{j-1} 
	\left[
	\preintRmeas_{ik} \left( \tilde\acc_k \!-\! \bias^a_i \right)^\wedge \rotvecpert_{ik} \Delta t 
	- \preintRmeas_{ik} \noise^{ad}_k \Delta t
	\right]
	  \nonumber \\
	&\;\;\doteq
    \preintVmeas_{ij} - \velpert_{ij}
\end{align}
where we defined the \emph{preintegrated velocity measurement} $\preintVmeas_{ij}\doteq  \sum_{k=i}^{j-1} \! \preintRmeas_{ik}  \! \left( \tilde\acc_k \!-\! \bias^a_i \right)  \Delta t$ and its noise~$\velpert_{ij}$.

Similarly, substituting~\eqref{eq:preintVarR} and~\eqref{eq:preintVarV} in the expression of $\Delta\tran_{ij}$ in~\eqref{eq:preintVar}, and using the first-order approximation~\eqref{eq:expFirstOrder}, we obtain:
\begin{small}
\begin{align}\label{eq:preintVarP}
\!\!\!\!\!\!\!\Delta\tran_{ij} 
    & \!\! \stackrel{\text{eq.}\eqref{eq:expFirstOrder}}{\simeq}\!
    \sum_{k=i}^{j-1} 
      \Big[
        (\preintVmeas_{ik} \!-\! \velpert_{ik})\Delta t
        \!+\! \frac{1}{2} \preintRmeas_{ik}(\eye \!-\! \rotvecpert_{ik}^\wedge) \left(\tilde\acc_k \!-\! \bias^a_i\right)\Delta t^2 
    \nonumber \\
    & \quad \quad \quad \quad  
       \!-\! \frac{1}{2} \preintRmeas_{ik} \noise^{ad}_k \Delta t^2 \Big]
    \nonumber \\
    %%%%%%%%%%%%
    & \!\! \overset{\text{eq.} \eqref{eq:skewProperty}}{\!\!=\!\!}
    \preintPmeas_{ij} \!+\! 
    \sum_{k=i}^{j-1} 
    \Big[
      - \velpert_{ik}\Delta t
      + \frac{1}{2} \preintRmeas_{ik}  \left(\tilde\acc_k \!-\! \bias^a_i \right)^\wedge \rotvecpert_{ik} \Delta t^2
      \nonumber \\
    & \quad \quad \quad \quad \quad \quad \quad  
      - \frac{1}{2} \preintRmeas_{ik} \noise^{ad}_k \Delta t^2 
    \Big]
    \nonumber \\
    & \doteq 
    \preintPmeas_{ij} - \tranpert_{ij}, 
\end{align}
\end{small}
where we defined the \emph{preintegrated position measurement} $\preintPmeas_{ij}$ and its noise~$\tranpert_{ij}$.

Substituting the expressions~\eqref{eq:preintVarR}, \eqref{eq:preintVarV}, \eqref{eq:preintVarP} back in the original definition of $\Delta\R_{ij},\Delta\vel_{ij},\Delta\tran_{ij}$ in~\eqref{eq:preintVar}, we finally get our \emph{preintegrated measurement model}
(remember $\expmap\left( -\rotvecpert_{ij} \right)^\transpose = \expmap\left( \rotvecpert_{ij} \right)$):
\begin{align}\label{eq:preintMeasModel}
  \preintRmeas_{ij}
    &= \R_i^\transpose \R_j \expmap\left( \rotvecpert_{ij} \right) \nonumber\\
  \preintVmeas_{ij} 
    &= \R_i^\transpose \left(\vel_j \!-\! \vel_i \!-\! \gravity\Delta t_{ij}\right) + \velpert_{ij} \nonumber\\
  \preintPmeas_{ij} 
    &= \R_i^\transpose \left( \tran_j-\tran_i - \vel_i\Delta t_{ij} - \frac{1}{2}\gravity \Delta t_{ij}^2\right) 
    + \tranpert_{ij}
\end{align} 
where our compound measurements are written as a function of the (to-be-estimated) state ``plus'' a random noise, described by the random vector $[\rotvecpert_{ij}^\transpose, \velpert_{ij}^\transpose, \tranpert_{ij}^\transpose]^\transpose$. 

To wrap-up the discussion in this section, we manipulated the measurement model~\eqref{eq:preintVar} 
and rewrote it as~\eqref{eq:preintMeasModel}. The advantage of~\Equation~\eqref{eq:preintMeasModel} 
is that, for a suitable distribution of the noise, it makes the definition of the log-likelihood straightforward. 
For instance the (negative) log-likelihood of measurements with zero-mean additive Gaussian noise (last two lines in~\eqref{eq:preintMeasModel}) is a quadratic function. %, while we reported the log-likelihood of 
Similarly, if $\rotvecpert_{ij}$ is a zero-mean Gaussian noise, we compute the (negative) log-likelihood 
 associated with $\preintRmeas_{ij}$.
The nature of the noise terms is discussed in the following section.

%%%%%%%%%%%%%%%%%%%%%%%%%%%%%%%%%%%%%%%%%%%%%%%%%%%%%%%%%%%%%%%%%%%%%%%
\subsection{Noise Propagation}
\label{sec:noisePropagation}

In this section we derive the statistics of the noise vector $[\rotvecpert_{ij}^\transpose, \velpert_{ij}^\transpose, \tranpert_{ij}^\transpose]^\transpose$. 
While we already observed that it is convenient to approximate the noise vector to be 
zero-mean Normally distributed, it is of paramount importance to accurately model the noise covariance. 
Indeed, the noise covariance has a strong influence on 
the MAP estimator (the inverse noise covariance is used to weight the terms in the optimization \eqref{eq:cost}).
In this section, we therefore provide a derivation of the covariance $\covimu_{ij}$ of the preintegrated measurements:
\beq 
\label{eq:preintMeasNoise}
\noise^{\Delta}_{ij} \doteq [\rotvecpert_{ij}^\transpose, \velpert_{ij}^\transpose, \tranpert_{ij}^\transpose]^\transpose 
\sim \calN(\Zero_{9 \times 1}, \covimu_{ij}). 
\eeq

We first consider the preintegrated rotation noise $\rotvecpert_{ij}$.
%Let us start by writing explicitly what the preintegrated rotation noise is. 
Recall from \eqref{eq:preintVarR} that
\beq 
  \textstyle
  \expmap\left( - \rotvecpert_{ij} \right) \doteq \prod_{k=i}^{j-1} 
  \expmap\left( - \preintRmeas_{k+1j}^\transpose \DExpk \; \noise^{gd}_k \; \Delta t\right).
\eeq
Taking the $\logmap$ on both sides and changing signs, we get:
\beq  
  \textstyle
  \rotvecpert_{ij} 
  = - \logmap\left(\prod_{k=i}^{j-1} 
      \expmap\left( - \preintRmeas_{k+1j}^\transpose \DExpk \; \noise^{gd}_k \; \Delta t\right)
      \right).
\eeq
Repeated application of the first-order approximation~\eqref{eq:logExpansion} (recall that $\noise^{gd}_k$ as well as $\rotvecpert_{ij}$ are small rotation noises, hence the right Jacobians are close to the identity) produces:
\beq 
  \textstyle
  \label{eq:firstOrderRotvecpert}
  \rotvecpert_{ij} \simeq
  \sum_{k=i}^{j-1} \preintRmeas_{k+1j}^\transpose \; \DExpk \; \noise^{gd}_k \; \Delta t
\eeq
Up to first order, the noise $\rotvecpert_{ij}$ is zero-mean and Gaussian, as it is a linear combination of zero-mean noise terms $\noise^{gd}_k$. % (the gyroscope noise). 
This is desirable, since it brings the rotation measurement model~\eqref{eq:preintMeasModel} exactly in the form~\eqref{eq:rightPerturbation}.

Dealing with the noise terms $\velpert_{ij}$ and $\tranpert_{ij}$ is now easy: these are linear combinations of the acceleration noise $\noise^{ad}_k$ and the preintegrated rotation noise $\rotvecpert_{ij}$, hence they are also zero-mean and Gaussian.
Simple manipulation leads to:
\begin{small}
\begin{align}\label{eq:noiseRVP}
\velpert_{ij}  
    &\simeq \!
    \sum_{k=i}^{j-1} 
    \left[ -
  \preintRmeas_{ik} \left( \tilde\acc_k \!-\! \bias^a_i \right)^\wedge \! \rotvecpert_{ik} \Delta t 
  + \preintRmeas_{ik} \noise^{ad}_k \Delta t
  \right]  \\
\tranpert_{ij} 
  &\simeq \!
  \sum_{k=i}^{j-1} 
    \Big[
      \velpert_{ik}\Delta t
      \!-\! \frac{1}{2} \preintRmeas_{ik}  \left(\tilde\acc_k \!-\! \bias^a_i \right)^\wedge \! \rotvecpert_{ik} \Delta t^2
      \!+\! \frac{1}{2} \preintRmeas_{ik} \noise^{ad}_k \Delta t^2 \Big]
      \nonumber
\end{align}
\end{small}
where the relations are valid up to the first order. 

Eqs.~\eqref{eq:firstOrderRotvecpert}-\eqref{eq:noiseRVP} express the preintegrated noise $\noise^{\Delta}_{ij}$ 
as a linear function of the IMU measurement noise $\noise^{d}_k \doteq [\noise^{gd}_k, \noise^{ad}_k]$, $k=1,\ldots,j-1$. 
Therefore, from the knowledge of the covariance of $\noise^{d}_k$ (given in the IMU specifications), we can compute the 
covariance of $\noise^{\Delta}_{ij}$, namely $\covimu_{ij}$, by a simple linear propagation.

In Appendix~\ref{sec:appendix_noisePropagation}, we provide a more clever way to compute  
$\covimu_{ij}$. In particular, we show that $\covimu_{ij}$ can be conveniently computed in iterative form: 
as a new IMU measurement arrive we only update $\covimu_{ij}$, rather 
than recomputing it from scratch. The iterative computation leads to 
simpler expressions and is more amenable for online inference.

%%%%%%%%%%%%%%%%%%%%%%%%%%%%%%%%%%%%%%%%%%%%%%%%%%%%%%%%%%%%%%%%%%%%%%%
\subsection{Incorporating Bias Updates}
\label{sec:biasUpdate}

In the previous section, we assumed that the bias $\{ \biasFixed^{a}_i, \biasFixed^{g}_i \}$ that is used during preintegration between $k=i$ and $k=j$ is correct and does not change.
However, more likely, the bias estimate changes by a small amount $\biaspert$ during optimization.
One solution would be to recompute the delta measurements when the bias changes; however, that is computationally expensive.
Instead, given a bias update $\bias \leftarrow \biasFixed+\biaspert$, we can update the delta measurements using a first-order expansion:
\begin{align}
\label{eq:biasUpdate}
  \preintRmeas_{ij}(\bias^g_i) 
    &\textstyle \simeq 
    \preintRmeas_{ij}(\biasFixed^g_i)\;
    \expmap\Big(\frac{\partial \pRot}{\partial\bias^g} \biaspert^g\Big) \\
  \preintVmeas_{ij}(\bias^g_i, \bias^a_i)
    &\textstyle \simeq 
    \preintVmeas_{ij}(\biasFixed^g_i, \biasFixed^a_i)
    + \frac{\partial \pVel}{\partial\bias^g} \biaspert^g_i
    + \frac{\partial \pVel}{\partial\bias^a} \biaspert^a_i \nonumber\\
  \preintPmeas_{ij}(\bias^g_i, \bias^a_i)
    &\textstyle \simeq 
    \preintPmeas_{ij}(\biasFixed^g_i, \biasFixed^a_i)
    + \frac{\partial \pTran}{\partial\bias^g} \biaspert^g_i
    + \frac{\partial \pTran}{\partial\bias^a} \biaspert^a_i \nonumber
\end{align}
This is similar to the bias correction in \cite{Lupton12tro} but operates directly on $\SOthree$.
The Jacobians $\{ \frac{\partial \pRot}{\partial\bias^g}, \frac{\partial \pVel}{\partial\bias^g}, \ldots  \}$
(computed at $\biasFixed_i$, the bias estimate at integration time) describe how the measurements change due to a change in the bias estimate. 
The Jacobians remain constant and can be precomputed during the preintegration.
The derivation of the Jacobians is very similar to the one we used in Section~\ref{sec:preintmeas} to express the measurements as a large value \emph{plus} a small perturbation and is given in Appendix~\ref{app:biasUpdate}.

%%%%%%%%%%%%%%%%%%%%%%%%%%%%%%%%%%%%%%%%%%%%%%%%%%%%%%%%%%%%%%%%%%%%%%%
\subsection{Preintegrated IMU Factors}

Given the preintegrated measurement model in~\eqref{eq:preintMeasModel} and since measurement noise is zero-mean and Gaussian (with covariance $\covimu_{ij}$) up to first order~\eqref{eq:preintMeasNoise}, it is now easy to write the residual errors
$
  \residual_{\measimu_\subimu}
  \doteq [\residual_{\Delta\R_{ij}}^\transpose, \residual_{\Delta\vel_{ij}}^\transpose, 
  \residual_{\Delta\tran_{ij}}^\transpose]^\transpose \in \Real^9
$,
where
\begin{align}
  \residual_{\Delta\R_{ij}}
    \doteq&\ \textstyle \logmap\left( 
    \left( \preintRmeas_{ij}(\biasFixed^g_i) \expmap\left(\frac{\partial\pRot}{\partial\bias^g} \biaspert^g\right)  \right)^\transpose 
     \R_i^\transpose \R_j  \right) \nonumber \\
  \residual_{\Delta\vel_{ij}} 
    \doteq&\ \textstyle \R_i^\transpose \left(\vel_j-\vel_i - \gravity\Delta t_{ij}\right)\nonumber\\ 
     -& \textstyle \left[ \preintVmeas_{ij}(\biasFixed^g_i, \biasFixed^a_i) 
                     + \frac{\partial\pVel}{\partial\bias^g} \biaspert^g
                     + \frac{\partial\pVel}{\partial\bias^a} \biaspert^a  \right] \nonumber\\
  \label{eq:residuals}
  \residual_{\Delta\tran_{ij}}
    \doteq&\ \textstyle  \R_i^\transpose \big( \tran_j-\tran_i - \vel_i\Delta t_{ij}
    - \frac{1}{2}\gravity \Delta t_{ij}^2\big) \nonumber \\
    -& \textstyle \left[ \preintPmeas_{ij}(\biasFixed^g_i, \biasFixed^a_i) 
           + \frac{\partial\pTran}{\partial\bias^g} \biaspert^g
           + \frac{\partial\pTran}{\partial\bias_a} \biaspert^a \right ],
\end{align} 
in which we also included the bias updates of Eq.~\eqref{eq:biasUpdate}.
 
According to the ``lift-solve-retract'' method (Section~\ref{sec:GNmanifold}), at each Gauss-Newton iteration we need to re-parametrize~\eqref{eq:residuals} using the retraction~\eqref{eq:retractionT}.
Then, the ``solve'' step requires to linearize the resulting cost around the current estimate.
For the purpose of linearization, it is convenient to compute analytic expressions of the Jacobians of the residual errors, which we derive in the Appendix~\ref{app:residualJacobians}.

\subsection{Bias Model}

When presenting the IMU model~\eqref{eq:omegaModel}, we said that biases are slowly time-varying quantities. 
Hence, we model them with a ``Brownian motion'', \emph{i.e.}, integrated white noise:
\beq
  \label{eq:continuouBias}
  \dot\bias^{g}(t) = \noise^{bg}, \qquad \dot\bias^a(t) = \noise^{ba}.
\eeq
Integrating~\eqref{eq:continuouBias} over the time interval $[t_i,t_j]$ between two consecutive keyframes $i$ and $j$ we get:
\beq
  \label{eq:discreteBias}
  \bias^{g}_j = \bias^{g}_i + \noise^{bgd}, \qquad \bias^a_j = \bias^a_i + \noise^{bad},
\eeq
where, as done before, we use the shorthand $\bias^{g}_i \doteq \bias^{g}(t_i)$, and we define the discrete noises $\noise^{bgd}$ and $\noise^{bad}$, which have zero mean and covariance $\Cov^{bgd} \doteq \Delta t_{ij} \text{Cov}(\noise^{bg})$ and $\Cov^{bad} \doteq \Delta t_{ij} \text{Cov}(\noise^{ba})$, respectively (\emph{cf.}~\cite[Appendix]{Crassidis06taes}).

The model~\eqref{eq:discreteBias} can be readily included in our factor graph, as a further additive term in~\eqref{eq:cost} for all consecutive keyframes:
\beq
  \textstyle
  \| \residual_{\bias_{ij}} \|^2 \doteq \| \bias^{g}_j - \bias^{g}_i \|^2_{\Cov^{bgd}} +\| \bias^{a}_j - \bias^{a}_i \|^2_{\Cov^{bad}}  
\eeq
%!TEX root = main.tex
\section{Structureless Vision Factors}
\label{sec:structureless}

In this section we introduce our structureless model for vision measurements.
The key feature of our approach is the linear elimination of landmarks.
Note that the elimination is repeated at each Gauss-Newton iteration, hence we are still guaranteed to obtain the optimal MAP estimate.

Visual measurements contribute to the cost~\eqref{eq:cost} via the sum:
\vspace{-4mm}
  \beq
  \label{eq:visionResiduals}
  \textstyle
  \sum_{i \in \mathcal{K}_k}\sum_{l \in \meascam_i} \| \residual_{\meascam_{il}} \|^2_{\covcam} 
  = 
  \sum_{l=1}^L\sum_{i \in \States(l)} \| \residual_{\meascam_{il}} \|^2_{\covcam} 
\eeq
which, on the right-hand-side, we rewrote as a sum of contributions of each landmark $l=1,\ldots,L$.
In~\eqref{eq:visionResiduals}, $\States(l)$ denotes the subset of keyframes in which $l$ is seen.

A fairly standard model for the residual error of a single image measurement $\pixel_{il}$ is the reprojection error:
\beq
  \label{eq:pixelResidual}
  \textstyle
  \residual_{\meascam_{il}} = \pixel_{il} - \pi(\R_i,\tran_i,\landmark_l), 
\eeq
where $\landmark_l \in \Real^3$ denotes the position of the $l$-th landmark, and $\pi(\cdot)$ is a standard perspective projection, which also encodes the (known) IMU-camera transformation $\T_{\imu\camera}$.

Direct use of~\eqref{eq:pixelResidual} would require to include the landmark positions $\landmark_l$, $l=1,\ldots,L$ in the optimization, and this impacts negatively on computation. 
Therefore, in the following we adopt a \emph{structureless} approach that avoids optimization over the landmarks, thus ensuring to retrieve the \MAP estimate.
 
As recalled in Section~\ref{sec:GNmanifold}, at each \GN iteration, we \emph{lift} the cost function, using the retraction~\eqref{eq:retractionT}. 
For the vision factors this means that the original residuals~\eqref{eq:visionResiduals} become:
\beq
  \label{eq:visionResidualsLifted}
  \textstyle
  \sum_{l = 1}^L \sum_{i \in \States(l)} \|  \pixel_{il} - \check{\pi}(\rotvecpert_i,\tranpert_i,\landpert_l) \|^2_{\covcam}
\eeq
where $\rotvecpert_i, \tranpert_i, \landpert_l$ are now Euclidean corrections, and $\check{\pi}(\cdot)$ is the lifted cost function. 
The ``solve'' step in the \GN method is based on linearization of the residuals:
\beq
  \label{eq:visionResidualsLinearized}
  \textstyle
  \sum_{l = 1}^L \sum_{i \in \States(l)} \|  \Fmat_{il} \posepert_i + \Emat_{il} \landpert_l - \bvec_{il}  \|^2,
\eeq
where $\posepert_i \doteq [\rotvecpert_i \; \tranpert_i]^\transpose$; the Jacobians $\Fmat_{il}, \Emat_{il}$, and the vector $\bvec_{il}$ (both normalized by $\covcam^{1/2}$) result from the linearization.
The vector $\bvec_{il}$ is the residual error at the linearization point.

Writing the second sum in~\eqref{eq:visionResidualsLinearized} in matrix form we get: 
\beq
  \label{eq:visionResidualsLinearized3}
  \textstyle
  \sum_{l = 1}^L \|  \Fmat_{l} \; \posepert_{\States(l)} + \Emat_{l} \; \landpert_l - \bvec_{l}  \|^2
\eeq
where $\Fmat_{l}, \Emat_{l}, \bvec_{l}$ are obtained by stacking $\Fmat_{il},\Emat_{il}, \bvec_{il}$, respectively, for all  $i \in \States(l)$.

Since a landmark $l$ appears in a single term of the sum~\eqref{eq:visionResidualsLinearized3}, for any given choice of the pose perturbation $\posepert_{\States(l)}$, 
the landmark perturbation $\landpert_l$ that minimizes the quadratic cost 
$\|  \Fmat_{l} \; \posepert_{\States(l)} + \Emat_{l} \; \landpert_l - \bvec_{l}  \|^2$ is:
\beq
  \label{eq:landmarkOpt}
  \landpert_l = - (\Emat_{l}^\transpose \Emat_{l})\inv \Emat_{l}^\transpose 
  ( \Fmat_{l} \; \posepert_{\States(l)}  - \bvec_{l} )
\eeq
Substituting~\eqref{eq:landmarkOpt} back into~\eqref{eq:visionResidualsLinearized3} 
we can \emph{eliminate} the variable $\landpert_l$ from the optimization problem:
\beq
  \label{eq:cameraOnly}
  \sum_{l = 1}^L \| (\eye - \Emat_{l} (\Emat_{l}^\transpose \Emat_{l})\inv \Emat_{l}^\transpose) 
  \left( \Fmat_{l} \; \posepert_{\States(l)} - \bvec_{l} \right)  \|^2,
\eeq
where $\eye - \Emat_{l}(\Emat_{l}^\transpose \Emat_{l})\inv \Emat_{l}^\transpose$ is an orthogonal projector of $\Emat_{l}$.
In Appendix~\ref{sec:appendix_structureless} we show that the cost~\eqref{eq:cameraOnly} can be further manipulated, leading to a more efficient implementation.

This approach is well known in the bundle adjustment literature as the \emph{Schur complement trick}, where a standard practice is to update the linearization point of $\landmark_l$ via \emph{back-substitution}~\cite{Hartley04book}.
In contrast, we obtain the updated landmark positions from the linearization point of the poses using a fast linear triangulation.
Using this approach, we reduced a large set of factors~\eqref{eq:visionResidualsLifted} which involve poses and landmarks into a smaller set of $L$ factors~\eqref{eq:cameraOnly}, which only involve poses.
In particular, the factor corresponding to landmark $l$ only involves the states $\States(l)$ observing $l$, creating the connectivity pattern of \Figure~\ref{fig:factor_graph}.
The same approach is also used in MSC-KF \cite{Mourikis07icra} to avoid the inclusion of landmarks in the state vector.
However, since MSC-KF can only linearize and absorb a measurement once, the processing of measurements needs to be delayed until all measurements of the same landmark are observed.
This does not apply to the proposed optimization-based approach, which allows for multiple relinearizations and the incremental inclusion of new measurements.
%!TEX root = main.tex
\section{Experimental Analysis}
\label{sec:experiments}

We tested the proposed approach on both simulated and real data.
Section~\ref{sec:experiments_synthetic} reports simulation results, showing that our 
approach is accurate, fast, and consistent.
Section~\ref{sec:real_experiments} compares our approach against the state-of-the-art, 
confirming its superior accuracy in real indoor and outdoor experiments.  

\subsection{Simulation Experiments}
\label{sec:experiments_synthetic}

\begin{figure}[t!]
  \centering
  \includegraphics[width=1\linewidth]{\figdirvin/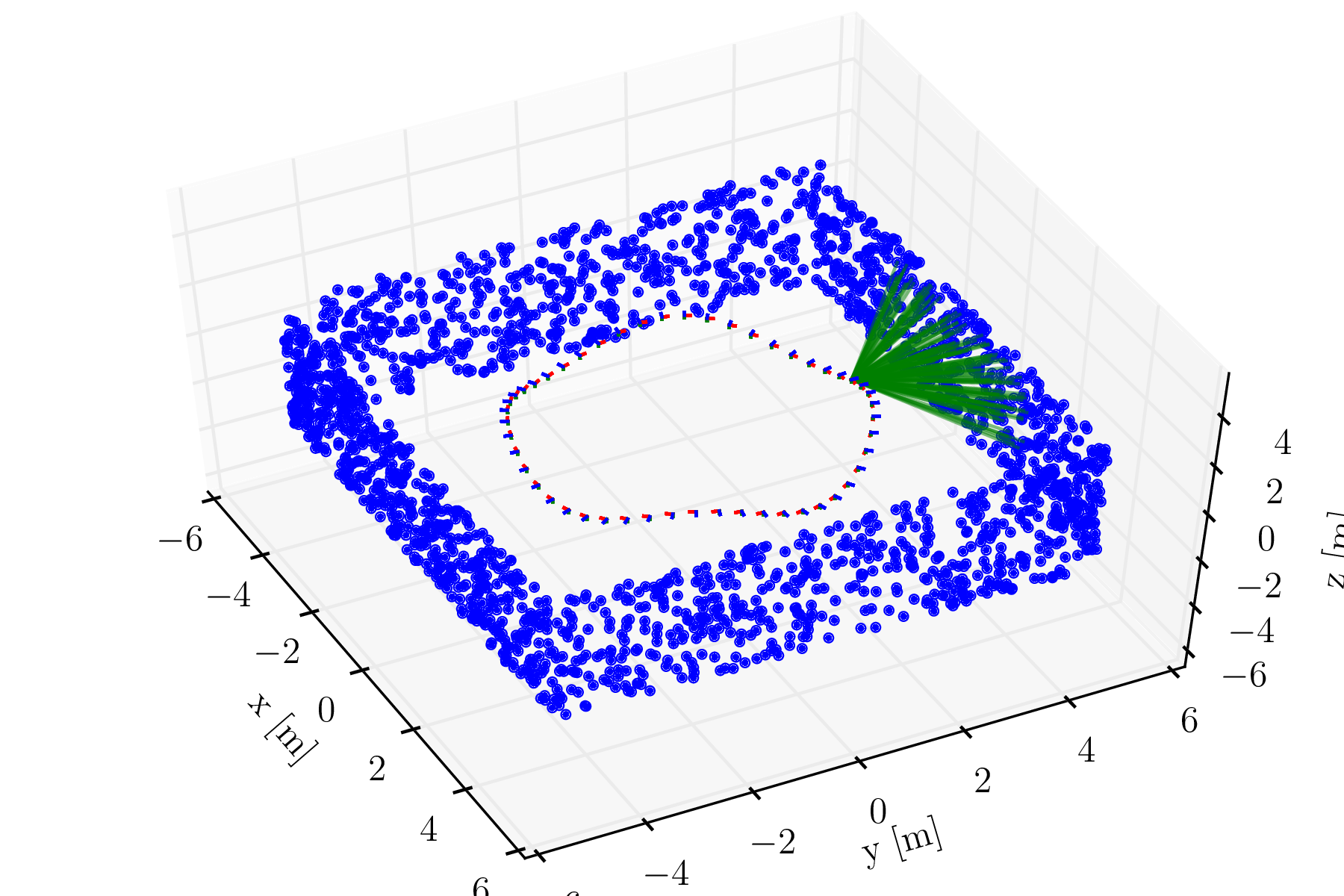}
  \caption{Simulation setup: The camera moves along a circular trajectory while observing features (green lines) on the walls of a square environment.}
  \label{fig:synthetic_environment}
  \vspace{-15pt}
\end{figure}

We simulated a camera following a circular trajectory of three meter radius with a sinusoidal vertical motion.
The total length of the trajectory is 120 meters.
While moving, the camera observes landmarks as depicted in \Figure \ref{fig:synthetic_environment}.
The number of landmark observations per frame is limited to 50. 
To simulate a realistic feature-tracker, we corrupt the landmark measurements with isotropic 
Gaussian noise with standard deviation $\sigma_\text{px}=1$~pixel.
The camera has a focal length of $315$~pixels and runs at a rate of 2.5~Hz (simulating keyframes).
The simulated acceleration and gyroscope measurements are computed from the analytic derivatives of the parametric trajectory and additionally corrupted by white noise and a slowly time-varying bias terms, according to the IMU model in \Equation\eqref{eq:omegaModel}.\footnote{We used the following IMU parameters:
Gyroscope and accelerometer continuous-time noise density: $\sigma^{g} = 0.0007 \ [\text{rad}/(\text{s} \sqrt{\text{Hz}})]$, $\sigma^{a} = 0.019 \ [\text{m}/(\text{s}^2 \sqrt{\text{Hz}})]$. Gyroscope and accelerometer \emph{bias} continous-time noise density: 
$\sigma^{bg} = 0.0004 \ [\text{rad}/(\text{s}^2 \sqrt{\text{Hz}})]$, 
$\sigma^{ba} = 0.012 \ [\text{m}/(\text{s}^3 \sqrt{\text{Hz}})]$.} 
To evaluate our approach, we performed a Monte Carlo analysis with 50 simulation runs, each with different realizations of process and measurement noise.
In each run we compute the MAP estimate using the IMU and the vision models presented in this paper.
The optimization (whose solution is the MAP estimate) is solved using 
 the incremental smoothing algorithm iSAM2~\cite{Kaess12ijrr}.
iSAM2 uses the Bayes tree~\cite{Kaess10tr} data structure to obtain efficient variable ordering that minimizes fill-in in the square-root information matrix and, thus, minimizes computation time.
Further, iSAM2 exploits the fact that new measurements often have only local effect on the MAP estimate, hence applies incremental updates directly to the square-root information matrix, only re-solving for the variables affected by a new measurement.

In the following we present the results of our experiments, organized in four subsections:
1) pose estimation accuracy and timing, 
2) consistency, 
3) bias estimation accuracy, and 
4) first-order bias correction. Then, in Section~\ref{sec:euler} we compare 
our approach against the original proposal of~\cite{Lupton12tro}.

%%%%%%%%%%%%%%%%%%%%%%%%%%%%%%%%%%%%%%%%%%%%%%%%%%%%%%%%%%%%
\subsubsection{Pose Estimation Accuracy and Timing}
\label{sec:experiments_batch_solution}

\begin{figure}[t!]
  \centering
  \includegraphics[trim=8.5pt 10pt 9pt 5pt, clip=true, width=1.01\linewidth]{\figdirvin/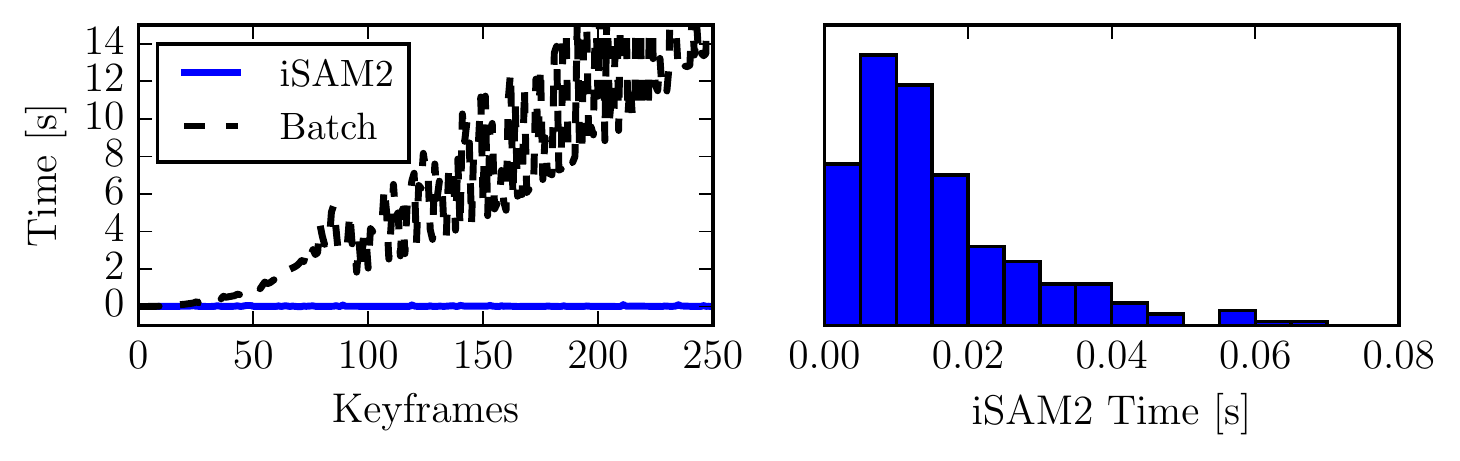}
  \caption{
  Left: CPU time required for inference, comparing  
  batch estimation against iSAM2. Right: histogram plot 
  of CPU time for the proposed approach.}
  \label{fig:timing_comparison}
  \vspace{-15pt}
\end{figure}

\begin{figure}[t!]
  \centering
  \includegraphics[trim=0pt 9pt 0pt 5pt, clip=true, width=1\linewidth]{\figdirvin/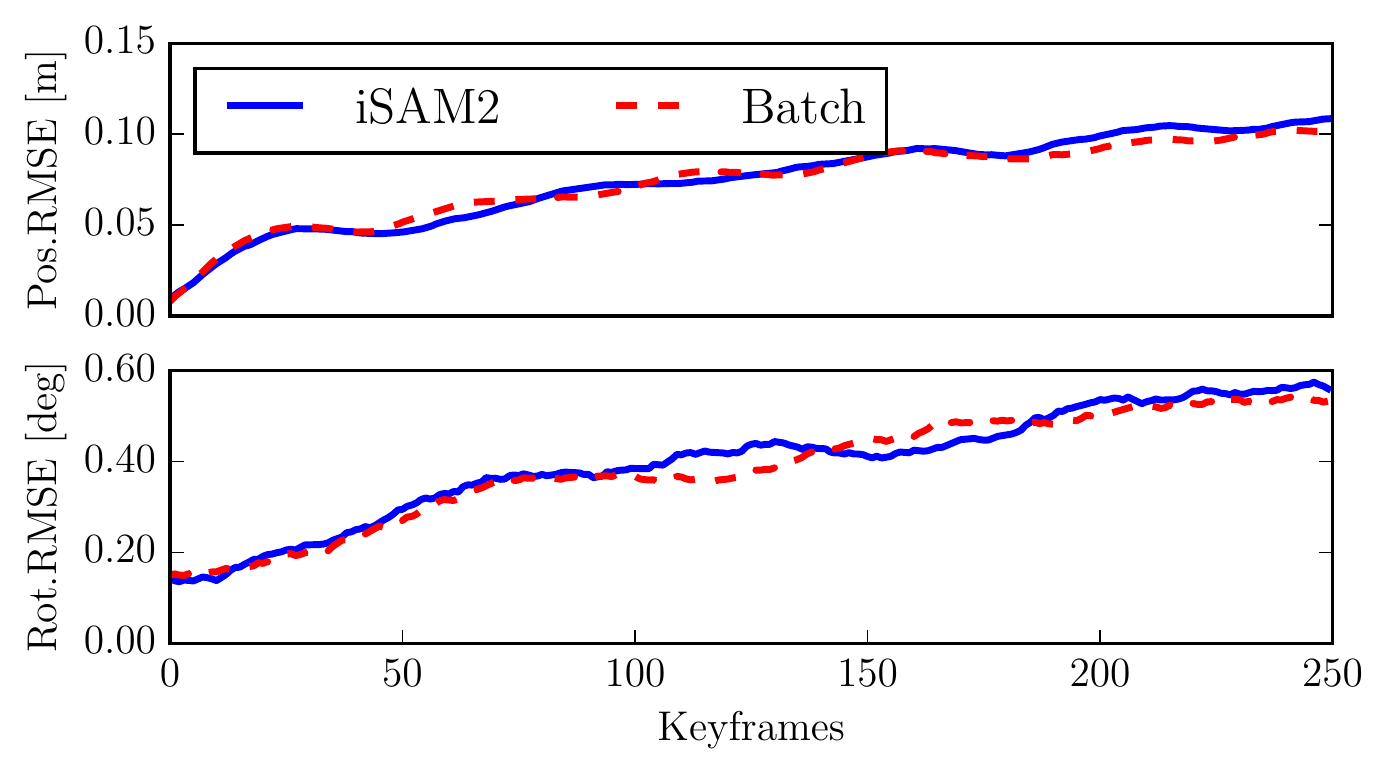}
  \caption{Root Mean Squared Error (RMSE) averaged over 50 Monte Carlo experiments, comparing 
  batch nonlinear optimization and iSAM2.}
  \label{fig:rms_average}
  \vspace{10pt}
  \includegraphics[trim=0pt 5pt 0pt -3pt, clip=true, width=1\linewidth]{\figdirvin/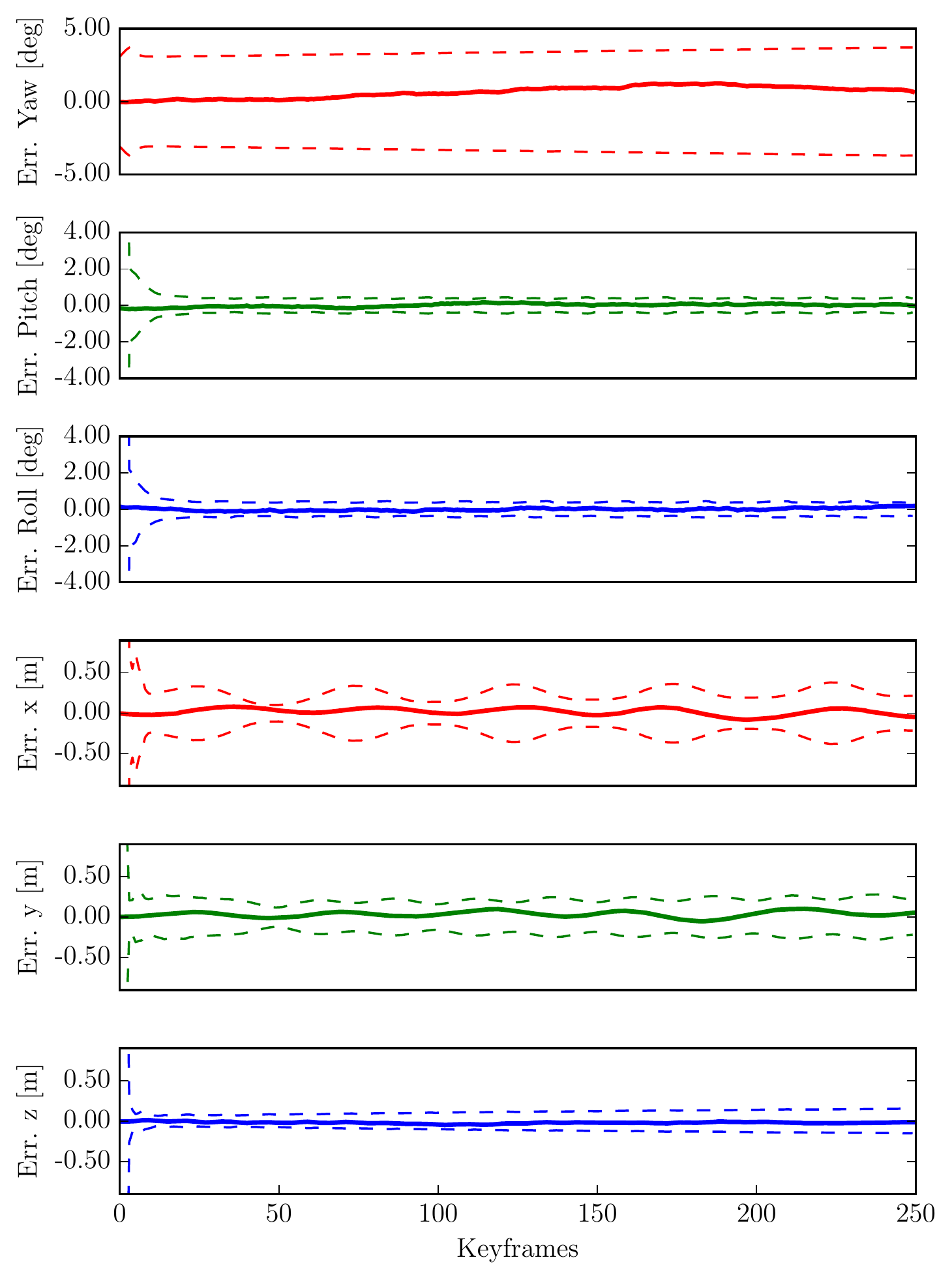}
  \caption{Orientation and position errors with 3$\sigma$ bounds (single simulation).}
  \label{fig:consistency_single_run}
  \vspace{-10pt}
\end{figure}

The optimal 
MAP estimate is given by the \emph{batch} nonlinear optimization of the least-squares objective in \Equation\eqref{eq:cost}. 
However, as shown on the left in \Figure\ref{fig:timing_comparison}, the computational cost of batch optimization 
quickly increases  as the trajectory length grows. 
A key ingredient that makes our approach extremely efficient is the use of the incremental smoothing algorithm iSAM2~\cite{Kaess12ijrr}, which performs close-to-optimal inference, while preserving real-time capability. \Figure\ref{fig:rms_average} shows that the accuracy of iSAM2 is 
practically the same as the batch estimate.
In odometry problems, the iSAM2 algorithm results in approximately constant update time per frame (\Figure\ref{fig:timing_comparison}, left), which in our experiment is approximately 10 milliseconds per update (\Figure\ref{fig:timing_comparison}, right).

%%%%%%%%%%%%%%%%%%%%%%%%%%%%%%%%%%%%%%%%%%%%%%%%%%%%%%%%%%%%
\subsubsection{Consistency}

For generic motion, the \VIO problem has four unobservable degrees of freedom, three corresponding to the global translation and one to the global orientation around the gravity direction (yaw), see \cite{Kottas2012iser}.
A \VIO algorithm must preserve these observability properties and avoid inclusion of spurious information along the unobservable directions, which would result in inconsistency~\cite{Kottas2012iser}.
\Figure\ref{fig:consistency_single_run} reports orientation and position errors with the corresponding $3\sigma$ bounds, confirming that our approach is consistent.
In the \VIO problem, the gravity direction is observable, hence the uncertainty on 
roll and pitch  remains bounded. 
In contrast, global yaw and position cannot be measured and the uncertainty slowly grows over time.

To present more substantial evidence of the fact that our estimator is consistent, we
recall a standard measure of consistency, the \emph{average} Normalized Estimation Error Squared (NEES)~\cite{BarShalom01}.
The NEES is the squared estimation error $\boldsymbol\epsilon_k$ normalized by the estimator-calculated covariance ${\boldsymbol\Sigma}_k$:
\begin{align}\label{eq:nees}
  \eta_k &\doteq \boldsymbol\epsilon_k^\transpose\hat{\boldsymbol\Sigma}_k^{-1}\boldsymbol\epsilon_k \qquad\quad \text{(NEES)}
\end{align}

The error in estimating the current pose is computed as:
\begin{align}\label{eq:error}
  \boldsymbol\epsilon_k \doteq \big[ \logmap(\hat\R_k^\transpose\R_k^\text{gt}), \ \hat\R_k^\transpose(\hat\tran_k-\tran_k^\text{gt}) \big]^\transpose
\end{align}
where the exponent ``gt'' denotes ground-truth states and $(\hat\R_k, \hat\tran_k)$ denotes the estimated pose at time~$k$. 
Note that the error~\eqref{eq:error} is expressed in the body frame and it is consistent with our choice of the retraction in 
\Equation\eqref{eq:retractionT} (intuitively, the retraction applies the perturbation in the body frame). 

The \emph{average} NEES over $N$ independent Monte Carlo runs, can be computed 
by averaging the NEES values:
\beq\label{eq:aNEES}
 \textstyle \bar{\eta}_k = \frac{1}{N}\sum_{i=1}^N \eta^{(i)}_{k} \qquad \quad \text{(average NEES)}
\eeq
where $\eta^{(i)}_{k}$ is the NEES computed at the $i$-th Monte Carlo run.
If the estimator is consistent, then $N\textstyle \bar{\eta}_k$ is $\chi^2_n$ chi-square distributed with 
$n=\text{dim}(\boldsymbol\epsilon_k) \cdot N$ degrees of freedom \cite[pp. 234]{BarShalom01}.
We evaluate this hypothesis with a $\chi^2_n$ acceptance test \cite[pp. 235]{BarShalom01}.
For a significance level $\alpha = 2.5\%$ and $n = \text{dim}(\boldsymbol\epsilon_k) \cdot N = 6 \cdot 50$, 
the acceptance region of the test is given by the two-sided probability concentration region $\bar{\eta}_k \in [5.0,7.0]$.
If $\bar{\eta}_k$ rises significantly higher than the upper bound, the estimator is overconfident, 
if it tends below the lower bound, it is conservative. In \VIO one usually wants to avoid 
overconfident estimators: the fact that $\bar{\eta}_k$ exceeds the upper bound is an indicator of the fact that  
the estimator is including spurious information in the inference process.

\begin{figure}[t!]
  \centering
  \includegraphics[trim=0pt 9pt 0pt 0pt, clip=true, width=1\linewidth]{\figdirvin/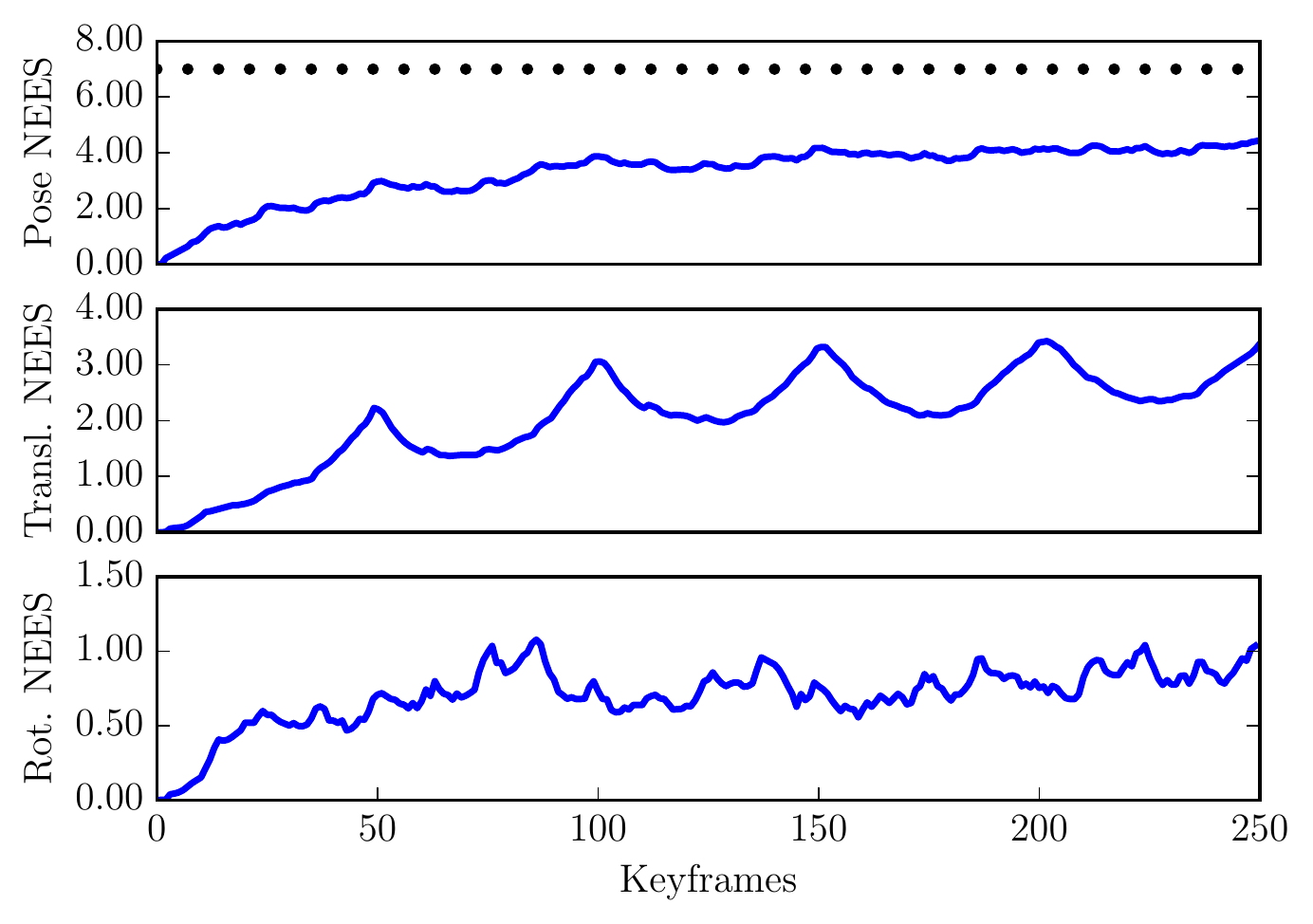}
  \caption{Normalized Estimation Error Squared (NEES) averaged over 50 Monte Carlo runs. 
  The average NEES is reported for the current pose (top), current position (middle), and current rotation (bottom).}
  \label{fig:nees_average}
  \vspace{3pt}
  \includegraphics[trim=0pt 15pt 0pt 5pt, clip=true, width=1\linewidth]{\figdirvin/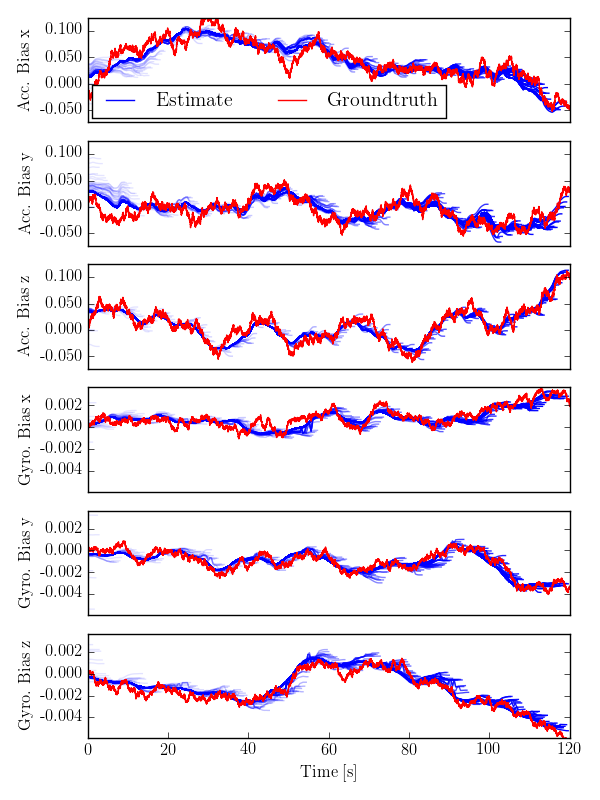}
  \caption{
  Comparison between ground truth bias (red line) and estimated bias (blue lines) in a Monte Carlo run.}
  \label{fig:synthetic_bias}
  \vspace{4pt}
  \includegraphics[width=0.97\linewidth]{\figdirvin/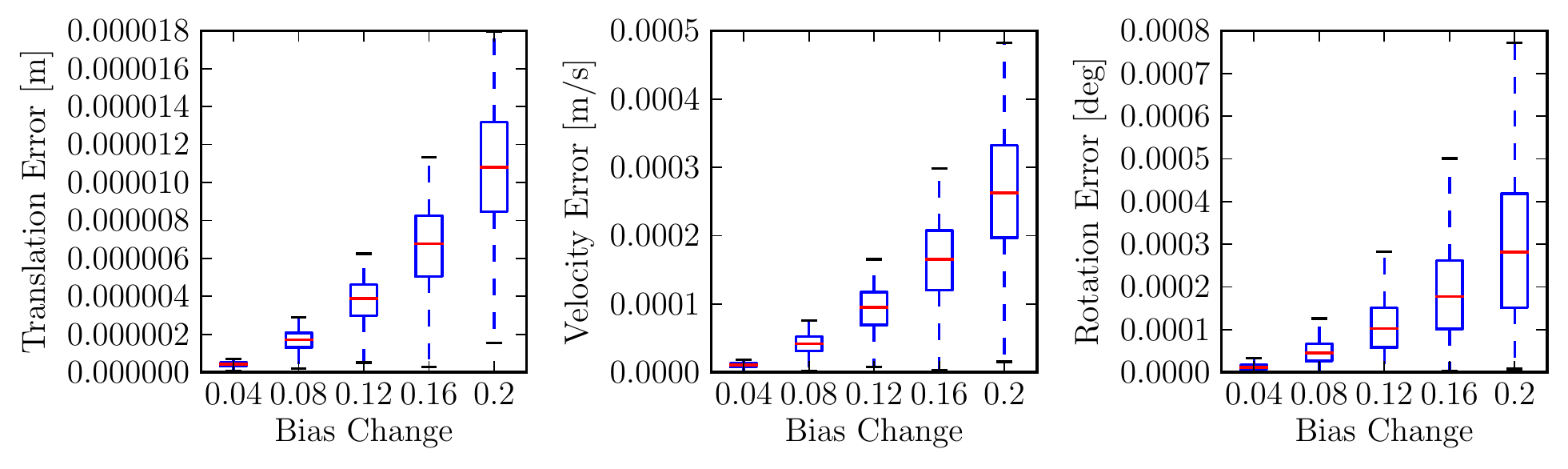}
  \caption{Error committed 
  when using the first-order approximation~\eqref{eq:biasUpdate} instead 
  of repeating the integration, for different bias perturbations. 
  Left: $\preintPmeas_{ij}(\biasFixed_i + \biaspert_i)$ error; 
  Center: $\preintVmeas_{ij}(\biasFixed_i + \biaspert_i)$ error;
  Right: $\preintRmeas_{ij}(\biasFixed_i + \biaspert_i)$ error.
  Statistics are computed over 1000 Monte Carlo runs.}
  \label{fig:bias_prediction_error}
\end{figure}

In \Figure\ref{fig:nees_average} we report the average NEES of the proposed approach. 
The average NEES approaches the lower bound but, more importantly, it remains below the upper bound at 7.0 (black dots), which assures that the estimator is not overconfident.
We also report the average rotational and translational NEES to allow a comparison with the observability-constrained EKF in \cite{Kottas2012iser,Hesch14ijrr}, which obtains similar results by enforcing explicitly the observability properties in EKF. 

%%%%%%%%%%%%%%%%%%%%%%%%%%%%%%%%%%%%%%%%%%%%%%%%%%%%%%%%%%%%
\subsubsection{Bias Estimation Accuracy}

Our simulations allow us to compare the estimated gyroscope and accelerometer bias with the true biases that were used to corrupt the simulated inertial measurements.
\Figure\ref{fig:synthetic_bias} shows that biases estimated by our approach (in blue) correctly track the ground truth biases (in red).
Note that, since we use a smoothing approach, at each step, we potentially change the entire history of the bias estimates, hence 
we visualize the bias estimates using multiple curves.
Each curve represents the history of the estimated biases from time zero (left-most extreme of the blue curve) to the current time (right-most extreme of the blue curve).

%%%%%%%%%%%%%%%%%%%%%%%%%%%%%%%%%%%%%%%%%%%%%%%%%%%%%%%%%%%%
\subsubsection{First-Order Bias Correction}

We performed an additional Monte-Carlo analysis to evaluate the a-posteriori bias correction proposed in Section~\ref{sec:biasUpdate}.
The preintegrated measurements are computed with the bias estimate at the time of integration. 
However, as seen in \Figure\ref{fig:synthetic_bias}, the bias estimate for an older preintegrated measurement may change when more information becomes available.
To avoid repeating the integration when the bias estimate changes, we perform a first-order correction of the preintegrated measurement according to \Equation\eqref{eq:biasUpdate}.
The accuracy of this first order bias correction is reported in \Figure\ref{fig:bias_prediction_error}.
To compute the statistics, we integrated 100 random IMU measurements with a given bias estimate $\biasFixed_i$ which results in the preintegrated measurements $\preintRmeas_{ij}(\biasFixed_i), \preintVmeas_{ij}(\biasFixed_i)$ and $\preintPmeas_{ij}(\biasFixed_i)$. 
Subsequently, a random perturbation $\biaspert_i$ with magnitude between $0.04$ and $0.2$ was applied to both the gyroscope and accelerometer bias.
We repeated the integration at $\biasFixed_i + \biaspert_i$ to obtain  $\preintRmeas_{ij}(\biasFixed_i + \biaspert_i), \preintVmeas_{ij}(\biasFixed_i + \biaspert_i)$ and $\preintPmeas_{ij}(\biasFixed_i + \biaspert_i)$.
This ground-truth result was then compared against the first-order correction in~\eqref{eq:biasUpdate} to compute the error of the approximation. The errors resulting from the first-order approximation are negligible, even for relatively large 
bias perturbations.

%%%%%%%%%%%%%%%%%%%%%%%%%%%%%%%%%%%%%%%%%%%%%%%%%%%%%%%%%%%%
\subsubsection{Advantages over the Euler-angle-based formulation}
\label{sec:euler}
\newcommand{\eulervec}{\boldsymbol\theta}

In this section we compare the proposed IMU preintegration with the original formulation of \cite{Lupton12tro}, based 
on Euler angles.
We observe three main problems with the preintegration using Euler angles, which are avoided in our formulation.

The first drawback is that, in contrast to the integration using the exponential map in \Equation\eqref{eq:euler}, 
the rotation integration based on Euler angles is only exact up to the first order. For the interested reader, we
 recall the rotation rate integration using Euler angles in Appendix~\ref{sec:appendix_euler}.
On the left of \Figure\ref{fig:euler_fairness}, we report the integration errors commited by the Euler angle parametrization 
when integrating angular rates with randomly selected rotation axes and magnitude in the range from 1 to 3 rad/s.
 Integration error in Euler angles accumulates quickly when 
the sampling time $\Delta t$ or the angular rate $\tilde\rotvel$ are large. On the other hand, the proposed 
approach, which performs integration directly on the rotation manifold, is exact, regardless the values of $\Delta t$ and~$\tilde\rotvel$.

The second drawback is that the Euler parametrization is not \emph{fair} \cite{Hornegger97}, which means that, given the preintegrated Euler angles $\tilde\eulervec$, the negative log-likelihood 
$
  \label{eq:eulerLikelihood}
  \calL(\eulervec) = \frac{1}{2}\|\tilde\eulervec - \eulervec\|^2_\Sigma
$ is not invariant under the action of rigid body transformations.
On the right of \Figure\ref{fig:euler_fairness} we show experimentally how the log-likelihood changes when the frame of reference is rotated around randomly selected rotation axes. 
This essentially means that an estimator using Euler angles may give different results for different choices of the 
world frame (\cf with \Figure\ref{fig:frames}). On the other hand, the 
 $\SOthree$ parametrization can be easily seen to be fair (the negative likelihood~\eqref{eq:minusloglikeSO3-2} 
 can be promptly seen to be left invariant), and this is confirmed by \Figure\ref{fig:euler_fairness} (right).

The third drawback is the existence of so-called \emph{gimball lock} singularities.
For a $zyx$ Euler angle parametrization, the singularity is reached at pitch values of $\theta = \frac{\pi}{2}+n\pi$, for $n \in \mathbb{Z}$.
To evaluate the effect of the singularity and how it affects the computation of preintegrated measurement noise, we performed 
the following Monte Carlo analysis.
We simulated a set of trajectories that reach maximum pitch values $\theta_\text{max}$ of increasing magnitude.
For each trajectory, we integrate the rotation uncertainty using the Euler parametrization and the proposed on-manifold approach.
The ground-truth covariance is instead obtained through sampling. % (and propagating the measurement noise).
We use the Kullback-Leibler (KL) divergence to quantify the mismatch between the estimated covariances and the ground-truth one.
The results of this experiment are shown in \Figure\ref{fig:euler_vs_so3_kld_new}, where we observe that the closer we get to the singularity, the worse is the noise propagation using Euler angles.
On the other hand, the proposed approach can accurately estimate the measurement covariance, 
independently on the motion of the platform.

\begin{figure}[t!]
  \centering
  \subfigure[]{\includegraphics[trim=7pt 8pt 7pt 8pt, clip=true, width=0.49\linewidth]{\figdirvin/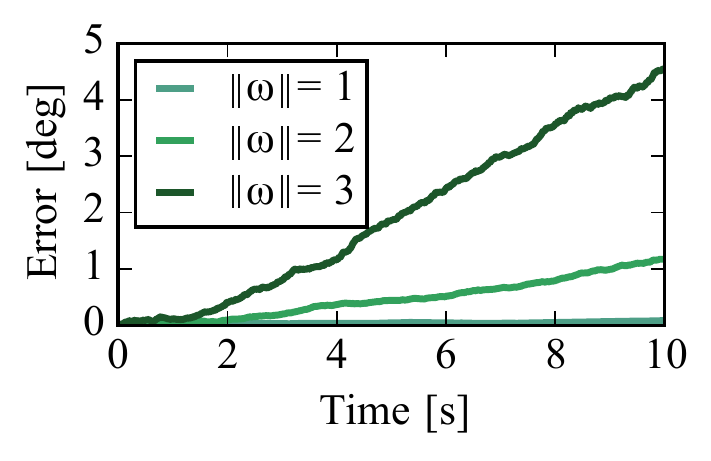}}
  \subfigure[]{\includegraphics[trim=7pt 8pt 7pt 8pt, clip=true, width=0.49\linewidth]{\figdirvin/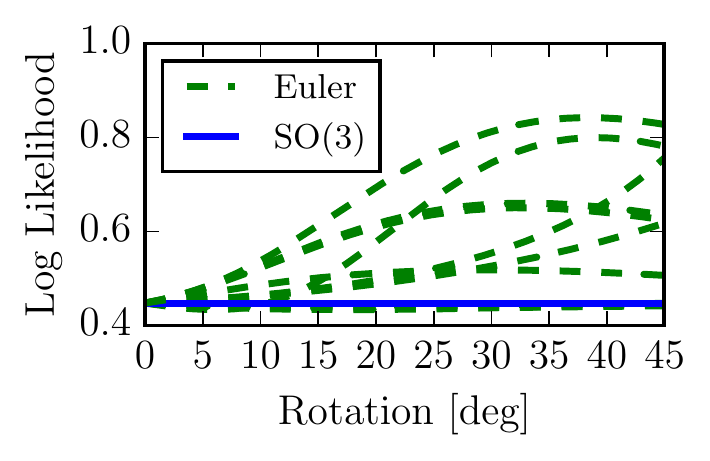}}
  \vspace{-7pt}
  \caption{(a) Integration errors committed with the Euler angle parametrization for angular velocities $\boldsymbol\omega$ of increasing magnitude [rad/s]. (b) Negative log-likelihood of a rotation measurement under the action of random rigid body transformations.}
  \label{fig:euler_fairness}
  \vspace{10pt}
  \includegraphics[trim=0pt 13pt 0pt 8pt, clip=true, width=1\linewidth]{\figdirvin/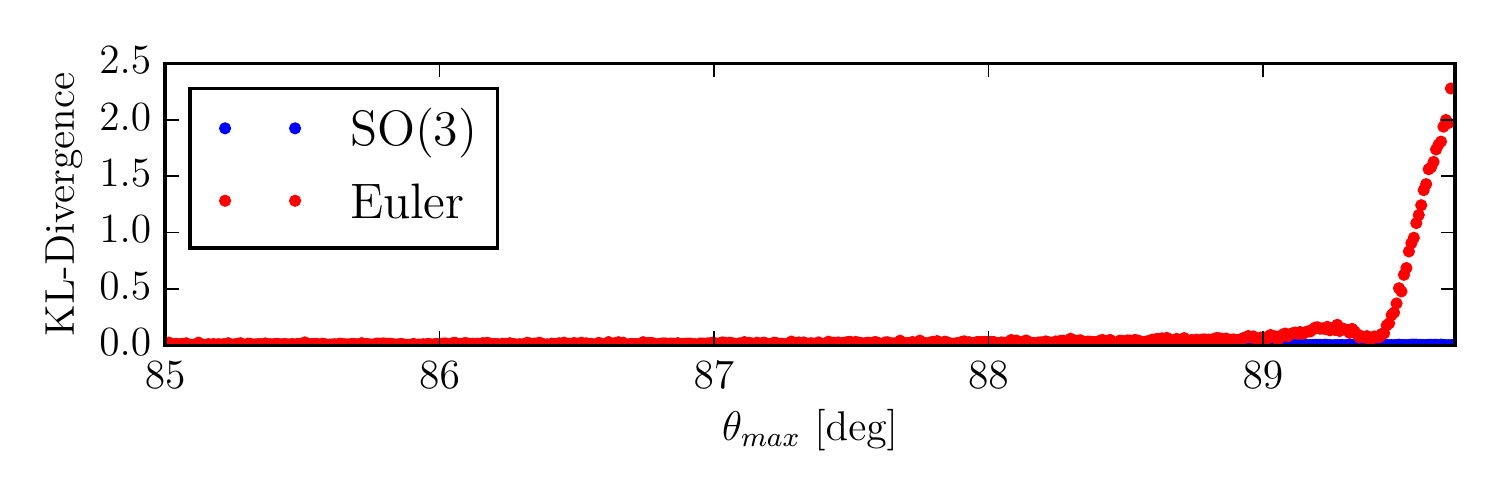}
  \caption{Kullback-Leibler divergence between the preintegrated rotation covariance --computed using Euler angles (red) and the proposed approach (blue)-- and the ground-truth covariance. The Euler angle parametrization degrades close to the singularity at $\theta_\text{max} = 90$ deg while the proposed on-manifold approach is accurate regardless of the motion.}
  \label{fig:euler_vs_so3_kld_new} 
  \vspace{-10pt}
\end{figure}
%!TEX root = main.tex
\subsection{Real Experiments}
\label{sec:real_experiments}

We integrated the proposed inertial factors in a monocular \VIO pipeline to benchmark its performance against the state of the art.
In the following, we first discuss our implementation, and then present results from an indoor experiment with motion-capture ground-truth.
Finally, we show results from longer trajectories in outdoor experiments.
The results confirm that our approach is more accurate than state-of-the-art filtering and fixed-lag smoothing algorithms, and enables fast inference in real-world problems.

%!TEX root = main.tex
\subsubsection{Implementation}
\label{sec:implementation}

% Front-end and Back-end
Our implementation consists of a high frame rate tracking front-end based on SVO\footnote{\url{http://github.com/uzh-rpg/rpg_svo}} \cite{Forster14icra} and an optimization back-end based on iSAM2 \cite{Kaess12ijrr}\footnote{\url{ http://borg.cc.gatech.edu}}.
The front-end tracks salient features in the image at camera rate while the back-end optimizes in parallel the state of selected \emph{keyframes} as described in this paper. 

% SVO
SVO \cite{Forster14icra} is a precise and robust monocular visual odometry system that employs \emph{sparse image alignment} to estimate incremental motion and tracks features by minimizing the photometric error between subsequent frames.
The difference to tracking features individually, as in standard Lucas-Kanade tracking, is that we exploit the known depth of features from previous triangulations.
This allows us to track all features as a bundle in a single optimization that satisfies epipolar constraints; hence, outliers only originate from erroneous triangulations.
In the visual-inertial setting, we further exploit the availability of accurate rotation increments, obtained by integrating angular velocity measurements from the gyroscope.
These increments are used as rotation priors in the sparse-image-alignment algorithm, and this increases the overall robustness of the system.
The motion estimation is combined with an outlier resistant probabilistic triangulation method that is implemented with a recursive Bayesian filter. The high frame-rate motion estimation combined with the robust depth estimation results in increased robustness in scenes with repetitive and high frequency texture (\emph{e.g.}, asphalt).
The output of SVO are selected keyframes with feature-tracks corresponding to triangulated landmarks.
This data is passed to the back-end that computes the visual-inertial MAP estimate in~\eq~\eqref{eq:cost} using iSAM2~\cite{Kaess12ijrr}.

We remark that our approach does not marginalize out past states. Therefore, while the approach is designed for fast visual-inertial 
odometry, if desired, it could be readily extended to incorporate loop closures.

\subsubsection{Indoor Experiments}
\label{sec:indoor_experiments}

\begin{figure}[t!]
  \centering
  \includegraphics[width=1.0\linewidth]{\figdirvin/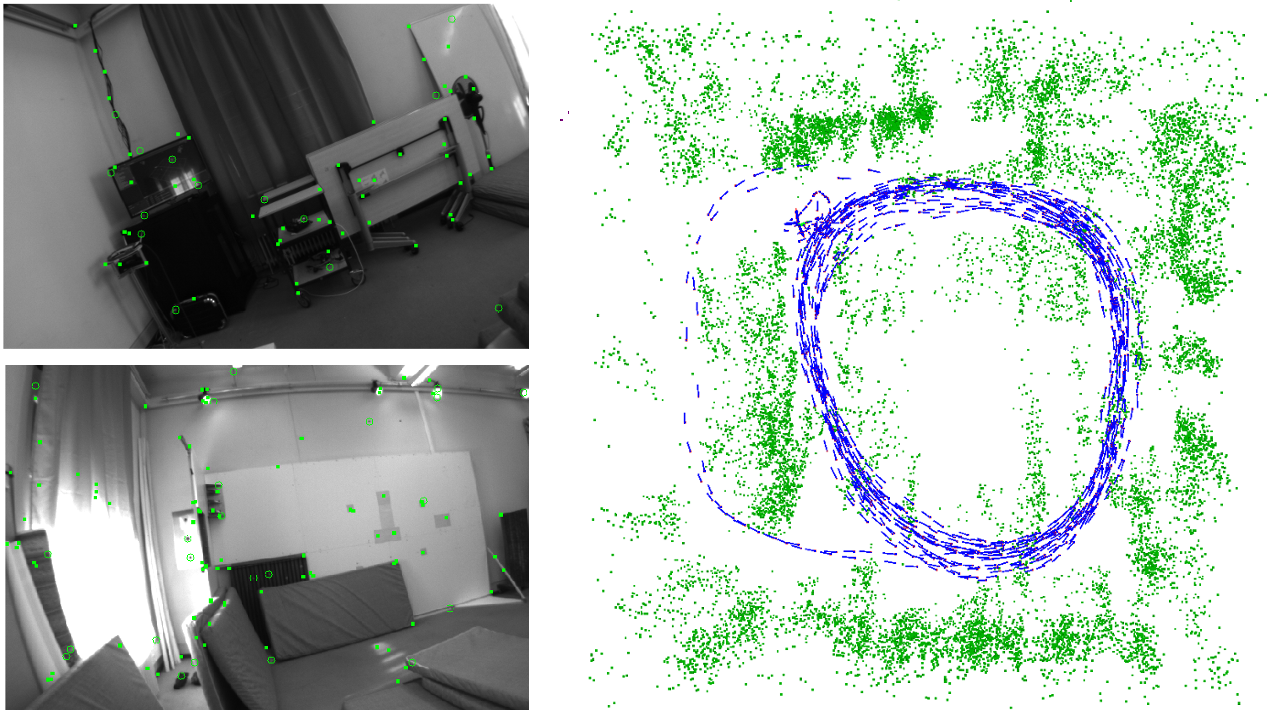}
  \caption{Left: two images from the indoor trajectory dataset with tracked features in green. 
  Right: top view of the trajectory estimate produced by our approach (blue) and 3D landmarks triangulated from the 
  trajectory (green).}
  \label{fig:vicon_trajectory}
  \vspace{10px}
  \includegraphics[width=1\linewidth]{\figdirvin/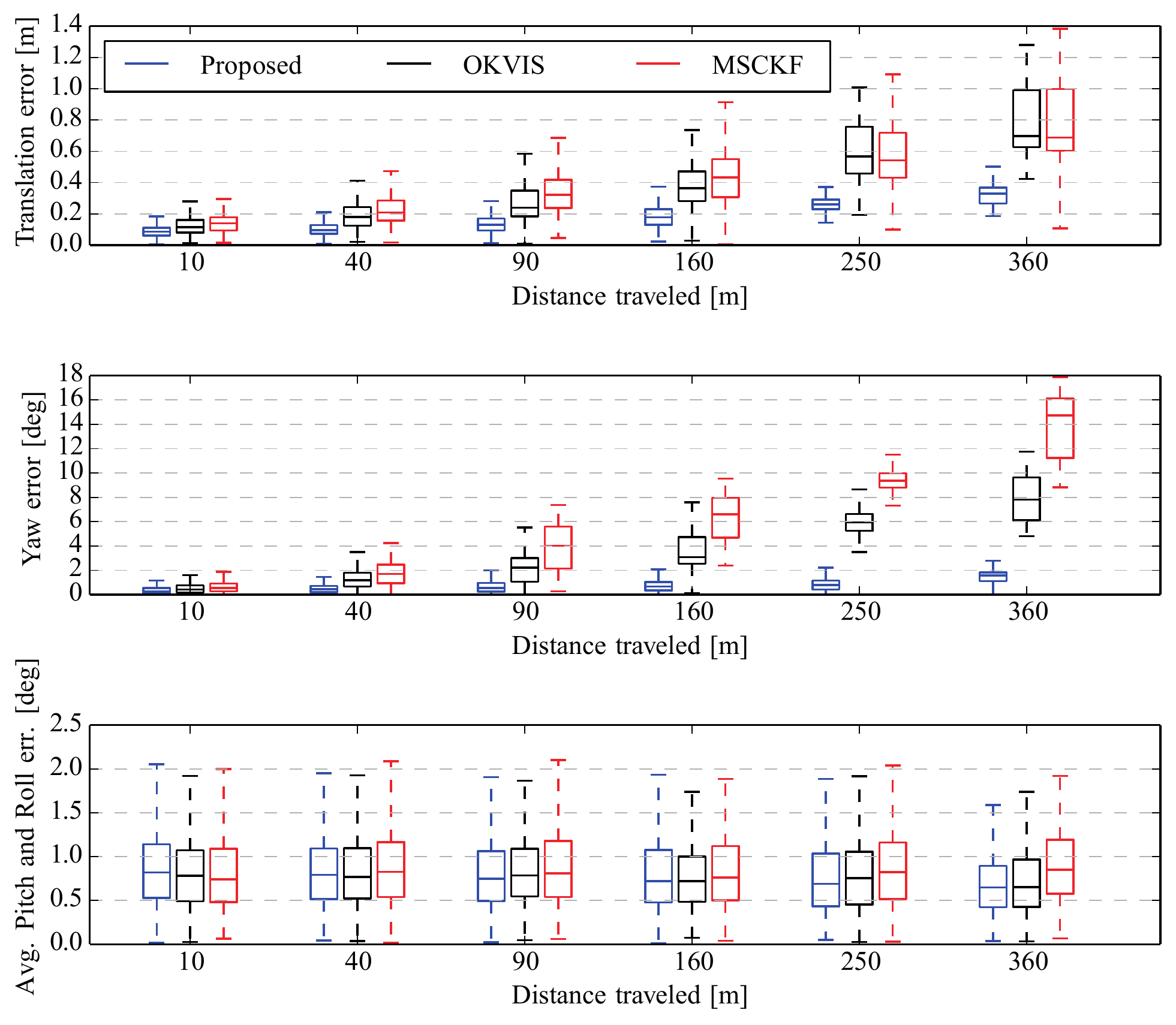}
  \caption{Comparison of the proposed approach 
  versus the OKVIS algorithm~\cite{Leutenegger15ijrr} and an implementation of the MSCKF filter~\cite{Mourikis07icra}.
  Relative errors are measured over different segments of the trajectory, of length $\{10,40,90,160,250,360\}$m, according to the 
  odometric error metric in~\cite{Geiger12cvpr}.}
  \label{fig:vicon_errors}
  \vspace{10px}
  \includegraphics[width=1\linewidth]{\figdirvin/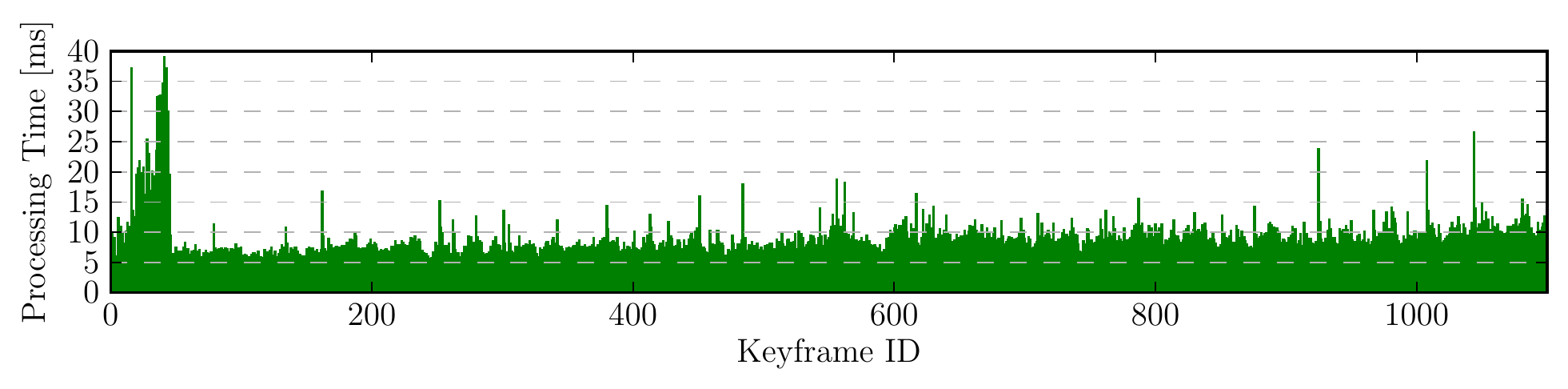}
  \caption{Processing-time per keyframe for the proposed \VIO approach.}
  \label{fig:processing_time}
  \vspace{-15pt}
\end{figure}

The indoor experiment shows that the  proposed approach is more accurate than two competitive state-of-the-art approaches, namely 
OKVIS\footnote{\url{https://github.com/ethz-asl/okvis}}~\cite{Leutenegger15ijrr}, and MSCKF~\cite{Mourikis07icra}.
% The source code of OKVIS is available at 
The experiment is performed on the 430m-long indoor trajectory of \Figure\ref{fig:vicon_trajectory}. 
The dataset was recorded with a forward-looking VI-Sensor \cite{Nikolic14icra} that consists of an ADIS16448 MEMS IMU and two embedded WVGA monochrome cameras (we only use the left camera).
Intrinsic and extrinsic calibration was obtained using~\cite{Furgale13iros}.
The camera runs at 20Hz and the IMU at 800Hz. 
Ground truth poses are provided by a Vicon system mounted in the room; the \emph{hand-eye} calibration between the Vicon markers and the camera is computed using a least-squares method~\cite{Park94tro}.

%!TEX root = main.tex

\Figure~\ref{fig:vicon_errors} compares the proposed system against the OKVIS algorithm~\cite{Leutenegger15ijrr}, and an implementation of the MSCKF filter~\cite{Mourikis07icra}.
Both these algorithms currently represent the state-of-the-art in \VIO, OKVIS for optimization-based approaches, and MSCKF for filtering methods.
We obtained the datasets as well as the trajectories computed with OKVIS and MSCKF from the authors of~\cite{Leutenegger15ijrr}.
We use the relative error metrics proposed in~\cite{Geiger12cvpr} to obtain error statistics. 
The metric evaluates the relative error by averaging the drift over trajectory segments of different length ($\{10,40,90,160,250,360\}$m in \Figure~\ref{fig:vicon_errors}).
Our approach exhibits less drift than the state-of-the-art, achieving $0.3$m drift on average over 360m traveled distance; OKVIS and MSCKF accumulate an average error of $0.7$m.
We observe significantly less drift in yaw direction in the proposed approach while the error in pitch and and roll direction is constant for all methods due to the observability of the gravity direction. 

\begin{figure}[t!]
  \centering
  \includegraphics[width=1.0\linewidth]{\figdirvin/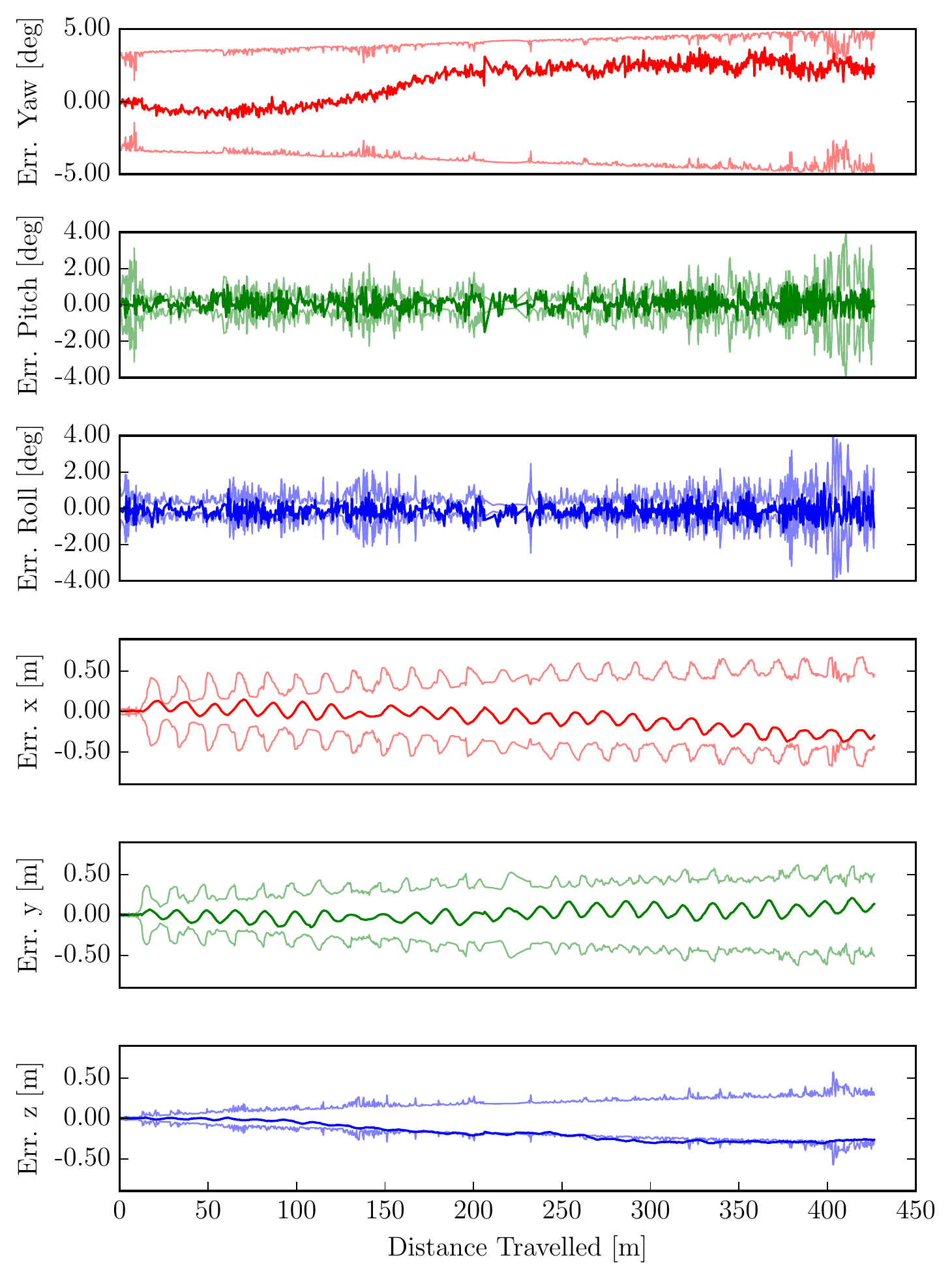}
  \caption{Orientation and position errors with 3$\sigma$ bounds for the real indoor experiment in \Figure\ref{fig:vicon_trajectory}.}
  \label{fig:vicon_consistency}
  \vspace{-15pt}
\end{figure}

We highlight that these three algorithms use different front-end feature tracking systems, which influence the overall performance of the approach. Therefore, while in Section~\ref{sec:experiments_synthetic} 
we discussed only aspects related to the preintegration theory, in this section we evaluate the proposed system as a whole 
(SVO, preintegration, structureless vision factors, iSAM2).

Evaluating consistency in real experiments by means of analysing the average NEES is difficult as one would have to evalate and average the results of multiple runs of the same trajectory with different realizations of sensor noise.
In Figure \ref{fig:vicon_consistency} we show the error plots with the 3-sigma bounds for a single run.
The result is consistent as the estimation errors remain within the bounds of the estimated uncertainty.
In this experiment, we aligned only the first frame of the trajectory with the vicon trajectory. Therefore, analyzing the drift over 400 meters is very prone to errors in the initial pose from the ground-truth or errors in the hand-eye calibration of the system.

% Timing
Figure \ref{fig:processing_time} illustrates the time required by the back-end to compute the full \MAP estimate, by running iSAM2 with 10 optimization iterations. 
The experiment was performed on a standard laptop (Intel i7, 2.4 GHz).
The average update time for iSAM2 is $10$ms.
The peak corresponds to the start of the experiment in which the camera was not moving.
In this case the number of tracked features becomes very large making the back-end slightly slower.
The SVO front-end requires approximately $3$ms to process a frame on the laptop while the back-end runs in a parallel thread and optimizes only keyframes. 
Although the processing times of OKVIS were not reported, the approach is described as computationally demanding~\cite{Leutenegger15ijrr}.
OKVIS needs to repeat IMU integration at every change of the linearization point, which we avoid by using the preintegrated IMU measurements.

%!TEX root = main.tex

\begin{figure}[t]
  \centering
  \hspace{-13.5px}
  \includegraphics[trim=20pt 10pt 30pt 10pt, clip, width=1.06\linewidth]{\figdirvin/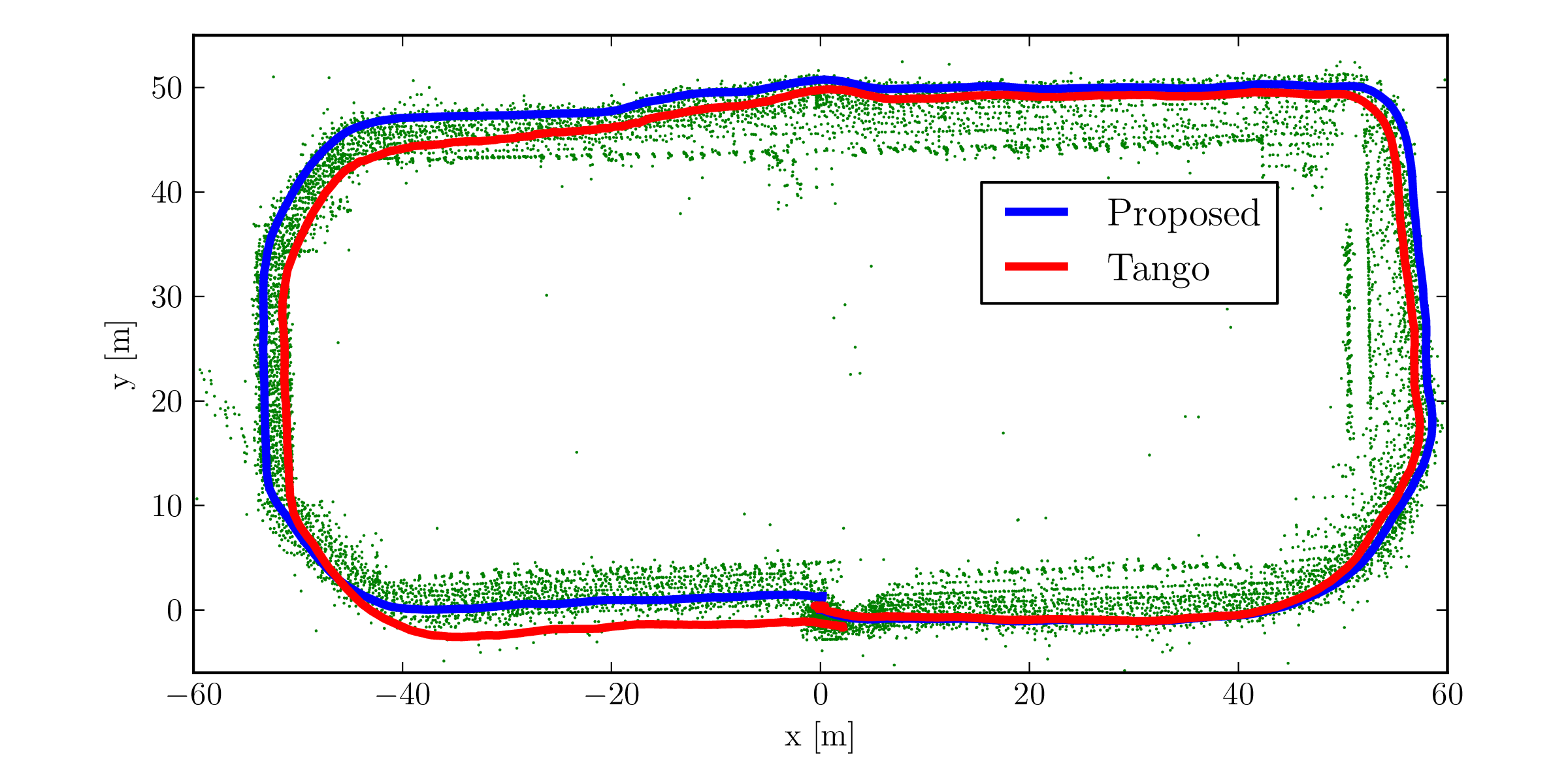} 
  \caption{Outdoor trajectory (length: $300\meters$) around a building with identical start and end point at coordinates $(0,0,0)$. The end-to-end error of the proposed approach is $1.0\meters$. Google Tango accumulated $2.2\meters$  drift. The green dots are the 3D points triangulated from our trajectory estimate.
  % obtained 
  %using our approach.
  %computed by the proposed approach.
  }
  \label{fig:outdoor_trajectory}
  \vspace{-5pt}
\end{figure}
%!TEX root = main.tex
\begin{figure}[t!]
  \centering
  \includegraphics[trim = 5mm 5mm 3mm 5mm, clip, width=1\linewidth]{\figdirvin/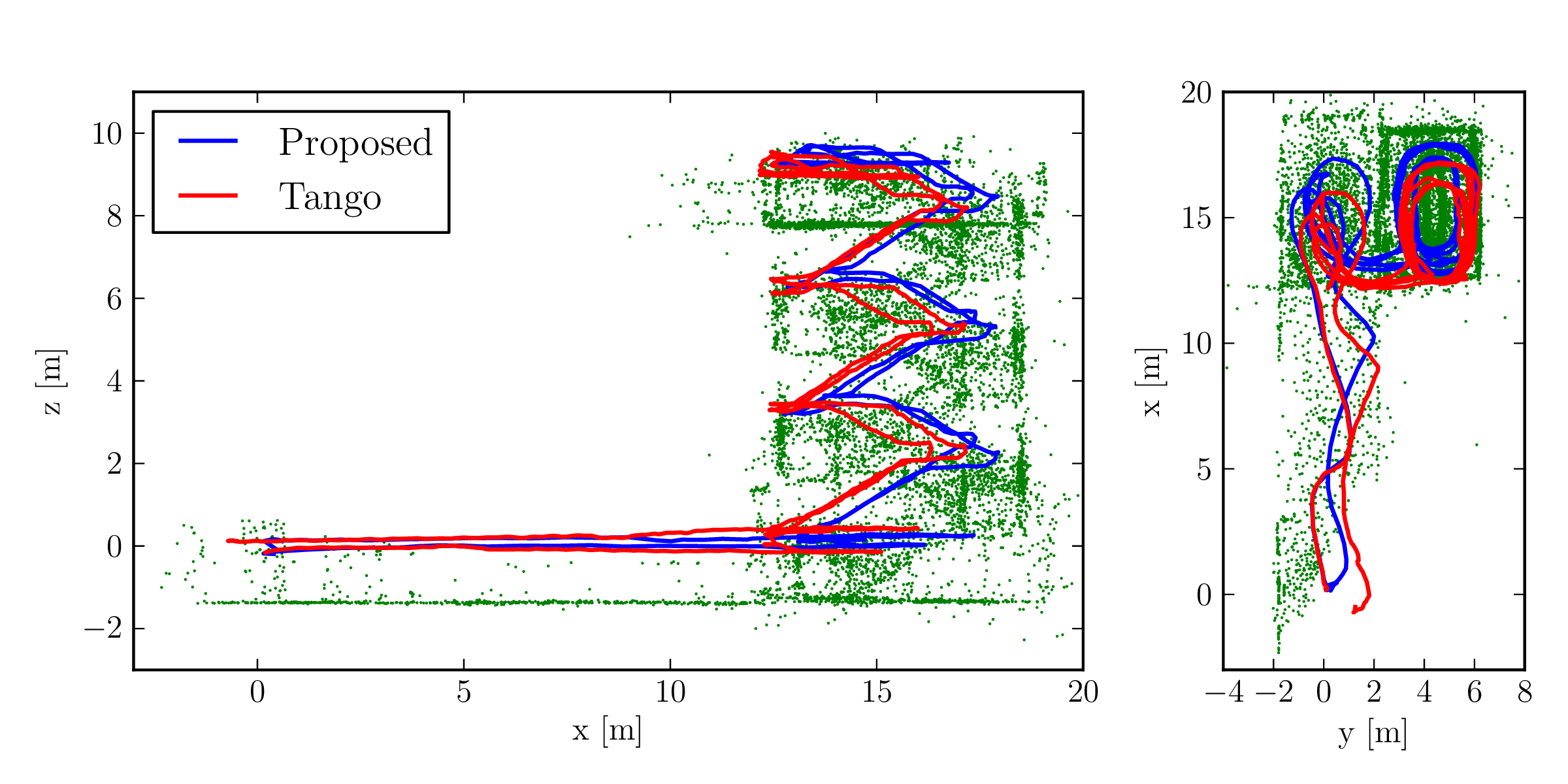}
  \caption{\label{fig:stairs}
  Real test comparing the proposed \VIO approach against Google \tango. 
  The $160$m-long  trajectory starts at $(0,0,0)$ (ground floor), goes up till the 3rd floor of a building, and 
  returns to the initial point. The figure shows a side view (left) and a top view (right) 
  of the trajectory estimates for our approach (blue) and \tango (red).
  Google \tango accumulates $1.4$m error, while the proposed approach only 
  has $0.5$m drift. 3D points triangulated from our trajectory estimate are shown in green for 
 visualization purposes.
   \vspace{-5pt}
}
\end{figure}

\subsubsection{Outdoor Experiments}
The second experiment is performed on an outdoor trajectory, and compares the proposed approach against the \emph{Google} \tango \emph{Peanut} sensor (mapper version 3.15), which is an \emph{engineered} \VIO system. 
We rigidly attached the VI-Sensor to a \tango device and walked around an office building. 
\Figure\ref{fig:outdoor_trajectory} depicts the trajectory estimates for our approach and Google \tango. 
The trajectory starts and ends at the same location, hence we can report the end-to-end error which is 1.5m for the proposed approach and 2.2m for the Google \tango sensor.

In \Figure\ref{fig:outdoor_trajectory}  we also show the estimated landmark positions (in green).
3D points are not estimated by our approach (which uses a structureless vision model), but are triangulated from our trajectory estimate for visualization purposes. 
  
The third experiment is the one in~\Figure~\ref{fig:stairs}.
The trajectory goes across three floors of an office building and eventually returns to the initial location on the ground floor.
Also in this case the proposed approach guarantees a very small end-to-end error ($0.5$m), while \tango accumulates $1.4$m error.

We remark that \tango and our system use different sensors, hence the reported end-to-end errors only allow for a qualitative comparison.
However, the IMUs of both sensors exhibit similar noise characteristics \cite{tangoSpecifications,adisSpecifications} and the \tango camera has a significantly larger field-of-view and better shutter speed control than our sensor.
Therefore, the comparison is still valuable to assess the accuracy of the proposed approach.

A video demonstrating the execution of our approach for the real experiments discussed in this section can be viewed at \url{\videolink}

%!TEX root = main.tex
\section{Conclusion} 
\label{sec:conclusion}

This paper proposes a novel preintegration theory, which provides a grounded way to model a large number of IMU measurements as a single motion constraint. 
Our proposal improves over related works that perform integration in a global frame, e.g.,~\cite{Leutenegger13rss,Mourikis07icra}, as we do not commit to a linearization point during integration. 
Moreover, it leverages the seminal work on preintegration~\cite{Lupton12tro}, bringing to maturity the preintegration and uncertainty propagation in~\SOthree.

As a second contribution, we discuss how to use the preintegrated IMU model in a \VIO pipeline; we adopt a structureless model for visual measurements which avoids optimizing over 3D landmarks.
Our \VIO approach uses iSAM2 to perform constant-time incremental smoothing.  

An efficient implementation of our approach requires $10$ms to perform inference (back-end), and $3$ms for feature tracking (front-end). Experimental results also confirm that our approach is more accurate than state-of-the-art alternatives, including filtering and optimization-based techniques.

We release the source-code of the IMU preintegration and the structurless vision factors in the GTSAM 4.0 optimization toolbox \cite{Dellaert12tr} and provide additional theoretical derivations and implementation details in the Appendix of this paper.
%%%%%%%%%%%%%%%%%%%%%%%%%%%%%%%%%%%%%%%%%%%%%%%%%%%%%%%%%%%%%%%%%%%%%%%%%%%%%%%%%%%%%%%%%%%%%%%%%%%%%%%%%

\small{
\vspace{0.2cm}
\textbf{Acknowledgments}
The authors gratefully acknowledge Stephen Williams and Richard Roberts 
for helping with an early implementation in GTSAM, and Simon Lynen and Stefan Leutenegger for providing datasets and results of their algorithms.
Furthermore, we are thankful to Mingyu Chen for the detailed comments and suggestions on the previous version of this paper.
}

\section*{Appendix}
%!TEX root = main.tex 
\subsection{Iterative Noise Propagation}
\label{sec:appendix_noisePropagation}

In this section we show that the computation of the preintegrated noise 
covariance, discussed in 
Section~\ref{sec:noisePropagation}, can be carried out in iterative form, 
which leads to simpler expressions and is more amenable for online inference.

Let us start from the preintegrated rotation noise in~\eqref{eq:firstOrderRotvecpert}.
To write $\rotvecpert_{ij}$ in iterative form, we simply take the last term ($k=j-1$) out of the sum and rearrange the terms:
\begin{align}
\label{eq:rotvecpert}
\rotvecpert_{ij} 
	& \simeq
	 \sum_{k=i}^{j-1} \preintRmeas_{k+1j}^\transpose \DExpk \noise^{gd}_k \Delta t \\
	 &= 
	 \sum_{k=i}^{j-2} \preintRmeas_{k+1j}^\transpose \DExpk \noise^{gd}_k \Delta t 
	 + \overbrace{\preintRmeas_{jj}^\transpose}^{= \eye_{3\times 3}} \DExp^{j-1} \noise^{gd}_{j-1} \Delta t 
	 \nonumber \\
	 &= 
	 \sum_{k=i}^{j-2} (\overbrace{\preintRmeas_{k+1j-1} \preintRmeas_{j-1j}}^{= \preintRmeas_{k+1j}})^\transpose \DExpk \noise^{gd}_k \Delta t
	 + \DExp^{j-1} \noise^{gd}_{j-1} \Delta t
   \nonumber\\
	 &=
	 \preintRmeas_{j-1j}^\transpose \sum_{k=i}^{j-2} \preintRmeas_{k+1j-1}^\transpose \DExpk \noise^{gd}_k \Delta t + 
	 \DExp^{j-1} \noise^{gd}_{j-1} \Delta t
	 \nonumber\\
	 &= 
	 \preintRmeas_{j-1j}^\transpose \rotvecpert_{ij-1} + 
	 \DExp^{j-1} \noise^{gd}_{j-1} \Delta t.
   \nonumber
\end{align}

Repeating the same process for $\velpert_{ij}$ in~\eqref{eq:noiseRVP}:
\begin{align}
\label{eq:velpert}
\velpert_{ij} 
	&=	
	\sum_{k=i}^{j-1} 
    \left[ -
	\preintRmeas_{ik} \left( \tilde\acc_k \!-\! \bias^a_i \right)^\wedge \rotvecpert_{ik} \Delta t 
	+ \preintRmeas_{ik} \noise^{ad}_k \Delta t
	\right]  \\
	&=
	\sum_{k=i}^{j-2} 
    \left[ -
	\preintRmeas_{ik} \left( \tilde\acc_k \!-\! \bias^a_i \right)^\wedge \rotvecpert_{ik} \Delta t 
	+ \preintRmeas_{ik} \noise^{ad}_k \Delta t
	\right] \nonumber\\
	&\quad \ -\!
	\preintRmeas_{ij-1} \left( \tilde\acc_{j-1} \!-\! \bias^a_i \right)^\wedge \rotvecpert_{i{j-1}} \Delta t 
	+ \preintRmeas_{i{j-1}} \noise^{ad}_{j-1} \Delta t
	\nonumber \\
	&\!\!\!\!\!\! =
	\velpert_{ij-1} \!-\! 
	\preintRmeas_{ij-1} \left( \tilde\acc_{j-1} \!-\! \bias^a_i \right)^\wedge \rotvecpert_{i{j-1}} \Delta t 
	\!+\! \preintRmeas_{i{j-1}} \noise^{ad}_{j-1} \Delta t \nonumber
\end{align}

Doing the same for $\tranpert_{ij}$ in~\eqref{eq:noiseRVP}, 
and noting that $\tranpert_{ij}$ can be written as a function of $\velpert_{ij}$ 
(\cf with the expression of $\velpert_{ij}$ in~\eqref{eq:noiseRVP}):
\begin{align}
\label{eq:tranpert}
&\tranpert_{ij} 
	\!=\!	
	\sum_{k=i}^{j-1} 
    \!\left[ \velpert_{ik} \Delta t
     \!-\! \frac{1}{2}
     \preintRmeas_{ik}  \left(\tilde\acc_k \!-\! \bias^a_i \right)^\wedge \rotvecpert_{ik} \Delta t^2 
    \!+\! \frac{1}{2} \preintRmeas_{ik} \noise^{ad}_k \Delta t^2 
    \!\right] 
    \nonumber \\
    &=	
	\sum_{k=i}^{j-2} 
    \left[ 
    \velpert_{ik} \Delta t
     - \frac{1}{2}
     \preintRmeas_{ik}  \left(\tilde\acc_k \!-\! \bias^a_i \right)^\wedge \rotvecpert_{ik} \Delta t^2 
    + \frac{1}{2} \preintRmeas_{ik} \noise^{ad}_k \Delta t^2 
    \right]  
    \nonumber \\
    &+\!
    \velpert_{ij-1} \Delta t
     \!-\! \frac{1}{2}
     \preintRmeas_{ij-1}\!  \left(\tilde\acc_{j-1} \!-\! \bias^a_i \right)^\wedge \! \rotvecpert_{ij-1} \Delta t^2 
     \!+\! \frac{1}{2} \preintRmeas_{i{j-1}} \noise^{ad}_{j-1} \Delta t^2 
    \nonumber \\
    & = 
    \tranpert_{ij-1} + \velpert_{ij-1} \Delta t 
    - \frac{1}{2}
     \preintRmeas_{ij-1}  \left(\tilde\acc_{j-1} \!-\! \bias^a_i \right)^\wedge \rotvecpert_{ij-1} \Delta t^2 \nonumber\\
    & \quad\quad\quad\quad \ + \frac{1}{2} \preintRmeas_{i{j-1}} \noise^{ad}_{j-1} \Delta t^2
\end{align}
Recalling that $\noise^{\Delta}_{ik} \doteq [\rotvecpert_{ik}, \velpert_{ik}, \tranpert_{ik}]$, 
and defining the IMU measurement noise $\noise^d_k \doteq [\noise^{gd}_k \; \; \noise^{ad}_k]$,\footnote{Both 
$\noise^{\Delta}_{ij}$ and $\noise^d_k$ are column vectors: we omit the transpose in the definition 
to keep notation simple.} 
we can finally write Eqs.~\eqref{eq:rotvecpert}-\eqref{eq:tranpert} in compact matrix form as:
\beq
  \label{eq:linearModel}
  \noise^{\Delta}_{ij} = \Amat_{j-1} \noise^{\Delta}_{ij-1} + \Bmat_{j-1} \noise^d_{j-1},
\eeq
From the linear model~\eqref{eq:linearModel} and given the covariance $\covimu_{\noise} \in \Real^{6 \times 6}$ of 
the raw IMU measurements noise $\noise^d_k$,
it is now possible to compute the preintegrated measurement covariance iteratively:
\beq
\label{eq:linearModelCovariance}
\covimu_{ij} = \Amat_{j-1} \covimu_{ij-1} \Amat_{j-1}^\transpose + \Bmat_{j-1} \covimu_{\noise} \Bmat_{j-1}^\transpose
\eeq
starting from initial conditions $\covimu_{ii} = \Zero_{9 \times 9}$.
%!TEX root = main.tex
\subsection{Bias Correction via First-Order Updates}
\label{app:biasUpdate}

In this section we provide a complete derivation of the first-order bias correction 
proposed in Section~\ref{sec:biasUpdate}.

Let us assume that we have computed the preintegrated variables 
at a given bias estimate $\biasFixed_i \doteq [\biasFixed^g_i\;\;\biasFixed^a_i]$, 
and let us denote the corresponding preintegrated measurements as 
\beq
  \pRot\doteq\preintRmeas_{ij}(\biasFixed_i), \
  \pVel\doteq\preintVmeas_{ij}(\biasFixed_i), \
  \pTran\doteq\preintPmeas_{ij}(\biasFixed_i).
\eeq
In this section we want to devise an expression to ``update'' $\pRot$, $\pVel$, $\pTran$ when our bias estimate changes. 

Consider the case in which we get a new estimate $\biasHat_i \leftarrow \biasFixed_i + \biaspert_i$, 
where $\biaspert_i$ is a \emph{small} correction w.r.t. the previous estimate $\biasFixed_i$.

We start with the bias correction for the preintegrated rotation measurement.
The key idea here is to write $\preintRmeas_{ij}(\biasHat_i)$ (the preintegrated measurement 
at the new bias estimate) as a function of $\pRot$ (the preintegrated 
measurement at the old bias estimate), ``plus'' a first-order correction.
Recalling \Equation\eqref{eq:preintVarR}, 
we write $\preintRmeas_{ij}(\biasHat_i)$ as:
\begin{align}
  \preintRmeas_{ij}(\biasHat_i)
    &= \prod_{k=i}^{j-1} \expmap\left(\left( \tilde\rotvel_k - \biasHat^g_i \right) \Delta t\right) 
\end{align}
Substituting $\biasHat_i = \biasFixed_i + \biaspert_i$ in the previous expression and using 
the first-order approximation~\eqref{eq:expFirstOrder} in each factor 
(we assumed small $\biaspert_i$):
\begin{align}
  \preintRmeas_{ij}(\biasHat_i)
    &= \prod_{k=i}^{j-1} \expmap\left(\left( \tilde\rotvel_k - (\biasFixed^g_i+\biaspert^{g}_i) \right) \Delta t\right) \\
    &\simeq \prod_{k=i}^{j-1} 
    \expmap\left(\left(\tilde\rotvel_k - \biasFixed^g_i \right) \Delta t\right) 
    \expmap\left( - \DExpk \; \biaspert^{g}_i \; \Delta t\right).
    \nonumber
\end{align}
Now, we rearrange the terms in the product, by ``moving'' the terms including $\biaspert^{gd}_i$ to the end, using the relation~\eqref{eq:adjointProperty2}:
\beq
  \preintRmeas_{ij}(\biasHat_i)
  \!=\! 
  \pRot
    \prod_{k=i}^{j-1} 
      \expmap\left(\ - \preintRmeas_{k+1j}(\biasFixed_i)^\transpose \DExpk \; \biaspert^{g}_i \; \Delta t\right),
\eeq
where we used the fact that by definition it holds that 
$
  \pRot
  %\doteq\preintRmeas_{ij}(\biasFixed_i)
  = \prod_{k=i}^{j-1} \expmap\left(\left(\tilde\rotvel_k - \biasFixed^{g}_i \right) \Delta t\right)
$.
Repeated application of the first-order approximation \eqref{eq:expExpansion} (recall that  $\biaspert^{g}_i$ is small, hence the right Jacobians are close to the identity) produces:
\begin{align} 
  \label{eq:updateR}
  \preintRmeas_{ij}(\biasHat_i)
    &\simeq
     \pRot\expmap\left(
      \sum_{k=i}^{j-1} 
        - \preintRmeas_{k+1j}(\biasFixed_i)^\transpose \DExpk \biaspert^{g}_i \Delta t
        \right)
        \nonumber \\
    &=\pRot
     \expmap\Big(\frac{\partial \pRot}{\partial\bias^g} \biaspert^g_i\Big)
\end{align}
Using~\eqref{eq:updateR} we can now update the preintegrated rotation measurement
$\preintRmeas_{ij}(\biasFixed_i)$ to get $\preintRmeas_{ij}(\biasHat_i)$ without repeating the integration.

Let us now focus on the bias correction of the preintegrated velocity $\preintVmeas_{ij}(\biasHat_i)$:
\begin{align}
  \!\!\!\!&\preintVmeas_{ij}(\biasHat_i) 
    = \sum_{k=i}^{j-1} \preintRmeas_{ik}(\biasHat_i) \left( \tilde\acc_k - \biasFixed^a_i - \biaspert^a_i \right)\Delta t \\
    &\!\!\stackrel{\eqref{eq:updateR}}{\simeq}
    \sum\limits_{k=i}^{j-1} \pRotk
    \expmap\Big(\frac{\partial \pRotk}{\partial\bias^g} \biaspert^g_i\Big) \left( \tilde\acc_k - \biasFixed^a_i - \biaspert^a_i \right)\Delta t
    \nonumber \\
    &\!\!\stackrel{\eqref{eq:expFirstOrder}}{\simeq}
    \sum\limits_{k=i}^{j-1} \pRotk
    \left(
      \eye + \Big(\frac{\partial \pRotk}{\partial\bias^g} \biaspert^g_i\Big)^\wedge 
    \right) \left( \tilde\acc_k - \biasFixed^a_i - \biaspert^a_i \right)\Delta t
    \nonumber \\
    &\!\!\stackrel{(a)}{\simeq}
    \pVel
    - \sum_{k=i}^{j-1} \pRotk \Delta t \biaspert^a_i 
    + \sum_{k=i}^{j-1} \pRotk
       \Big(\frac{\partial \pRotk}{\partial\bias^g} \biaspert^g_i\Big)^\wedge \left( \tilde\acc_k - \biasFixed^a_i \right)\Delta t \nonumber \\
    &\!\!\stackrel{\eqref{eq:skewProperty}}{=}
    \pVel
    - \sum_{k=i}^{j-1} \pRotk \Delta t \biaspert^a_i 
    - \sum_{k=i}^{j-1} \pRotk
    \left(\tilde\acc_k - \biasFixed^a_i \right)^\wedge \frac{\partial \pRotk}{\partial\bias^g} \Delta t \; \biaspert^g_i  
    \nonumber \\
    &\!\!= 
    \pVel + \frac{\partial \pVel}{\partial\bias^a} \biaspert^a_i
    + \frac{\partial \pVel}{\partial\bias^g} \biaspert^g_i
    \nonumber
\end{align}
Where for $(a)$, we used $\pVel = \sum_{k=i}^{j-1} \pRotk
 \left( \tilde\acc_k - \biasFixed^a_i \right)\Delta t$.
Exactly the same derivation can be repeated for $\preintPmeas_{ij}(\biasHat_i)$. 
Summarizing, the Jacobians used for the a posteriori bias update in \Equation~\eqref{eq:biasUpdate} are:
\begin{equation}
\begin{array}{ccl}
  \frac{\partial \pRot}{\partial\bias^g}
    &=& - \sum_{k=i}^{j-1} \left[ \preintRmeas_{k+1j}(\biasFixed_i)^\transpose \; \DExpk \;  \Delta t \right] \\
  \frac{\partial \pVel}{\partial\bias^a} 
    &=& - \sum_{k=i}^{j-1} \pRotk \Delta t
	  \nonumber \\
  \frac{\partial \pVel}{\partial\bias^g} 
    &=& - \sum_{k=i}^{j-1} \pRotk \left( \tilde\acc_k - \biasFixed^a_i \right)^\wedge  \frac{\partial \pRotk}{\partial\bias^g} \Delta t
	  \nonumber \\
  \frac{\partial \pTran}{\partial\bias^a}
    &=& \sum_{k=i}^{j-1}
    \frac{\partial \pVelk}{\partial\bias^a} \Delta t
    -\frac{1}{2} \pRotk \Delta t^2
    \nonumber \\
  \frac{\partial \pTran}{\partial\bias^g}
    &=& \sum_{k=i}^{j-1}
    \frac{\partial \pVelk}{\partial\bias^g} \Delta t
    -\frac{1}{2} \pRotk \left(\tilde\acc_k - \biasFixed^a_i \right)^\wedge   \frac{\partial \pRotk}{\partial\bias^g} \Delta t^2 \nonumber
\end{array}
\end{equation}
Note that the Jacobians can be computed incrementally, as new measurements arrive.

%!TEX root = main.tex
\subsection{Jacobians of Residual Errors}
\label{app:residualJacobians}

In this section we provide analytic expressions for the Jacobian matrices of the 
residual errors in \Equation~\eqref{eq:residuals}.
These Jacobians are crucial when using iterative optimization techniques (e.g., 
the Gauss-Newton method of Section~\ref{sec:GNmanifold}) to minimize the cost in \Equation\eqref{eq:cost}.

``Lifting'' the cost function (see Section~\ref{sec:GNmanifold}) consists in substituting the following retractions:
\begin{equation}
\begin{array}{rclrcl}
  \R_i    &\leftarrow&  \R_i \ \expmap(\rup_i),  &
  \R_j &\leftarrow&  \R_j \ \expmap(\rup_j), \\
  \tran_i &\leftarrow&  \tran_i + \R_i \tup_i, &
  \tran_j &\leftarrow&  \tran_j + \R_j \tup_j, \\  
  \vel_i  &\leftarrow&  \vel_i + \vup_i, &
  \vel_j  &\leftarrow&  \vel_j + \vup_i, \\
  \biaspert^g_i  &\leftarrow&  \biaspert^g_i + \bupg, &
  \biaspert^a_i  &\leftarrow&  \biaspert^a_i + \bupa, \\
\end{array}
\end{equation}
The process of lifting makes the residual errors a function defined on a vector space, on which it is easy to compute Jacobians.
Therefore, in the following sections we derive the Jacobians w.r.t. the vectors $\rup_i, \tup_i, \vup_i, \rup_j, \tup_j, \vup_j, \bupg, \bupa$.\\

\subsubsection{Jacobians of \texorpdfstring{$\residual_{\Delta\tran_{ij}}$}{dr\_tran}}

Since $\residual_{\Delta\tran_{ij}}$ is linear in $\biaspert^g_i$ and $\biaspert^a_i$, 
and the retraction is simply a vector sum, the Jacobians of $\residual_{\Delta\tran_{ij}}$ w.r.t. $\bupg, \bupa$ are simply the matrix coefficients of $\biaspert^g_i$ and $\biaspert^a_i$.
Moreover, $\R_j$ and $\vel_j$ do not appear in $\residual_{\Delta\tran_{ij}}$, hence the Jacobians w.r.t. $\rup_j, \vup_j$ are zero.
Let us focus on the remaining Jacobians:
\begin{align}
  \residual_{\Delta\tran_{ij}}(\tran_i\!+\!\R_i\tup_i)
    &=  \R_i^\transpose
        \left(
          \tran_j \!-\! \tran_i \!-\! \R_i \tup_i 
           \!-\! \vel_i \Delta t_{ij} \!-\! \frac{1}{2}\gravity \Delta t_{ij}^2
        \right) \!-\! C \nonumber\\
    &= \residual_{\Delta\tran_{ij}}(\tran_i) + (-\eye_{3\times 1}) \tup_i
\end{align}
\begin{align}
  \residual_{\Delta\tran_{ij}}(\tran_j\!+\!\R_j\tup_j)
    &=  \R_i^\transpose
        \left(
          \tran_j \!+\! \R_j\tup_j \!-\! \tran_i  \!-\! \vel_i \Delta t_{ij}
          \!-\! \frac{1}{2}\gravity \Delta t_{ij}^2
        \right) \!-\! C \nonumber\\
    &= \residual_{\Delta\tran_{ij}}(\tran_j) + (\R_i^\transpose \R_j) \tup_j
\end{align}
\begin{align}
  \residual_{\Delta\tran_{ij}}(\vel_i\!+\!\vup_i)
    &=  \R_i^\transpose
        \left(
          \tran_j \!-\! \tran_i  \!-\! \vel_i \Delta t_{ij} \!-\! \vup_i \Delta t_{ij} \!-\! \frac{1}{2}\gravity \Delta t_{ij}^2
        \right) \!-\! C\nonumber\\
    &= \residual_{\Delta\tran_{ij}}(\vel_i) + (- \R_i^\transpose \Delta t_{ij})\vup_i
\end{align}
\begin{align}
  \residual_{\Delta\tran_{ij}}&(\R_i \; \expmap(\rup_i)) \\
    &=(\R_i \expmap(\rup_i))^\transpose
      \left(
        \tran_j -\tran_i  - \vel_i \Delta t_{ij} - \frac{1}{2}\gravity \Delta t_{ij}^2
      \right)
    - C \nonumber\\
    &\stackrel{\eqref{eq:expFirstOrder}}{\simeq} 
     (\eye - \rup_i^\wedge)\R_i^\transpose
     \left(
       \tran_j -\tran_i  - \vel_i \Delta t_{ij} - \frac{1}{2}\gravity \Delta t_{ij}^2
     \right)
    - C \nonumber\\
    &\stackrel{\eqref{eq:skewProperty}}{=}
     \residual_{\Delta\tran_{ij}}(\R_i)
    +  \left(
        \R_i^\transpose 
          \left( \tran_j -\tran_i  - \vel_i \Delta t_{ij}
          - \frac{1}{2}\gravity \Delta t_{ij}^2\right)
        \right)^\wedge \rup_i. \nonumber
\end{align}
Where we used the shorthand
$C \doteq \mTran
          + \frac{\partial\pTran}{\partial\bias^g_i} \delta\bias^g_i
          + \frac{\partial\pTran}{\partial\bias^a_i} \delta\bias^a_i$.
Summarizing, the Jacobians of $\residual_{\Delta\tran_{ij}}$ are:
\begin{equation}
\begin{array}{ll}
\jacobian{\residual_{\Delta\tran_{ij}}}{\rup_i}
  = (\R_i^\transpose \!(\tran_j \!-\! \tran_i \!-\! \vel_i \Delta t_{ij} \!-\! \frac{1}{2}\gravity \Delta t_{ij}^2))^\wedge 
&
\jacobian{\residual_{\Delta\tran_{ij}}}{\rup_j} = \Zero 
\\
\jacobian{\residual_{\Delta\tran_{ij}}}{\tup_i} = -\eye_{3\times 1}
&
\jacobian{\residual_{\Delta\tran_{ij}}}{\tup_j} = \R_i^\transpose \R_j 
\\
\jacobian{\residual_{\Delta\tran_{ij}}}{\vup_i} = - \R_i^\transpose \Delta t_{ij} 
&
\jacobian{\residual_{\Delta\tran_{ij}}}{\vup_j} = \Zero  
\\
\jacobian{\residual_{\Delta\tran_{ij}}}{\bupa} = - \frac{\partial\pTran}{\partial\bias^a_i} 
&
\jacobian{\residual_{\Delta\tran_{ij}}}{\bupg} = - \frac{\partial\pTran}{\partial\bias^g_i}\nonumber
\end{array}
\end{equation}

\subsubsection{Jacobians of \texorpdfstring{$\residual_{\Delta\vel_{ij}}$}{r\_vel}}

As in the previous section, $\residual_{\Delta\vel_{ij}}$ is linear in $\biaspert^g_i$ and $\biaspert^a_i$, 
hence the Jacobians of $\residual_{\Delta\vel_{ij}}$ w.r.t. $\bupg, \bupa$ are simply the matrix coefficients of 
$\biaspert^g_i$ and $\biaspert^a_i$.
Moreover, $\R_j$, $\tran_i$, and $\tran_j$ do not appear in $\residual_{\Delta\vel_{ij}}$,
 hence the Jacobians w.r.t. $\rup_j, \tup_i, \tup_j$ are zero. The 
 remaining Jacobias are computed as:

\begin{align}
  \residual_{\Delta\vel_{ij}}(\vel_i + \vup_i)
    &= \R_i^\transpose
       \left( \vel_j-\vel_i-\vup_i - \gravity\Delta t_{ij}\right)
     - D \nonumber\\
    &= \residual_{\Delta\vel}(\vel_i) -\R_i^\transpose \vup_i
\end{align}
\begin{align}
  \residual_{\Delta\vel_{ij}}(\vel_j + \vup_j)
      &= \R_i^\transpose \left(\vel_j + \vup_j-\vel_i - \gravity\Delta t_{ij}\right)
      - D \nonumber\\
      &= \residual_{\Delta\vel}(\vel_j) + \R_i^\transpose \vup_j
\end{align}
\begin{align}
  \residual_{\Delta\vel_{ij}}&(\R_i \; \expmap(\rup_i))
    = \left(
        \R_i \; \expmap(\rup_i) \right)^\transpose \left(\vel_j-\vel_i - \gravity\Delta t_{ij}
      \right)
    - D \nonumber\\
    &\stackrel{\eqref{eq:expFirstOrder}}{\simeq} 
      (\eye-\rup_i^\wedge) \R_i^\transpose \left(\vel_j-\vel_i - \gravity\Delta t_{ij}\right)
    - D \nonumber\\
    &\stackrel{\eqref{eq:skewProperty}}{=} 
      \residual_{\Delta\vel}(\R_i) 
       + \left(\R_i^\transpose \left(\vel_j-\vel_i - \gravity\Delta t_{ij}\right)\right)^\wedge \rup_i,
\end{align}
with $D \doteq \left[
         \mVel
         + \frac{\partial\pVel}{\partial\bias^g_i} \delta\bias^g_i
         + \frac{\partial\pVel}{\partial\bias^a_i} \delta\bias^a_i
       \right]$. Summarizing, the Jacobians of $\residual_{\Delta\vel_{ij}}$ are:

\begin{equation}
\begin{array}{ll}
\jacobian{\residual_{\Delta\vel_{ij}}}{\rup_i} = \left(\R_i^\transpose \left(\vel_j-\vel_i - \gravity\Delta t_{ij}\right)\right)^\wedge 
&
\jacobian{\residual_{\Delta\vel_{ij}}}{\rup_j} = \Zero
\\
\jacobian{\residual_{\Delta\vel_{ij}}}{\tup_i} = \Zero
&
\jacobian{\residual_{\Delta\vel_{ij}}}{\tup_j} = \Zero
\\
\jacobian{\residual_{\Delta\vel_{ij}}}{\vup_i} = -\R_i^\transpose   
&
\jacobian{\residual_{\Delta\vel_{ij}}}{\vup_j} = \R_i^\transpose 
\\
\jacobian{\residual_{\Delta\vel_{ij}}}{\bupa} = - \frac{\partial\pVel}{\partial\bias^a_i}
&
\jacobian{\residual_{\Delta\vel_{ij}}}{\bupg} = - \frac{\partial\pVel}{\partial\bias^g_i}
\nonumber
\end{array}
\end{equation}

\subsubsection{Jacobians of \texorpdfstring{$\residual_{\Delta\R_{ij}}$}{r\_R}}
The derivation of the Jacobians of $\residual_{\Delta\R_{ij}}$ is slightly more involved.
We first note that $\tran_i, \tran_j, \vel_i, \vel_j, \biaspert^a_i$ do not appear in the 
expression of $\residual_{\Delta\R_{ij}}$, hence the corresponding Jacobians are zero. 
The remaining Jacobians can be computed as follows:

\begin{align}
  \residual_{\Delta\R_{ij}}(\R_i \; \expmap(\rup_i))
  &=  \logmap\left( 
    \left( \preintRmeas_{ij}(\biasFixed^g_i) E  \right)^\transpose 
     \left( \R_i \; \expmap(\rup_i) \right)^\transpose \R_j  \right)
  \nonumber\\
  &=  \logmap\left( 
    \left( \preintRmeas_{ij}(\biasFixed^g_i) E  \right)^\transpose 
      \expmap(-\rup_i)  \R_i^\transpose \R_j  \right)
  \nonumber\\
  &
  \overset{\eqref{eq:adjointProperty2}}{=}  
  \logmap\left( 
    \left( \preintRmeas_{ij}(\biasFixed^g_i) E  \right)^\transpose 
    \R_i^\transpose \R_j   \expmap(-\R_j^\transpose \R_i  \rup_i)  \right)
  \nonumber\\
  &\overset{\eqref{eq:logExpansion}}{\simeq}  
    \residual_{\Delta\R}(\R_i) - \DLog(\residual_{\Delta\R}(\R_i)) \R_j^\transpose \R_i  \rup_i
\end{align}
\begin{align}
  \residual_{\Delta\R_{ij}}(\R_j \; \expmap(\rup_j)) 
  &=  \logmap\left( 
    \left( \preintRmeas_{ij}(\biasFixed^g_i) E  \right)^\transpose 
     \R_i^\transpose (\R_j \; \expmap(\rup_j))  \right)
 \nonumber\\
 &\overset{\eqref{eq:logExpansion}}{\simeq}  
  \residual_{\Delta\R}(\R_j)
   + \DLog(\residual_{\Delta\R}(\R_j)) \rup_j
\end{align}
\begin{align}
\textstyle
  \residual_{\Delta\R_{ij}}&(\biaspert^g_i+\bupg)\\
  	&= 
    \logmap\Big( 
      \Big(
        \preintRmeas_{ij}(\biasFixed^g_i)
          \expmap\big(
            \frac{\partial\pRot}{\partial\bias^g} (\biaspert^g_i + \bupg) 
          \big)
       \Big)^\transpose \R_i^\transpose \R_j
     \Big) 
     \nonumber\\
    &\overset{\eqref{eq:expExpansion}}{\simeq}
      \logmap\Big( 
        \Big(
          \preintRmeas_{ij}(\biasFixed^g_i) 
          \ E \
          \expmap\big(
            \DExp^b \frac{\partial\pRot}{\partial\bias^g}  \bupg
          \big) 
        \Big)^\transpose 
        \R_i^\transpose \R_j
     \Big) 
     \nonumber\\
    &= 
      \logmap\Big( 
      \expmap\Big( -\DExp^b \frac{\partial\pRot}{\partial\bias^g}  \bupg \Big)
      \left( 
      \preintRmeas_{ij}(\biasFixed^g_i) 
      \ E
       \right)^\transpose 
       \R_i^\transpose \R_j  \Big) 
       \nonumber\\
       &= 
       \logmap\Big( 
        \expmap\Big( -\DExp^b \frac{\partial\pRot}{\partial\bias^g}  \bupg \Big)
      \expmap\left( 
      \residual_{\Delta\R_{ij}}(\biaspert^g_i)
       \right) 
       \Big) 
       \nonumber\\
    &\overset{\eqref{eq:adjointProperty2}}{=}   
    	\logmap\Big( 
    	\expmap\big( \residual_{\Delta\R_{ij}}(\biaspert^g_i) \big)
      \nonumber\\
      &\quad\quad\cdot
         \expmap\big( - \expmap\left( \residual_{\Delta\R_{ij}}(\biaspert^g_i) \right)^\transpose 
         \DExp^b
         \frac{\partial\pRot}{\partial\bias^g}  \bupg \big)\Big)
         \nonumber\\
       &\overset{\eqref{eq:logExpansion}}{\simeq}   
    	\residual_{\Delta\R_{ij}}(\biaspert^g_i)
      \nonumber\\
  	&\quad\quad - \DLog\left( \residual_{\Delta\R_{ij}}(\biaspert^g_i) \right)
  	 \expmap\left( \residual_{\Delta\R_{ij}}(\biaspert^g_i) \right)^\transpose 
       \DExp^b
       \frac{\partial\pRot}{\partial\bias^g} \ \bupg, \nonumber 
\end{align}
where we used the shorthands $E \doteq \expmap\left(\frac{\partial\pRot}{\partial\bias^g} \biaspert^g\right)$ and $\DExp^b \doteq  \DExp\left( \frac{\partial\pRot}{\partial\bias^g} \biaspert^g_i \right)$.
In summary, the Jacobians of $\residual_{\Delta\R_{ij}}$ are:

\begin{equation}
\begin{array}{ll}
\jacobian{\residual_{\Delta\R_{ij}}}{\rup_i} = 
  - \DLog(\residual_{\Delta\R}(\R_i)) \R_j^\transpose \R_i
&
\jacobian{\residual_{\Delta\R_{ij}}}{\tup_i} = \Zero
\\
\jacobian{\residual_{\Delta\R_{ij}}}{\vup_i} = \Zero   
&
\jacobian{\residual_{\Delta\R_{ij}}}{\rup_j} = \DLog(\residual_{\Delta\R}(\R_j)) 
\\
\jacobian{\residual_{\Delta\R_{ij}}}{\tup_j} = \Zero 
&
\jacobian{\residual_{\Delta\R_{ij}}}{\vup_j} = \Zero 
\\
\jacobian{\residual_{\Delta\R_{ij}}}{\bupa} = \Zero
&
\jacobian{\residual_{\Delta\R_{ij}}}{\bupg} =  \alpha
\end{array}
\end{equation}
with $
\alpha = -\DLog\left( \residual_{\Delta\R_{ij}}(\biaspert^g_i) \right)
         \expmap\left( \residual_{\Delta\R_{ij}}(\biaspert^g_i) \right)^\transpose 
         \DExp^b \frac{\partial\pRot}{\partial\bias^g}$.

%!TEX root = main.tex
\subsection{Structureless Vision Factors: Null Space Projection}
\label{sec:appendix_structureless}

In this section we provide a more efficient implementation of the structureless 
vision factors, described in Section~\ref{sec:structureless}.

Let us consider \Equation\eqref{eq:cameraOnly}.
Recall that
$
	\Qmat \doteq (\eye - \Emat_{l} (\Emat_{l}^\transpose \Emat_{l})\inv \Emat_{l}^\transpose) \in \Real^{2n_l \times 2n_l}
$ 
is an \emph{orthogonal projector} of $\Emat_{l}$, where $n_l$ is the number of cameras observing landmark $l$.
Roughly speaking, $\Qmat$ projects any vector in $\Real^{2n_l}$ to the null space of the matrix $\Emat_{l}$.
Since $\Emat_{l} \in \Real^{2 n_l \times 3}$ has rank 3, the dimension of its null space is $2n_l-3$. 
Any basis $\Emat^\perp_{l} \in \Real^{2n_l \times 2n_l-3}$ of the null space of $\Emat_{l}$ satisfies the following relation~\cite{Meyer00siam}:
\beq
\label{eq:orthogonalProjectors}
\Emat^\perp_{l} \left( (\Emat^\perp_{l})^\transpose \Emat^\perp_{l} \right)\inv (\Emat^\perp_{l})^\transpose = 
  \eye - \Emat_{l} (\Emat_{l}^\transpose \Emat_{l})\inv \Emat_{l}^\transpose.
\eeq
A basis for the null space can be easily computed from $\Emat_{l}$ using SVD.
Such basis is \emph{unitary}, i.e.,  satisfies $(\Emat^\perp_{l})^\transpose \Emat^\perp_{l} = \eye$.
Substituting~\eqref{eq:orthogonalProjectors} into~\eqref{eq:cameraOnly}, and recalling that 
$\Emat^\perp_{l}$ is a unitary matrix, we obtain:
\begin{align}
\label{eq:cameraOnlyJac}
&\sum_{l = 1}^L \| \Emat^\perp_{l} (\Emat^\perp_{l})^\transpose
\left( \Fmat_{l} \; \posepert_{\States(l)} - \bvec_{l} \right)  \|^2 \\
&= 
\sum_{l = 1}^L
\big(
	\Emat^\perp_{l} (\Emat^\perp_{l})^\transpose 
	\!( \Fmat_{l} \; \posepert_{\States(l)} - \bvec_{l} )
\big)^\transpose  
\!\big( \Emat^\perp_{l} (\Emat^\perp_{l})^\transpose 
( \Fmat_{l} \; \posepert_{\States(l)} - \bvec_{l}) \big) 
 \nonumber \\
&= \sum_{l = 1}^L  
\left( \Fmat_{l} \; \posepert_{\States(l)} - \bvec_{l} \right)^\transpose
 \Emat^\perp_{l}
 \overbrace{(\Emat^\perp_{l})^\transpose  \Emat^\perp_{l}}^{=\eye_{3\times 3}}
 (\Emat^\perp_{l})^\transpose 
\left( \Fmat_{l} \; \posepert_{\States(l)} - \bvec_{l} \right)  
\nonumber \\ 
&%=
%\sum_{l = 1}^L  
%\left( \Fmat_{l} \; \posepert_{\States(l)} - \bvec_{l} \right)^\transpose
% \Emat^\perp_{l}  (\Emat^\perp_{l})^\transpose 
%\left( \Fmat_{l} \; \posepert_{\States(l)} - \bvec_{l} \right)
=\sum_{l = 1}^L  
\| (\Emat^\perp_{l})^\transpose 
\left( \Fmat_{l} \; \posepert_{\States(l)} - \bvec_{l} \right) \|^2 \nonumber
\end{align}
which is an alternative representation of the cost function~\eqref{eq:cameraOnly}. 
This representation is usually preferable from a computational standpoint, as it does not include matrix inversion and can be computed using a smaller number of matrix multiplications.

%!TEX root = main.tex
\subsection{Rotation Rate Integration Using Euler Angles}
\label{sec:appendix_euler}

In this section, we recall how to integrate rotation rate measurements using the Euler angle parametrization. 
Let $\tilde\rotvel_k$ be the rotation rate measurement at time $k$ and $\noise^{g}_k$ be the corresponding noise.
Then, given the vector of Euler angles at time $k$, namely $\eulervec_k \in \Real^3$, 
 we can integrate the rotation rate measurement $\tilde\rotvel_k$ and get $\eulervec_{k+1}$ as follows:
\beq
  \label{eq:eulerIntegration}
  \eulervec_{k+1} = \eulervec_k + [E'(\eulervec_k)]^{-1} (\tilde\rotvel_k - \noise^{g}_k)\Delta t,
\eeq
where the matrix $E'(\eulervec_k)$ is the \emph{conjugate Euler angle rate matrix}~\cite{Diebel06}.
The covariance of $\eulervec_{k+1}$ can be approximated by a
first-order propagation as: 
\beq
\covimu^{\text{Euler}}_{k+1} = \Amat_{k} \covimu^{\text{Euler}}_{k} \Amat_{k}^\transpose + \Bmat_{k} \covimu_{\noise} \Bmat_{k}^\transpose
\eeq
where $\Amat_k \doteq \Ithree + \frac{\partial[E'(\eulervec_k)]^{-1}}{\partial\eulervec_k}\Delta t$, $\Bmat_k = - [E'(\eulervec_k)]^{-1} \Delta t$, and $\covimu_{\noise}$ is the covariance of the measurement noise $\noise^{gd}_k$.

\normalsize
\bibliographystyle{unsrtnat}
\bibliography{main}

\newpage

\begin{IEEEbiography}[{\includegraphics[width=1in,height=1.25in,clip,keepaspectratio]{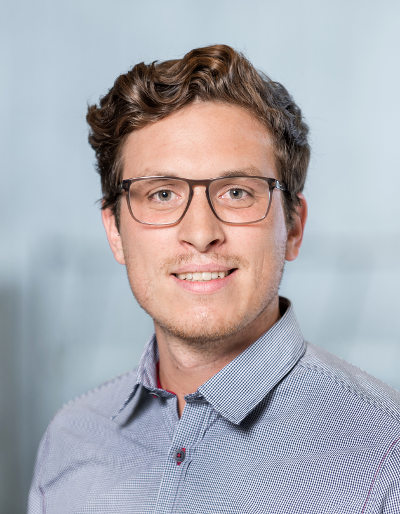}}]{Christian Forster} (1986, Swiss) obtained his Ph.D. in Computer Science (2016) at the University of Zurich under the supervision of Davide Scaramuzza. Previously, he received a B.Sc. degree in Mechanical Engineering (2009) and a M.Sc. degree in Robotics, Systems and Control (2012) at ETH Zurich, Switzerland. In 2011, he was a visiting researcher at CSIR (South Africa), and in 2014 at Georgia Tech in the group of Frank Dellaert. He is broadly interested in developing real-time computer vision algorithms that enable robots to perceive the three dimensional environment.
\end{IEEEbiography}

\begin{IEEEbiography}[{\includegraphics[width=1in,height=1.25in,clip,keepaspectratio]{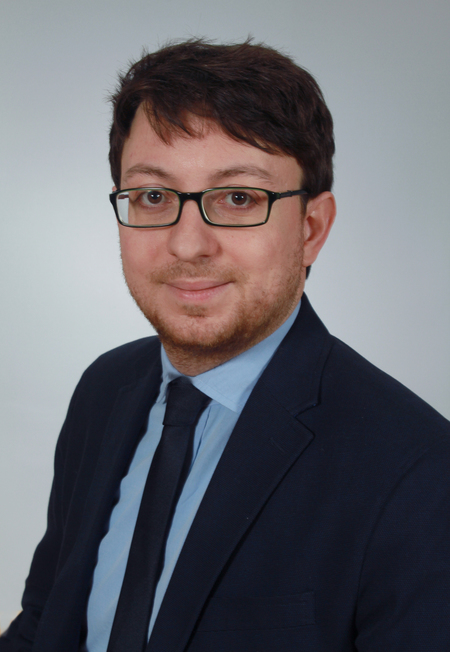}}]{Luca Carlone} is a research scientist in the Laboratory for Information and Decision Systems at the Massachusetts Institute of Technology. Before joining MIT, he was a postdoctoral fellow at Georgia Tech (2013-2015),  a visiting researcher at the University of California Santa Barbara (2011), and a visiting researcher at the University of Zaragoza (2010). He got his Ph.D. from Politecnico di Torino, Italy, in 2012.  His research interests include nonlinear estimation, optimization, and control applied to robotics, with special focus on localization, mapping, and decision making for navigation in single and multi robot systems.
\end{IEEEbiography}

\begin{IEEEbiography}[{\includegraphics[width=1in,height=1.25in,clip,keepaspectratio]{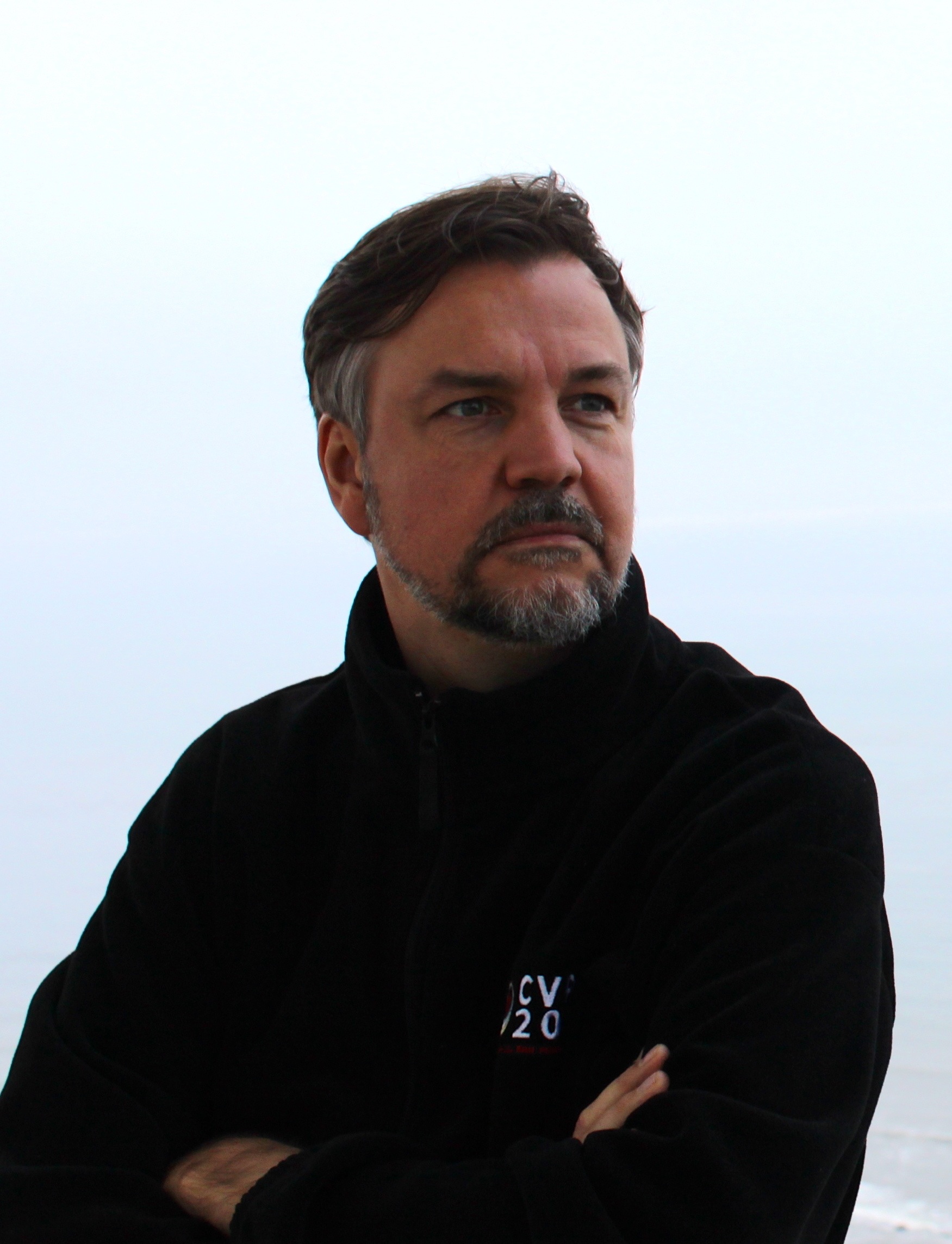}}]{Frank Dellaert} is Professor in the School of Interactive Computing at Georgia Tech. He graduated from Carnegie Mellon in 2001 with a Ph.D. in Computer Science, after obtaining degrees from Case Western Reserve University and K.U. Leuven before that. Professor Dellaert's research focuses on large-scale inference for autonomous robot systems, on land, air, and in water. He pioneered the use of several probabilistic methods in both computer vision and robotics. With Dieter Fox and Sebastian Thrun, he has introduced the Monte Carlo localization method for estimating and tracking the pose of robots, which is now a standard and popular tool in mobile robotics. Most recently, he has investigated 3D reconstruction in large-scale environments by taking a graph-theoretic view, and pioneered the use of sparse factor graphs as a representation in robot navigation and mapping. He and his group released an open-source large-scale optimization toolbox, GTSAM.
\end{IEEEbiography}

\begin{IEEEbiography}[{\includegraphics[width=1in,height=1.25in,clip,keepaspectratio]{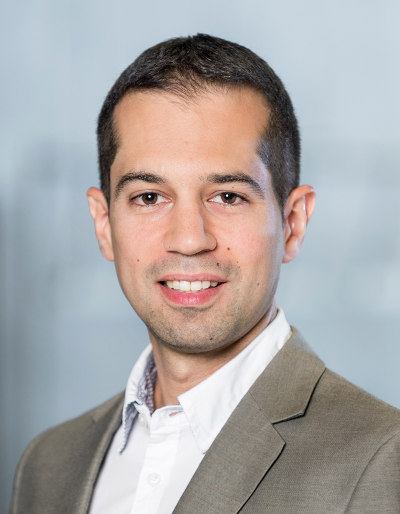}}]{Davide Scaramuzza} (1980, Italian) is Professor of Robotics at the University of Zurich, where he does research at the intersection of robotics, computer vision, and neuroscience. He did his PhD in robotics and computer Vision at ETH Zurich and a postdoc at the University of Pennsylvania. From 2009 to 2012, he led the European project sFly, which introduced the world’s first autonomous navigation of micro drones in GPS-denied environments using visual-inertial sensors as the only sensor modality. For his research contributions, he was awarded an SNSF-ERC Starting Grant, the IEEE Robotics and Automation Early Career Award, and a Google Research Award. He coauthored the book Introduction to Autonomous Mobile Robots (published by MIT Press). 
\end{IEEEbiography}

\end{document}